\documentclass{kais}

\usepackage{url}
\usepackage{graphicx}

\usepackage{amsmath}
\usepackage{xcolor}
\usepackage{booktabs}
\usepackage{comment}

\usepackage{textcomp}
\usepackage{siunitx}
\usepackage{amssymb}

\usepackage{cite}

\usepackage{array} 
\usepackage{floatrow}
\usepackage{adjustbox}
\usepackage{arydshln}
\usepackage{multirow}
\usepackage[ruled,linesnumbered]{algorithm2e}
\usepackage{subcaption}
\usepackage{hyperref}
\usepackage{enumitem}
\usepackage{pbox}

\usepackage[colorinlistoftodos,prependcaption]{todonotes}

\usepackage{mathtools}

\DeclareMathOperator*{\argminC}{\arg\min}   

\usepackage{appendix}
\usepackage{authblk}

\received{xxx}
\revised{xxx}
\accepted{xxx}

\pubyear{2000}
\pagerange{\pageref{firstpage}--\pageref{lastpage}}
\volume{xxx}

\begin{document} 
\label{firstpage}
 
%
\title{Parity-based Cumulative Fairness-aware Boosting}

 
 \author[*]{Vasileios Iosifidis\thanks{iosifidis@el3s.de}}
\author[**]{Arjun Roy\thanks{roy@l3s.de}}
\author[**]{Eirini Ntoutsi\thanks{eirini.ntoutsi@fu-berlin.de}}
\affil[*]{Leibniz University Hannover \& L3S Research Center, Germany.}
\affil[**]{Freie Universität Berlin \& L3S Research Center, Germany.}
\maketitle

\begin{abstract}
Data-driven AI systems can lead to discrimination on the basis of protected attributes like gender or race. One reason for this behavior is the encoded societal biases in the training data (e.g., females are underrepresented), which is aggravated in the presence of unbalanced class distributions (e.g., ``granted'' is the minority class). 
State-of-the-art fairness-aware machine learning approaches focus on preserving the \emph{overall} classification accuracy while improving fairness. In the presence of class-imbalance, such methods may further aggravate the problem of discrimination by denying an already underrepresented group (e.g., \textit{females}) the fundamental rights of equal social privileges (e.g., equal credit opportunity).

To this end, we propose AdaFair, a fairness-aware boosting ensemble that changes the data distribution at each round, taking into account not only the class errors but also the fairness-related performance of the model defined cumulatively based on the partial ensemble. Except for the in-training boosting of the  group discriminated over each round, AdaFair directly tackles imbalance during the post-training phase by optimizing the number of ensemble learners for balanced error performance (BER). AdaFair can facilitate different parity-based fairness notions and mitigate effectively discriminatory outcomes. Our experiments show that our approach can achieve parity in terms of statistical parity, equal opportunity, and disparate mistreatment while maintaining good predictive performance for all classes.
\end{abstract}


\section{Introduction}%
\label{sec:intro}

Increasing concerns about accountability, fairness, and transparency of AI-based decision making systems, especially for domains of high societal impact, have been raised over the recent years~\cite{united2014big} as a plethora of discrimination incidents by such systems have been reported
~\cite{AmazonPrime,datta2015automated,airbnb,sweeney2013discrimination,larson2016we}.  The discriminations are mainly against individuals or groups who share specific characteristics like gender or race (referred to as \emph{protected groups} hereafter) compared to other groups (referred to as \emph{non-protected groups} hereafter). A growing body of research has been proposed over the recent years to address fairness and algorithmic discrimination. These methods propose ``interventions'' at the input data (the so-called, pre-processing methods), learning algorithm (the so-called, in-processing methods), or the output model (the so-called post-processing methods) to ensure that the model decisions are not only correct in terms of predictive performance but also fair according to some definition of fairness. 

The vast majority of these methods, e.g.,~\cite{krasanakis2018adaptive,zafar2017fairness,calders2009building,calmon2017optimized,kamiran2012data,hardt2016equality,fish2016confidence,kamiran2009classifying,kamiran2010discrimination}, focus on optimizing for fairness while maintaining an \emph{overall} high classification performance.
In the case of class-imbalance though, overall performance is not a good performance indicator 
for the different classes. In the binary classification case, for example, it would ignore the performance in the minority (also called, positive) class. 
Such approaches might achieve fairness by getting an overall performance parity between the protected and non-protected groups; however, the predictive performance of the model on the minority class is poor as confirmed by our experiments. 

Class-imbalance is an inherited problem of fairness; therefore, tackling fairness also requires tackling imbalance~\cite{galar2012review,iosifidis2020semi}. Our proposed approach, AdaFair, overcomes this issue and achieves fairness while preserving good predictive performance across all classes. AdaFair is based on AdaBoost and extends its instance weighting strategy in each round based on the fairness of the partial ensemble. 
This way, in each round, the corresponding weak learner focuses on both complex classification examples (as in traditional boosting) and on the discriminated group, which is dynamically identified in each boosting round.
The effect on the instance weighting is evaluated based on a cumulative notion of fairness that considers the fairness behaviour of the partial ensemble model until the particular round. 
Moreover, post-training, we select the best sequence of weak learners, achieving high performance across all classes and ensuring fairness. Our experiments indicate that AdaFair can provide the best trade-off among state-of-the-art fairness-aware methods in terms of balanced error rate and fairness.

Our contributions are summarized as follows: i) we propose AdaFair, a fairness-aware boosting method, that achieves \emph{parity} between the protected and non-protected groups (thus achieving fairness) while maintaining good predictive performance for \emph{all} classes (thus tackling class-imbalance). 
ii) We define the notion of cumulative fairness for the ensemble for three different parity-based fairness notions: statistical parity, equal opportunity, and disparate mistreatment.
Based on that, we propose a dynamic instance re-weighting schema that considers both predictive- and fairness-performance of the model.
iii) We show the superiority of our
cumulative notions of fairness vs non-cumulative alternatives in terms of model performance and stability. 

This work is an extension of our previous work~\cite{iosifidis2019adafair}. The significant changes include: i) extending AdaFair to facilitate two more parity-based notions, namely statistical parity and equal opportunity, ii) introducing the cumulative versions of the newly added fairness notions including the new instance weighting schemes, iii) providing a theoretical analysis on bounding the training error of AdaFair and iv) comparison of AdaFair to current state-of-the-art fairness-aware methods \cite{iosifidis2019fae,fish2016confidence}. Data and source code are made available\footnote{AdaFair (source code and data) available at: \url{https://iosifidisvasileios.github.io/AdaFair}.}.

\section{Related Work}%
\label{sec:related}
\noindent \textbf{Fairness notions.}
``Fair is not fair everywhere"~\cite{schafer2015fair}, but rather the definition of fairness depends on context. As a result, 
more than twenty fairness notions exist for fairness in classification~\cite{verma2018fairness,ntoutsi2020bias} 
. 
One of the earliest measures of discrimination, the so-called \textit{statistical} or \textit{demographic parity}~\cite{kamiran2012data}, measures the percentage difference in the positive predictions, , e.g., grant a loan application, between the protected and non-protected groups. However, this definition only requires a balanced representation of both groups in the positive class ignoring whether the selected instances are qualified or not~\cite{dwork2012fairness}. 
A more recent measure, called equal opportunity~\cite{hardt2016equality,verma2018fairness}, alleviates this pitfall by measuring the percentage difference between true positive rates for both groups. 
Disparate mistreatment~\cite{zafar2017fairness} extends equal opportunity by considering the difference of correctly classified instances between protected and non-protected groups for all classes. 

\noindent \textbf{Pre-processing approaches.} 
One of the most common causes of ML discrimination arises from biases in historical data. Pre-processing methods aim to deal with this issue by changing the underlying data to restore balance between the protected and non-protected groups. 
Methods falling under this category are typically model-agnostic; therefore, any classifier is applicable after pre-processing. Example methods include 
\textit{massaging}~\cite{kamiran2009classifying} that changes the class-labels of carefully selected instances, \textit{re-weighting}~\cite{calders2009building} that assigns different weights to the instances of the protected group, \textit{uniform- or preferential-sampling}~\cite{kamiran2012data} and \textit{data augmentation} \cite{iosifidisdealing} that enriches the minority group through pseudo-instances. Another interesting line of works by~\cite{calmon2017optimized,hu2020fairnn}, perform data transformations to eliminate existing correlations between class labels and protected attributes.
Some of these approaches have been also extended to non-stationary data~\cite{iosifidis2019fairness}. 

\noindent \textbf{In-processing approaches.} 
These approaches aim to mitigate discrimination during model training by extending the objective function of the learner to account for discrimination using regularization or constraints. In~\cite{kamiran2010discrimination,zhang2019faht}, for example, the entropy-based splitting criterion for decision tree induction is modified based on statistical parity to also consider the fairness of the splitting decisions. In~\cite{kamishima2012fairness} a regularization approach is proposed that scales down the correlation (mutual information) of the sensitive features and the class attribute to avoid outcomes based on these features. \cite{dwork2012fairness} introduces the notion of ``individual fairness-constraints" that impose similar treatment to similar instances. 
In \cite{zafar2017fairness}, a set of constraints which minimize disparate mistreatment, is added to a logistic regression model. Finally,~\cite{krasanakis2018adaptive} assume that the ground truth is biased and propose to estimate the true unbiased labels by iteratively re-adjusting the instance weights while minimizing disparate mistreatment.

\noindent \textbf{Post-processing approaches.} 
Post-processing approaches can be divided into two subcategories: the ones that change the decision boundary of the model (white-box approaches) and the ones that directly change the predictions of the model (black-box approaches). 
In the first category, belongs~\cite{fish2016confidence} which shifts the decision boundary of AdaBoost to minimize statistical parity. An extension of this idea has been applied for non-stationary data~\cite{iosifidis2020fabboo,iosifidis2021online}: the decision boundary is tweaked online to tackle concept drifts and to mitigate discriminatory outcomes as evaluated by statistical parity or equal opportunity.   
In~\cite{DBLP:conf/sdm/PedreschiRT09} the authors alter the confidence of CPAR classification rules and in~\cite{calders2010three} the authors change Na\"ive Bayes probabilities considering fairness. 
Black-box approaches have no access to the learner but only to its decisions. For example,~\cite{hardt2016equality} set thresholds to the model predictions to achieve the same error rates for protected and non-protected groups. An extension of this work~\cite{pleiss2017fairness} analyzes how to obtain calibrated classifiers with the same error rates among different groups.


\section{Basic concepts and definitions}%
\label{sec:preliminaries}
We assume a dataset $D$ of $N$ i.i.d. samples drawn from the joint distribution $P(F,S,y)$: 
$S$ denotes the set of \emph{protected attributes} such as gender or race, $F$ denotes other \emph{non-protected attributes} and $y$ is the class label.
For simplicity, we consider that the classification problem is binary, that is, $y \in \{+, -\}$ and that there exists a single protected attribute $S$, also binary: $S \in \{s, \bar{s}\}$ with $s$ and $\bar{s}$ denoting the \emph{protected} and \emph{non-protected group}, respectively.
We use the notation $s_+$ ($s_{-}$), $\bar{s}_+$ ($\bar{s}_{-}$) to denote the protected and non-protected groups for the positive (negative, respectively) class.

\noindent\textbf{Fairness measures:} As already discussed in Section~\ref{sec:related}, we adopt three different parity-based fairness measures: \emph{\underline{St}atistical \underline{Parity}} (shortly $St.Parity$), \emph{\underline{Eq}ual \underline{Op}portunity} (shortly $Eq.Op.$), and \emph{\underline{D}isparate \emph{\underline{M}}istreatment} (shortly $D.M.$).

$St.Parity$ measures the difference in the positive prediction rates between the protected and non-protected groups and is defined as:
\begin{equation} 
\label{eq:statistical_parity}
St.Parity = P( \hat{y} = + | \bar{s}) - P(\hat{y} = + | s) 
\end{equation}
where $\hat{y}$ denotes the prediction. $St.Parity$ takes values in the [0-1] range with 0 standing for no discrimination and 1 for maximum/worse discrimination. 

$Eq.Op.$ measures the difference of the false negative prediction rates between the protected and non-protected groups. It extends $St.Parity$ by considering not only the predictions but also the ground truth labels. Formally:
\begin{equation} 
\label{eq:eqop}
Eq.Op. = P(y\neq \hat{y} | \bar{s}_+) - P(y\neq \hat{y} | s_+) = \delta FNR
\end{equation}
We denote by $\delta FNR$ the difference in false negative rates between the protected and non-protected groups. 
$Eq.Op.$ also takes values in the [0-1] range. 

Finally, $D.M.$ measures the difference in prediction errors between  the protected and non-protected 
groups. It extends $Eq.Op.$ by considering not only the positive misclassifications but also the negative misclassifications. 
Similar to $\delta FNR$ above, we denote by $\delta FPR$ the difference in false positive rates  between the protected and non-protected groups, namely:
\begin{equation} 
\label{eq:FPR_FNR}
\begin{aligned}
\delta FPR = P(y\neq \hat{y} | \bar{s}_-) - P(y\neq \hat{y} | s_-) 
\end{aligned}
\end{equation}

Then, $D.M.$ is defined as follows:
\begin{equation}
D.M. = |\delta FPR |+ |\delta FNR|
\label{eq:EqOdds}
\end{equation}
$D.M.$ values lie in the [0-2] range as $|\delta FPR|$, $|\delta FNR|$ lie in the  [0-1] range.


The goal of fairness-aware classification is to find a mapping $f:(F,S) \rightarrow y$ that achieves good predictive- and fairness-performance (in our case, the latter is evaluated in terms of $St.Parity$, $Eq.Op.$, or $D.M.$).

\noindent\textbf{Predictive performance:}
In the context of fairness-aware learning, predictive performance is typically assessed via \emph{error rate}, e.g.,~\cite{krasanakis2018adaptive,zafar2017fairness,calders2009building,calmon2017optimized,kamiran2012data,hardt2016equality,fish2016confidence,kamiran2009classifying,kamiran2010discrimination}, defined as:
\begin{equation}\label{eq:standard_error}
ER=\frac{FN+FP}{TP+TN+FN+FP}
\end{equation}
where $FP, FN, TP, TN$ are the false positive, false negative, true positive, true negative cases, respectively.
However, optimizing for the error rate is problematic in cases of \emph{class imbalance}. A possible outcome in such a case is that the classifier will misclassify most (in the extreme case all) of the minority instances while correctly classifying the majority. In this scenario, the error rate ($ER$) will still be low despite the poor performance in the minority class, and w.r.t fairness, a classifier might still be fair, e.g., $D.M. \approx 0$, as the difference between the FPRs, FNRs for each group will be low (c.f., Equations~\eqref{eq:FPR_FNR},\eqref{eq:EqOdds}).
However, in such a case, the non-discriminatory behavior of the classifier would be achieved by drastically reducing the correct predictions for the minority class, with the extreme case of misclassifying all minority instances. 
As we will see in the experiments section, many of the datasets in this domain exhibit high class
imbalance (c.f., Table~\ref{tbl:datasets}) and therefore, tackling fairness requires
also tackling imbalance.

Our goal in this work is to minimize discriminatory outcomes (as measured by $St.Parity$, $Eq.Op.$, or $D.M.$) while maintaining good predictive performance for \emph{both} classes.
To this end, we propose (c.f., Section~\ref{sec:AdaFair_tuning}) to replace the error rate (which is not a good performance indicator in case of class-imbalance, as we also show experimentally in Section~\ref{sec:evaluation}) with the \emph{balanced error rate (BER)} which is the average of the errors on each class~\cite{brodersen2010balanced}:
\begin{equation}\label{eq:balanced_error}
BER = 1 - \frac{1}{2}\cdot(\frac{TP}{TP + FN} + \frac{TN}{TN + FP})
= 1 - \frac{1}{2}\cdot(TPR + TNR)
\end{equation}
where $TPR$ ($TNR$) is the true positive rate (true negative rate, respectively).

\noindent\textbf{AdaBoost:} 
Our classification model is based on AdaBoost~\cite{schapire1999brief}, an ensemble technique that combines multiple weak learners to create a strong learner. The weak learners are trained sequentially, each trying to correct the errors of its predecessor by adjusting the instance weights accordingly. AdaBoost takes as input the number of boosting rounds $T$ and trains an ensemble of weak learners $h_{[1-T]}$, by training each weak learner $h_{j+1},j:1-T$ with the updated weight distribution from the previous round $j$. The weight distribution is updated based on the errors of the previous weak learner as follows: 
\begin{equation}\label{eq:Adaboost_updated}
D_i^{j+1} \leftarrow \frac{1}{Z_j} D_i^j \cdot\exp{(-\alpha_j \cdot y_i\cdot h_j(x_i))}
\end{equation}
where $D_i^j$ is the weight of instance $i$ in the current boosting round $j$, $Z_j$ is the normalization factor and $\alpha_j$ is the weight of the weak learner $h_j$ defined as: $\alpha_j = \frac{1 - err_j}{err_j}$ (where $err_j$ is the error rate of $h_j$ on the training set). 

The final ensemble uses a weighted majority schema at the prediction phase, i.e., for an instance $x$ the prediction is derived as: $H(x) = \sum_{j=1}^T \alpha_jh_j(x)$.

\section{Parity-Based Fairness-Aware Cumulative Boosting}%
\label{sec:method}
AdaBoost and boosting in general divide the complex learning problem into lower complexity sub-problems and then combine their solutions (sub-models) into an overall (global) model. Intuitively, such a technique is highly promising for fairness-aware learning, as it is easier to tackle the fairness problem in the simpler sub-models than in the complex global model. However, adopting AdaBoost for fairness requires careful interventions in the data distribution that take into account both predictive and fairness-related performance (Section~\ref{sec:method}).

We tailor AdaBoost to fairness by adjusting the re-weighting process, which traditionally focuses on the misclassifications of the previous weak learner $h_j$ for training the next weak learner $h_{j+1}$
(c.f., Equation~\eqref{eq:Adaboost_updated}). 
In particular, we directly consider the fairness behaviour of the model in the weighting process by introducing fairness-related costs. Moreover, for the fairness-related costs, we don't rely only on the fairness behavior of the previous single weak learner $h_j$, but on the fairness behavior of the \emph{partial ensemble} $H_{1:j}$. 
By taking into account the ``history" of the weak learners for fairness-related interventions, we aim to achieve smoother interventions based on the \emph{cumulative} performance of the model rather than on the varying performances of individual weak learners.

We first introduce the cumulative fairness costs based on the adopted fairness notions and their corresponding cumulative versions, namely cumulative $St.Parity$ ( Section~\ref{sec:cumulativeSP}), cumulative $Eq.Op.$ (Section~\ref{sec:cumulativeEQOP}) and cumulative $D.M.$ (Section~\ref{sec:cumulativeEO}). 
The fairness-aware interventions in the distribution re-weighting process are described in Section~\ref{sec:AdaFair_reweighting}. A theoretical analysis of the training error is provided in Section~\ref{sec:AdaFair_training_error}. Finally, we optimize the number of weak learners in the final ensemble based on the balanced error rate and thus directly considering class imbalance in the best model selection (Section~\ref{sec:AdaFair_tuning}). 

\subsection{Cumulative fairness notions and fairness costs}
\label{sec:cumulativeFairness}
Let $j \in [1,T]$ be the current boosting round
and $H_{1:j}=\{h_1, \cdots, h_j\}$ the sequence of weak learners up to $j$, i.e., the partial ensemble. For an instance $x$, the partial ensemble decides according to: $H_{1:j}(x) = sign(\sum_{i=1}^j \alpha_ih_i(x))$.
We define the fairness-related costs based on the cumulative behavior of the model, i.e., based on the partial ensemble $H_{1:j}$.
In the following subsections, for each fairness notion ($St.Parity$, $Eq.Op.$, $D.M.$), we first define their cumulative counterparts and then the fairness costs.

\subsubsection{Cumulative Statistical Parity}
\label{sec:cumulativeSP}

The \emph{cumulative statistical parity} in round $j$, denoted by $\delta SP_{1:j}$, evaluates the 
parity in the positive predictions of the partial ensemble $H_{1:j}$ between the protected and non-protected groups. Formally:
\begin{equation}\footnotesize
\begin{aligned}
\label{eq:cumulsp}
\delta SP_{1:j} = \frac{\sum\limits_{i,x_i \in \bar{s}} 1 \cdot\mathbb{I}\left[\sum\limits_{k=1}^j \alpha_kh_k(x_i) = +\right]}{|\bar{s}|} - \frac{\sum\limits_{i,x_i \in s} 1 \cdot\mathbb{I}\left[\sum\limits_{k=1}^j \alpha_kh_k(x_i) = + \right]}{|s|}
\end{aligned}
\end{equation}\normalsize
where the function $\mathbb{I}(\cdot)$ returns 1 iff the expression within is true, otherwise 0. 

If there is no parity, i.e., $\delta SP \neq 0$, we change the weights of the training instances so that the discriminated group is boosted extra in the next round $j+1$. Note that vanilla AdaBoost already boosts the misclassified instances for the next round. Our weighting, therefore, aims at achieving parity between the protected and non-protected groups. To this end, we assign fairness-related costs to the discriminated group. More formally, the fairness-related cost $u_i^j$, for an instance $x_i$ in the boosting round $j$ is computed as follows:
\begin{equation}\footnotesize
\begin{aligned}
\label{eq:fairnessCostsSP}
u_i^j = 
 \begin{cases}
|\delta SP_{1:j}|, & if~\mathbb{I}((y_i \neq h_j(x_i)) \land |\delta SP_{1:j}| >\epsilon), x_i \in s, sign(\delta SP_{1:j}) = +\\
|\delta SP_{1:j}|, & if~\mathbb{I}((y_i \neq h_j(x_i)) \land |\delta SP_{1:j}| >\epsilon), x_i \in \bar{s},sign(\delta SP_{1:j}) = - \\ 
0, & otherwise\\
 \end{cases}
\end{aligned}
\end{equation}\normalsize
where $u_i^j \in [0,1]$, $sign()$ is the sign function, and parameter $\epsilon \in \mathbf{R^+}$ reflects the tolerance to unfairness and is typically set to zero or to a very small value\footnote{The notions $u_i^j$ and $\epsilon$ will bear the same meaning for the rest of the section.}.
The signs (+/-) of $\delta SP_{1:j}$ denote which group is discriminated and should be boosted w.r.t. fairness, while $\epsilon$ is a condition for the necessity of fairness-related costs in the upcoming round $j+1$. 
For example, if in round $j$ the group $s$ is discriminated, which means $\delta SP_{1:j} > \epsilon$, then misclassified instances $x_i$ in this group will receive fairness-related costs for the next round. 
Note that all misclassified instances of the discriminated group will receive the same cost in the boosting round $j$. However, the costs are dynamically estimated in each round.


\subsubsection{Cumulative Equal Opportunity}
\label{sec:cumulativeEQOP}
The \emph{cumulative equal opportunity} in round $j$ evaluates the parity in the 
false negative prediction rates
of the partial ensemble $H_{1:j}$ between the protected and non-protected groups. Formally:
\begin{equation}\footnotesize
\begin{aligned}
\label{eq:cumuleqop}
\delta FNR_{1:j} = \frac{\sum\limits_{i,x_i \in \bar{s}_+} 1 \cdot\mathbb{I}\left[\sum\limits_{k=1}^j \alpha_kh_k(x_i) \neq y_i\right]}{|\bar{s}_+|} - \frac{\sum\limits_{i,x_i \in s_+} 1 \cdot\mathbb{I}\left[\sum\limits_{k=1}^j \alpha_kh_k(x_i) \neq y_i\right]}{|s_+|}
\end{aligned}
\end{equation}\normalsize

Similar to cumulative statistical parity, cumulative equal opportunity assigns fairness-related costs in each round to instances that belong to an unfairly treated group.
For an instance $x_i$ in the boosting round $j$, the fairness-related cost $u_i^j$ is computed as follows: 
\begin{equation}\footnotesize
\begin{aligned}
\label{eq:fairnessCostsEQOP}
u_i^j = 
 \begin{cases}
 |\delta FNR_{1:j}|, & if~\mathbb{I}((y_i \neq h_j(x_i))\land |\delta FNR_{1:j}| >\epsilon), x_i \in s_+, sign(\delta FNR_{1:j}) = +\\
 |\delta FNR_{1:j}|, & if~\mathbb{I}((y_i \neq h_j(x_i))\land |\delta FNR_{1:j}| >\epsilon), x_i \in \bar{s}_+,sign(\delta FNR_{1:j}) = - \\
 0, & otherwise\\
 \end{cases}
\end{aligned}
\end{equation}\normalsize
For example, if the group $s_+$ is discriminated in round $j$, which means $\delta FNR_{1:j} > \epsilon$, then the misclassified instances in this group will be boosted in the next round based on Equation~\ref{eq:fairnessCostsEQOP}.

\subsubsection{Cumulative Disparate Mistreatment}
\label{sec:cumulativeEO}
\emph{Cumulative disparate mistreatment} extends cumulative equal opportunity by considering parity among protected and non-protected groups for both the positive and negative classes. 
We define it in terms of $\delta FPR$, $\delta FNR$ of the partial ensemble $H_{1:j}$. Similar to $\delta FNR_{1:j}$ (c.f., Equation~\eqref{eq:cumuleqop}), $\delta FPR_{1:j}$ is defined as:
\begin{equation}\footnotesize
\begin{aligned}
\label{eq:cumulFairEO}
\delta FPR_{1:j} = \frac{\sum\limits_{i,x_i \in \bar{s}_-} 1 \cdot\mathbb{I}\left[\sum\limits_{k=1}^j \alpha_kh_k(x_i) \neq y_i\right]}{|\bar{s}_-|} - \frac{\sum\limits_{i,x_i \in s_-} 1 \cdot\mathbb{I}\left[\sum\limits_{k=1}^j \alpha_kh_k(x_i) \neq y_i\right]}{|s_-|}
\end{aligned}
\end{equation}\normalsize

In the boosting round j, miss-classified  instances $x_i$ of the discriminated group are boosted extra based on fairness-related costs $u_i^j$ defined as:
\begin{equation}\footnotesize
\begin{aligned}
\label{eq:fairnessCostsEO}
u_i^j = 
 \begin{cases}
 |\delta FNR_{1:j}|, & if~\mathbb{I}((y_i \neq h_j(x_i))\land |\delta FNR_{1:j}| >\epsilon), x_i \in s_+, sign(\delta FNR_{1:j}) = + \\
 |\delta FNR_{1:j}|, & if~\mathbb{I}((y_i \neq h_j(x_i))\land |\delta FNR_{1:j}| >\epsilon), x_i \in \bar{s}_+,sign(\delta FNR_{1:j}) = - \\
 |\delta FPR_{1:j}|, & if~\mathbb{I}((y_i \neq h_j(x_i))\land |\delta FPR_{1:j}| >\epsilon), x_i \in s_-, sign(\delta FPR_{1:j}) = +\\
 |\delta FPR_{1:j}|, & if~\mathbb{I}((y_i \neq h_j(x_i))\land |\delta FPR_{1:j}| >\epsilon), x_i \in \bar{s}_-,sign(\delta FPR_{1:j}) = - \\ 0, & otherwise\\
 \end{cases}
\end{aligned}
\end{equation}\normalsize

Again, only misclassified instances are susceptible to receive the fairness-related cost (the order of the cost assignments inside the equation does not matter). Similar to the other measures, all miss-classified instances of the discriminated group will receive the same cost in a given boosting round $j$. Still, the costs might change across the rounds as they depend on the partial ensemble.

\subsection{The AdaFair Algorithm}
AdaFair is a sequential ensemble that extends Adaboost for fairness-aware learning under class-imbalance. 
The algorithm consists of two steps. The first step (Section~\ref{sec:AdaFair_reweighting}) is the in-training phase of the ensemble based on the selected cumulative fairness notion and its fairness-related costs (Section~\ref{sec:cumulativeFairness}). The second step (Section~\ref{sec:AdaFair_tuning}) is the post-processing phase, in which the algorithm selects the optimal partial ensemble that offers the minimum weighted summed loss of (balanced) predictive performance and fairness.

\subsubsection{In-processing Distribution Update}
\label{sec:AdaFair_reweighting}
The main difference to vanilla AdaBoost is the weight distribution update formula (c.f., Equation~\eqref{eq:Adaboost_updated}) which now also considers the fairness-related costs $u_i^j$.
In particular, the data distribution is updated as follows:
\begin{equation}
\label{eq:distribution}
D^{j+1}(i) = \frac{D^j(i)C^{j}_i\exp{(-\alpha_j y_i h_j(x_i)})}{Z_j}
\end{equation}
For convenience, we use $C_i^j = (1+u_i^j)$ instead of $u_i^j$. For example, if $u_i^j = 0$ then $C_i^j=1$ and the instance is not affected. In addition, $u_i^j \in [0,1]$; therefore, the fairness-related cost would degrade the instance's weight instead of boosting it. 

The normalization factor $Z_j$ ensures $D^{j+1}$ is a probability distribution:
\begin{equation}
    \begin{aligned}\label{eq: normalization}
        Z_j = \sum\limits_{i=1}^N D^j(i)C^j_i\exp{(-\alpha_jy_ih_j(x_i))}
    \end{aligned}
\end{equation}

The ensemble training is shown in  Algorithm~\ref{alg:method}. 
Instance weights $D_i^1$ and fairness-related costs $u_i^1$ are initialized (line 1). In each boosting round $j:1-T$ (lines 2--12), a weak learner $h_j$ is trained upon the current weight distribution $D^j$ (line instance 3) and the $\alpha_j$, and fairness-related costs $u_i^j$ are computed (lines 4, and 5, respectively). The new instance weights are estimated (line 8). 

After the in-training learning phase, AdaFair~selects, post-training, the best sequence of weak learners (line 13, Algorithm~\ref{alg:method}), which achieves the best trade-off between balanced and standard error rate as specified by a user-defined parameter $c$ (Equation~\eqref{eq:argmin}). The post-training phase directly tackles imbalance and is discussed hereafter (Section~\ref{sec:AdaFair_tuning}).


\begin{algorithm}
 \SetKwInput{KwData}{Input}
	\KwData{$D = (x_i,y_i)_1^N, T, \epsilon, c$ }	
 	\KwResult{Ensemble $H$}
 Initialize $D_i^1 = 1/N$, $C_i^1=1$, and $u_i^1=0$, for $i = 1, 2, \dots, N$\;
 \For{j=1 \KwTo $T$} {
	 Train a classifier $h_j$ to the training data using weights $D^j$\;
 Compute the weight $\alpha_j$ (Equation~\eqref{eq:adafair_at_final})\;
 Compute fairness-related costs $u_i^j$ based on a given fairness notion (Equations~\eqref{eq:cumulsp},~\eqref{eq:cumuleqop},~\eqref{eq:cumulFairEO})\;
 $C_i^j = (1 + u_i^j)$\;
 
 Update the distribution as: \\$D_i^{j+1} \leftarrow \frac{1}{Z_j} D_i^j \cdot C_i^j \cdot \exp{(-\alpha_j \cdot y_i\cdot h_j(x_i))}$ \\// $Z_j$ is normalization factor\; 
 \uIf{ Termination condition does not hold (Equation~\eqref{eq:condition})}{
 break\;
 }
 }
 Return best weak learner sequence (Equation~\eqref{eq:argmin}) using parameter $c$ 
 \caption{AdaFair algorithm}
 \label{alg:method}
\end{algorithm}

\subsubsection{Post-processing Model Selection based on Balanced Performance}
\label{sec:AdaFair_tuning}
The number of weak learners $T$ is provided as input to AdaFair, similarly to AdaBoost. 
We propose to refine the model by  
finding the best (sub)sequence of weak learners $1 \cdots \theta, \theta \leq T$ that achieves good performance for both classes and is fair according to the chosen fairness measure. To this end, we propose to optimize for the balanced error rate $BER$ (Equation~\eqref{eq:balanced_error}) instead of the standard error rate $ER$ (Equation~\eqref{eq:standard_error}).
In case of balanced data, BER corresponds to ER. To allow for different combinations of ER and BER in the $\theta$ computation, we consider both ER and BER in the objective function as follows: 
\begin{equation}\label{eq:argmin}
\argminC_\theta ~ (c\cdot BER_\theta + (1-c)\cdot ER_\theta + F.M._\theta)
\end{equation}
where $F.M.$ can be one of the aforementioned cumulative fairness measures (Equations~\eqref{eq:cumulsp},~\eqref{eq:cumuleqop},~\eqref{eq:cumulFairEO}). The parameter $c$ controls the impact of BER and ER in the computation. The selection of $\theta$ is performed based on a validation set (more details on the validation set in Section~\ref{sec:parameter_selection}).
A detailed evaluation of parameter's $c$ impact in the performance of AdaFair is presented in Section~\ref{sec:exp_parameter_c}.

Our approach \emph{directly} tackles class-imbalance in the post-processing phase by selecting the best sequence of weak learners according to Equation~\eqref{eq:argmin}. 
We have also investigated the in-training mitigation of class imbalance. In particular, we transformed AdaFair into a cost-sensitive learner by inserting misclassification costs for each class. However, in our preliminary investigations, the interplay between fairness-related costs and class-related miss-classification costs resulted in an unstable model. We still believe that such an approach is promising
and we plan to pursue this direction in our future work.

\subsection{Bounding the training error}
\label{sec:AdaFair_training_error}

The update of the weight distribution of AdaFair is given in Equation~\eqref{eq:distribution}. Following the same reasoning as in~\cite{sun2007cost}, by unravelling Equation~\eqref{eq:distribution}, we obtain:
\begin{equation} 
\begin{aligned}
\label{eq:unravel_adafair}
D^{t+1}(i) & = D^1(i)\times\frac{C^{1}_i\exp{\left(-\alpha_1y_ih_1(x_i)\right)}}{Z_1}\times\cdots \times \frac{C^{t}_i\exp{\left(-\alpha_ty_ih_t(x_i)\right)}}{Z_t} \\
 & = \frac{D^1(i)\prod\limits_{j=1}^t C^{j}_i\exp{(- \sum\limits_{j=1}^t\alpha_j y_i h_j(x_i) )}}{\prod\limits_{j=1}^t Z_j}
\end{aligned}
\end{equation}

From Equation~\eqref{eq:unravel_adafair}, we get:
\begin{equation}\label{eq:1_N}
    \begin{aligned}
       D^1(i)\exp{(- \sum\limits_{j=1}^t\alpha_j y_i h_j(x_i) )}= \frac{D^{t+1}(i)}{\prod\limits_{j=1}^t C^{j}_i} (\prod\limits_{j=1}^t Z_j)\\
       \implies \frac{1}{N}\exp{(- \sum\limits_{j=1}^t\alpha_j y_i h_j(x_i) )}= \frac{D^{t+1}(i)}{\prod\limits_{j=1}^t C^{j}_i} (\prod\limits_{j=1}^t Z_j)~~~~[\because \forall i, D^1(i)=\frac{1}{N}]
    \end{aligned}
\end{equation}
Then, the training error of the final classifier $H$ is bounded as:
\begin{equation}\label{eq:loss_probability}
    \begin{aligned}
       Pr [H(x_i) \neq y_i]=\frac{1}{N}\sum\limits_{H(x_i) \neq y_i}{\mathbf{1}} \leq \sum\limits_{i=1}^{N}\frac{1}{N}\exp{(- \sum\limits_{j=1}^t\alpha_j y_i h_j(x_i) )}\\
       \implies Pr [H(x_i) \neq y_i] \leq \sum\limits_{i=1}^{N}\frac{D^{t}(i)}{\prod\limits_{j=1}^t C^{j}_i} (\prod\limits_{j=1}^t Z_j) 
    \end{aligned}
\end{equation}
There exists a constant $\gamma$, such that $\forall i, \gamma < \prod\limits_{j=1}^t C^{j}_i$. Then,
\begin{equation}\label{eq:final_bound}
    \begin{aligned}
       Pr [H(x_i) \neq y_i] \leq (\prod\limits_{j=1}^t Z_j)\sum\limits_{i=1}^{N}\frac{D^{t}(i)}{\gamma} \\
        \implies Pr [H(x_i) \neq y_i] \leq \frac{1}{\gamma}\prod\limits_{j=1}^t Z_j~~~~~[\because \sum\limits_{i=1}^N D^{t}(i)=1]
    \end{aligned}
\end{equation}
Since $\gamma$ is a constant, in order to minimize the training error (Equation~\eqref{eq:final_bound}), parameter $Z$ needs to be minimized on each boosting round. Therefore, the objective in each boosting round $t$ is to find $\alpha_t$ that minimizes $Z_t$. According to~\cite{schapire1999improved}, once $y_ih_t(x_i) \in \{-1, 1\}$ holds, the choice of $\alpha_t$ for each $h_t$ can be obtained with the help of the following approximation: 

\begin{equation}
\label{eq:adafair_at}
\begin{split}
 & Z_t=\sum\limits_{i=1}^N D^t(i)C^t_i\exp{\left(- \alpha_t  y_i h(x_i)\right)}\\
 & \leq \sum\limits_{i=1}^N D^t(i)C^t_i\left( \frac{1-y_ih_t(x_i)}{2}e^{\alpha_t} + \frac{1+y_ih_t(x_i)}{2}e^{-\alpha_t} \right)\\
 \end{split}
\end{equation}
To estimate the $\alpha_t$ that minimizes $Z_t$, we need to solve for $\alpha_t$ that minimizes the approximation upper bound in  Equation~\eqref{eq:adafair_at}. 
\begin{equation}
\label{eq:adafair_at_final}
\begin{split}
 & \displaystyle{  \frac{\partial }{\partial \alpha_t} \left( \sum\limits_{i=1}^N D^t(i)C^t_i\left( \frac{1-y_ih_t(x_i)}{2}e^{\alpha_t}\right) + \sum\limits_{i=1}^N D^t(i) C^t_i\left( \frac{1+y_ih_t(x_i)}{2}e^{-\alpha_t} \right) \right) = 0 \Rightarrow}\\
 & e^{\alpha_t}\sum\limits_{i=1}^N D^t(i)C^t_i\left( \frac{1-y_ih_t(x_i)}{2}\right) = e^{-\alpha_t}\sum\limits_{i=1}^N D^t(i)C^t_i\left( \frac{1+y_ih_t(x_i)}{2}\right)\Rightarrow\\
& \alpha_t = \frac{1}{2}\log\left( \frac{\sum\limits_{i,y_i=h_t(x_i)}^N C_i^tD^t(i)}{\sum\limits_{i,y_i\neq h_t(x_i)}^N C_i^tD^t(i)} \right)
 \end{split}
\end{equation}

To preserve the property of AdaBoost, i.e., to ensure $\alpha_t$ is strictly positive, the following must hold:

\begin{equation}\label{eq:condition}
  \sum\limits_{i,y_i=h(x_i)}C^t_i D^t(i) > \sum\limits_{i,y_i\neq h(x_i)} C^t_i D^t(i) 
\end{equation}

\section{Evaluation}%
\label{sec:evaluation}
We evaluate the predictive performance and fairness behavior of AdaFair vs other related approaches (Sections~\ref{sec:exp_comparison_statistical_parity},~\ref{sec:exp_comparison_equal_opportunity} and~\ref{sec:exp_comparison}).
Regarding predictive performance, we report on both accuracy (Equation~\eqref{eq:standard_error}) and balanced accuracy (Equation~\eqref{eq:balanced_error}), whereas for fairness we report on statistical parity (S.P.), equal opportunity (Eq.Op.) and disparate mistreatment (D.M.) (c.f., Section~\ref{sec:preliminaries}).

Another goal of our experiments is to understand the behaviour of AdaFair. To this end, we investigate the effect of cumulative vs non-cumulative fairness (Section~\ref{sec:single_vs_accum}) and the impact of adopting balanced error rate vs error rate (Section~\ref{sec:exp_parameter_c}) for the post-processing model selection. 
We provide the details on the datasets, baselines, parameter selection and evaluation in Section~\ref{sec:expsetup}.

\subsection{Experimental setup}
\label{sec:expsetup}

\subsubsection{Datasets}%
\label{sec:data}
We evaluate our approach on four real-world datasets whose characteristics are summarized in Table~\ref{tbl:datasets}. They comprise a suitable benchmark due to their diverse characteristics, namely cardinality, dimensionality and class imbalance.

\begin{table}[htbp]
\begin{adjustbox}{width=\columnwidth,center}
\begin{tabular}{lcccc}
\hline
 & Adult census~\cite{dua2017}  & Bank~\cite{dua2017} & Compass~\cite{larson2016we} & KDD census~\cite{dua2017} \\ \hline
\#Instances & 45,175 & 40,004 & 5,278 & 299,285 \\
\#Attributes & 14 & 16 & 9 & 41 \\
Sen.Attr. & Gender & Marit. Status & Gender &  Gender \\
Prot.Group (s) & Female & Married & Female & Female \\
Class ratio ($+$:$-$) & 1:3.03 & 1:7.57 & 1:1.12 & 1:15.11 \\
Positive class & \textit{\textgreater{}50K}  & \textit{subscription} & \textit{recidivism} & \textit{\textgreater{}50K}  \\ \hline
\end{tabular}
\end{adjustbox}
\caption{An overview of the datasets used in our experiments.}
\label{tbl:datasets}
\end{table}

\noindent \textbf{Adult census}~\cite{dua2017} dataset contains demographic data from the U.S. The task is to predict whether the annual income of a person will exceed 50K dollars. 
The protected attribute is $S=Gender$ with $s=female$ being the protected group; the positive class is people receiving more than 50K. 
We remove duplicate instances and instances containing missing values. The positive to negative class ratio is ~1:3 (exact ratio 24\%:76\%).

\noindent \textbf{Bank} dataset~\cite{dua2017} is related to direct marketing campaigns of a Portuguese banking institution. The task is to determine if a person will subscribe to the product (bank term deposit). As positive class we consider people who subscribed to a term deposit. We consider as $S=marital~status$ with $s=married$ being the protected group. The dataset suffers from severe class imbalance, with a positive to negative ratio of ~1:8 (exact ratio 11\%:89\%).

\noindent \textbf{Compass} dataset~\cite{larson2016we} stores record about prisoners in Broward County. The task is to predict  \textit{(recidivism)}, namely if a person will be re-arrested within two years. We consider \textit{recidivism} as the positive class and $S=Gender$ with $s=female$ as the protected group. For this dataset, we followed the pre-processing steps of~\cite{zafar2017fairness}. The dataset with a positive to negative ratio of $46\%:54\%$ is almost balanced.

\noindent \textbf{KDD census}~\cite{dua2017} has the same prediction task as the Adult census dataset. However, in KDD census ``the class labels were drawn from the total person income field rather than the adjusted gross income''~\cite{dua2017}.
\color{black}

\subsubsection{Baselines}%
\label{sec:baselines}
We evaluate AdaFair against state-of-the-art methods for each fairness measure. 

\noindent For \emph{statistical parity}, we employ \textbf{AdaBoost SDB}~\cite{fish2016confidence}, which trains a vanilla AdaBoost, and afterwards, it tweaks the decision boundary of the induced model (based on a validation set) to mitigate discriminatory outcomes.

\noindent For \emph{equal opportunity}, we employ \textbf{FAE}~\cite{iosifidis2019fae}, an ensemble that combines pre- and post-processing steps to mitigate unfair outcomes and to tackle class imbalance. It pre-processes the data by sampling the dataset to generate equi-sized samples w.r.t the protected attribute and assigns them to a set of bags used to induce an ensemble of AdaBoost models. In the post-processing phase, it shifts the decision boundary of the ensemble to account for fairness. 

\noindent For \emph{disparate mistreatment}, we employ the methods by \textbf{Zafar et al.}~\cite{zafar2017fairness} and \textbf{Krasanakis et al.}~\cite{krasanakis2018adaptive}. \textbf{Zafar et al.}~\cite{zafar2017fairness} presents the fairness problem as a set of convex-concave constraints to minimize discriminatory outcomes and solve it using a logistic regression model. \textbf{Krasanakis et al.}~\cite{krasanakis2018adaptive} assume the existence of latent fair classes and propose an iterative training approach towards those classes by altering the instance weights. We have selected these methods as they follow a different line of reasoning, constraints~\cite{zafar2017fairness} vs hidden unbiased labels~\cite{krasanakis2018adaptive}. 

In addition, we compare against two \emph{fairness-agnostic} boosting methods: 
\textbf{vanilla AdaBoost}~\cite{schapire1999brief} and \textbf{SMOTEBoost}~\cite{chawla2003smoteboost}. SMOTEBoost is an extension of AdaBoost for imbalanced data which tackles imbalance by generating in each boosting round new synthetic instances of the minority class using SMOTE~\cite{chawla2002smote}. The goal of employing SMOTEBoost is to see whether the fairness problem can be addressed by only tackling class imbalance. 

Finally, to study the behaviour of AdaFair, we also compare it against a variation \textbf{AdaFair NoCumul} that computes the fairness-related costs in each round based on only the fairness evaluation of the current weak learner $h_j$,  instead of the partial ensemble $H_{1:j}$. 
The goal of this baseline is to clarify the understanding of the impact of the cumulative fairness notion (Section~\ref{sec:single_vs_accum}).



\subsubsection{Parameter selection and evaluation}%
\label{sec:parameter_selection}
We follow the evaluation setup as in~\cite{zafar2017fairness,krasanakis2018adaptive} by splitting each dataset randomly into train $(50\%)$ and test set $(50\%)$ and report on the average of 10 random splits. 
We set $\epsilon=0$ as a threshold for all fairness measures, which means zero tolerance to discrimination. For extracting the validation set (which is used for estimating $\theta$, Section~\ref{sec:AdaFair_tuning}) we perform a stratified split (67\% training and 33\% validation set). Our method is instantiated in each boosting rounds with decision trees of depth 1 (decision stumps) as weak learners. For the total number of boosting rounds, we set $T=200$ (same for the other ensemble approaches: AdaBoost, AdaBoost SDB, FAE, and SMOTEBoost, c.f., Section~\ref{sec:baselines}). 
We analyze the effect of $T$ in Section~\ref{sec:over_the_rounds}. For Krasanakis et al. and Zafar et al. methods, we employ their default (suggested) parameters. For SMOTEBoost, we set $N$ (the number of synthetic instances generated per round) to 2, 100, 100 and 500 for datasets Compass, Adult census, Bank and KDD census, respectively. Furthermore, for experiments in Sections~\ref{sec:exp_comparison},~\ref{sec:over_the_rounds} and \ref{sec:single_vs_accum}, we set parameter $c=1$ (c.f., Equation~\eqref{eq:argmin}), that is the proposed AdaFair optimized for balanced error rate; the effect of $c$ is studied in Section~\ref{sec:exp_parameter_c}.

\subsection{Statistical Parity: Predictive and fairness performance}%
\begin{figure*}[htp!]
 \centering
 \begin{subfigure}[t]{0.40\textwidth}
 \centering
 \includegraphics[width=1.0\columnwidth]{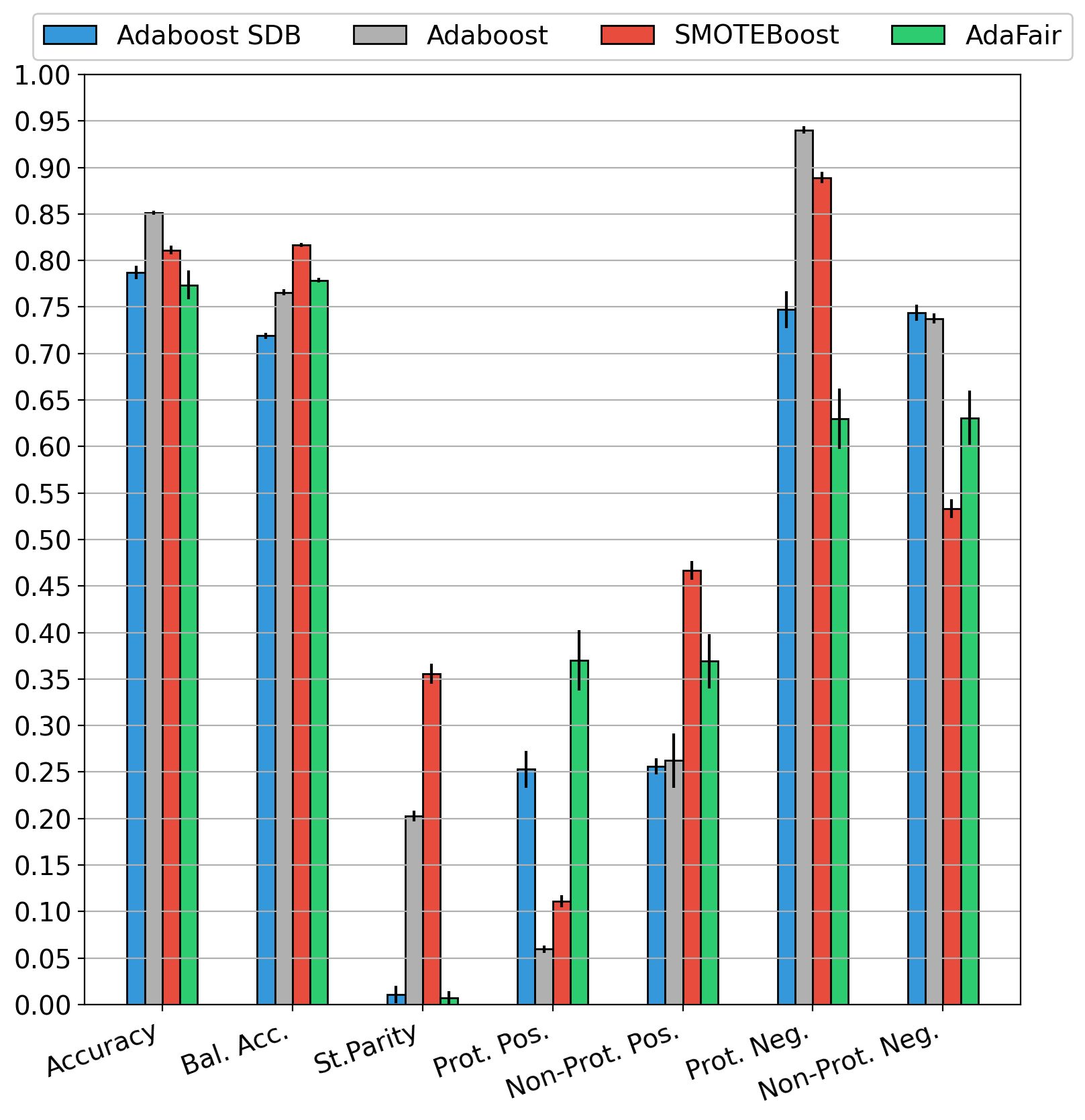}
 \caption{Adult census}
  \label{fig:st_p_adult}
 \end{subfigure}
 \centering
 \begin{subfigure}[t]{0.40\textwidth}
 \centering
 \includegraphics[width=1.0\columnwidth]{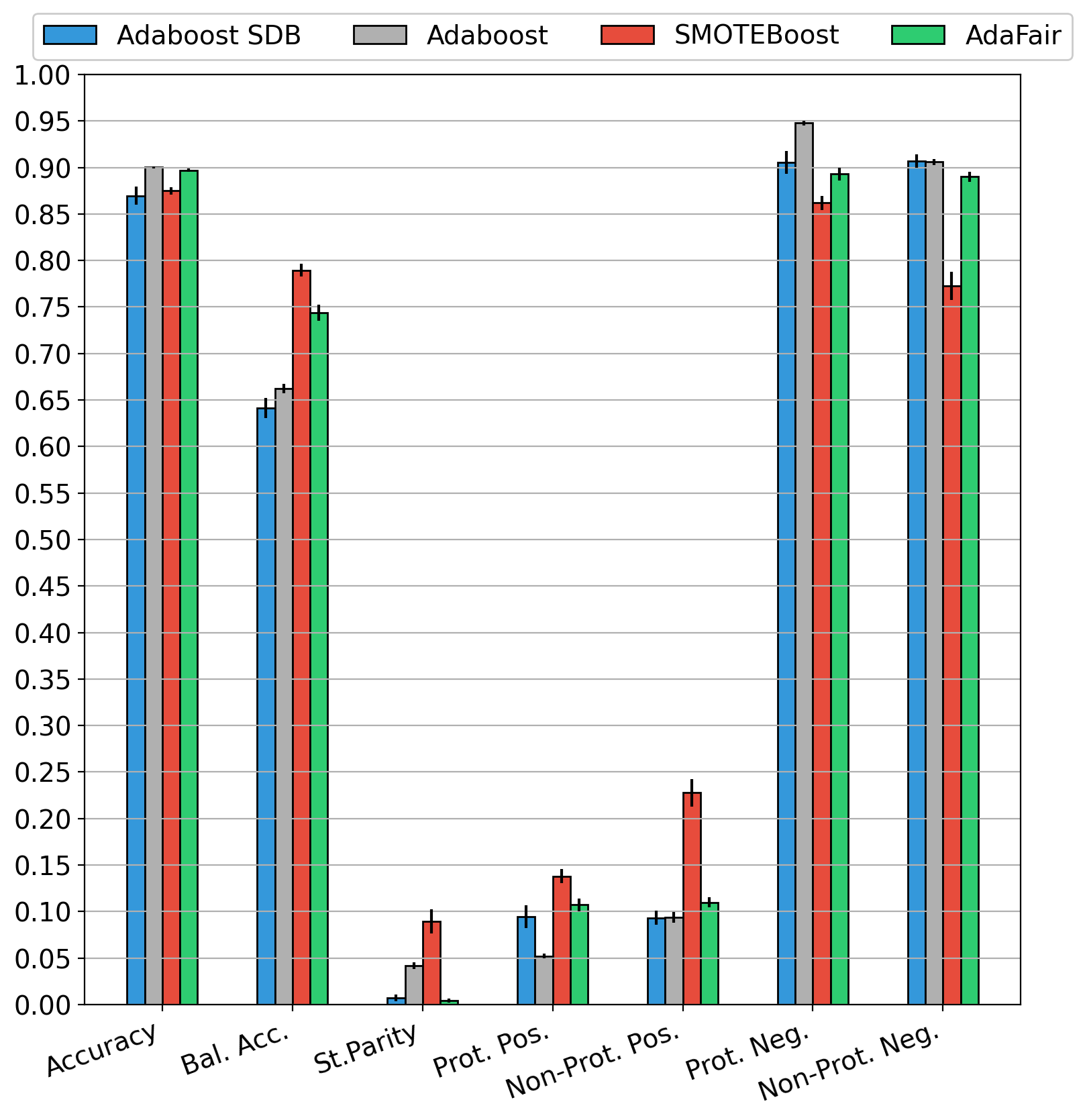}
 \caption{Bank}
   \label{fig:st_p_bank}
 \end{subfigure}
 \begin{subfigure}[t]{0.40\textwidth}
 \centering
 \includegraphics[width=1.0\columnwidth]{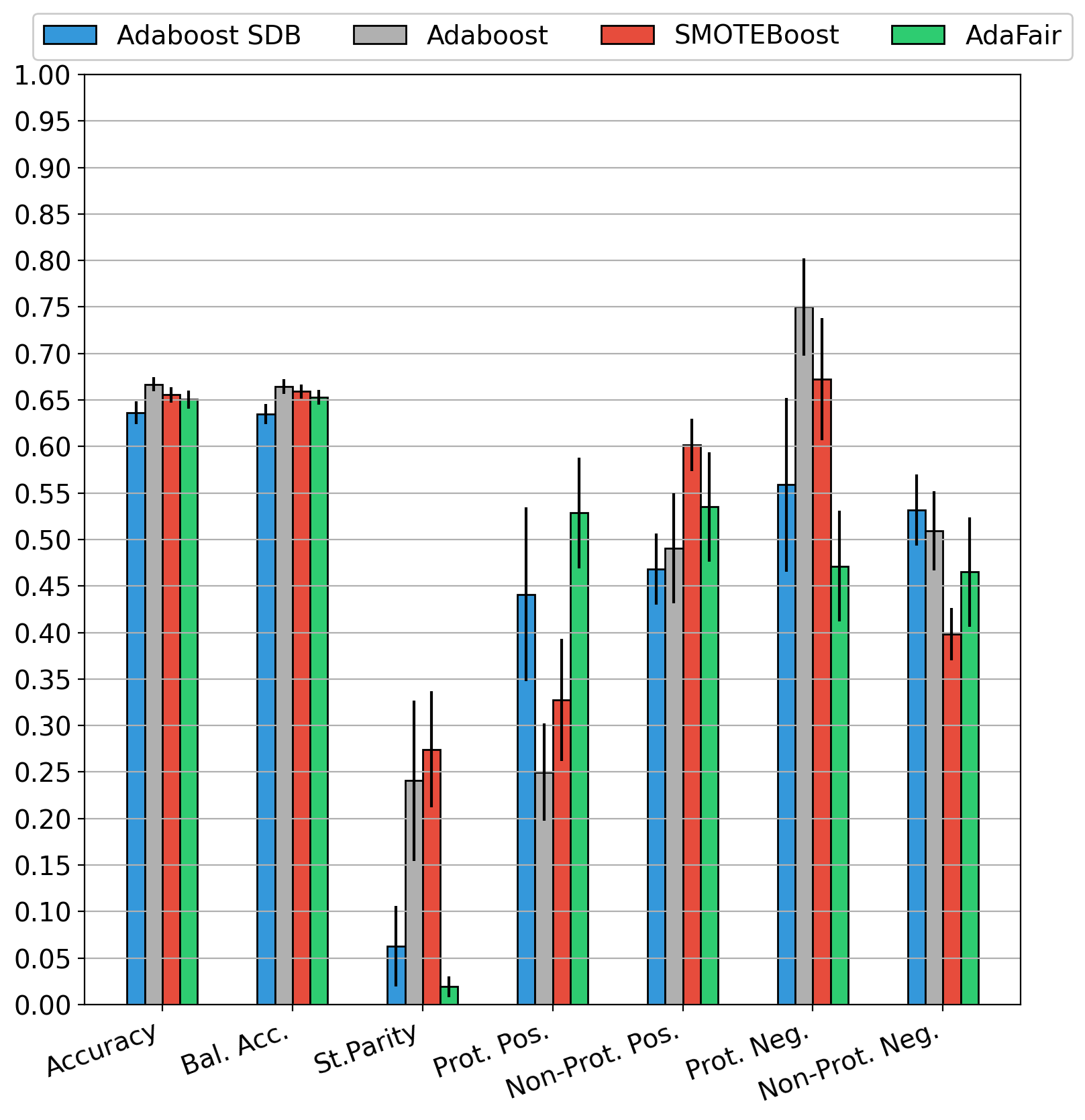}
 \caption{Compass} 
\label{fig:st_p_compass}
\end{subfigure}
 \begin{subfigure}[t]{0.40\textwidth}
 \centering
 \includegraphics[width=1.0\columnwidth]{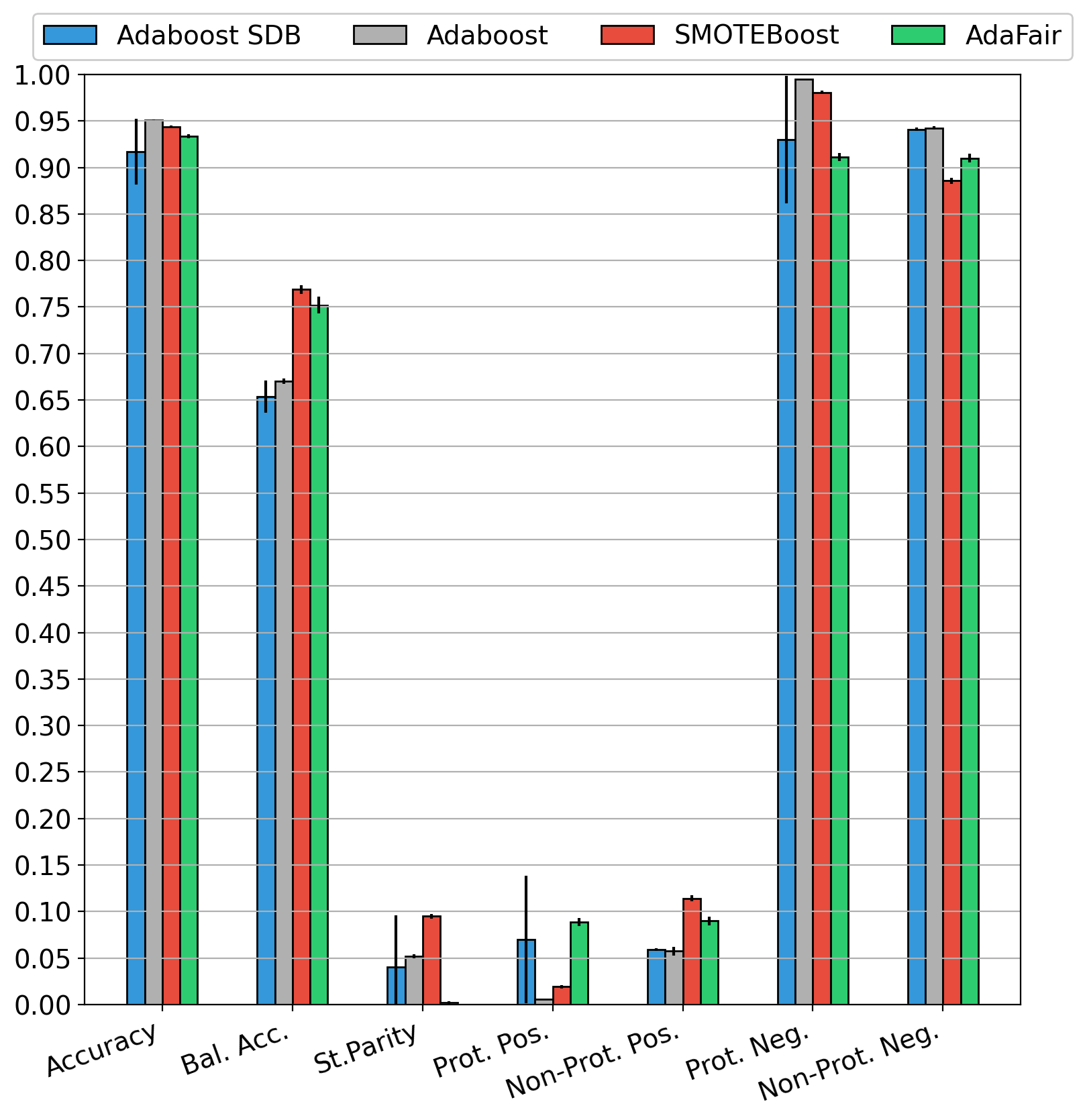}
 \caption{KDD census} 
 \label{fig:st_p_kdd}
 \end{subfigure}
 \caption{Predictive and fairness, based on Statistical Parity, performance - higher values are better; for Statistical Parity, lower values are better.} 
 \label{fig:performance_statistical_parity}
\end{figure*}

\label{sec:exp_comparison_statistical_parity}
In Figure~\ref{fig:performance_statistical_parity} we report on predictive performance (both accuracy and balanced accuracy, Bal.Acc. for short) and on fairness-related performance, namely statistical parity (St. Parity, for short). We also report on the TPR and TNR for both protected and non-protected groups to showcase the fairness and accuracy of the approaches for both groups. 

\noindent\textbf{Adult census:} 
In Figure \ref{fig:st_p_adult}, we show the performance of the different approaches on the Adult census dataset. AdaFair achieves the best (lowest) statistical parity score, closely followed by AdaBoost SDB. This is also reflected in the almost identical percentage of positive predictions for the protected (denoted as Prot. Pos.) and non-protected groups (denoted as Non-prot. Pos.).
In terms of balanced accuracy, we see that AdaFair is only second best to SMOTEBoost by a 3.5\%$\downarrow$ drop (note that SMOTEBoost has the worst discriminatory behaviour). In comparison to AdaBoost SDB, AdaFair produces slightly fairer outcomes and 5\%$\uparrow$ better performance in terms of balanced accuracy. 


\noindent\textbf{Bank:} 
The results on the Bank dataset are shown in Figure \ref{fig:st_p_bank}. AdaFair again achieves the best fairness score, and in balanced accuracy it outperforms AdaBoost SDB by a margin of 10\%$\uparrow$. SMOTEBoost is 5\%$\uparrow$ better than AdaFair and achieves the best balanced accuracy. 
AdaBoost achieves the best performance in terms of accuracy, while all the other methods, including AdaFair, have similar accuracy. The high accuracy of AdaBoost is because it predicts most of the instances as negative (the majority class) and very few instances as positive (the minority) class, which is evident in the last four-bar plots Prot. Pos, Non-prot Pos, Prot. Neg, and Non-Prot Neg. 

\noindent\textbf{Compass:} 
The results on Compass dataset are shown in Figure \ref{fig:st_p_compass}. We observe that AdaFair achieves the best statistical parity score. All the methods have similar performance in balanced accuracy and accuracy since this dataset does not suffer from class imbalance. SMOTEBoost and AdaBoost perform very poorly in terms of statistical parity. AdaFair produces 3.5\%$\downarrow$ fairer outcomes in contrast to AdaBoost SDB, and slightly higher balanced accuracy (1\%$\uparrow$).

\noindent\textbf{KDD census:} 
Figure \ref{fig:st_p_kdd} depicts the results on KDD census income dataset. Once again, we observe that AdaFair achieves the best statistical parity score compared to the baselines (4\%$\downarrow$ lower than AdaBoost SDB). Moreover, our method AdaFair also achieves higher balanced accuracy than AdaBoost (8.5\%$\uparrow$), AdaBoost SDB (10\%$\uparrow$). SMOTEBoost has higher balanced accuracy (2\%$\uparrow$) and high discriminatory outcomes (10\%$\uparrow$) in contrast to AdaFair.

\noindent\textbf{Conclusion:}  
AdaFair performs better than AdaBoost SDB in terms of fairness and predictive performance. However, SMOTEBoost outperforms AdaFair in balance accuracy, mainly due to the nature of the fairness notion. Statistical parity forces AdaFair to shift the decision boundary to achieve parity between the different groups without considering the true label distribution.

\subsection{Equal Opportunity: Predictive and fairness performance}%
\label{sec:exp_comparison_equal_opportunity}

\begin{figure*}[htp!]
 \centering
 \begin{subfigure}[t]{0.40\textwidth}
 \centering
 \includegraphics[width=1.0\columnwidth]{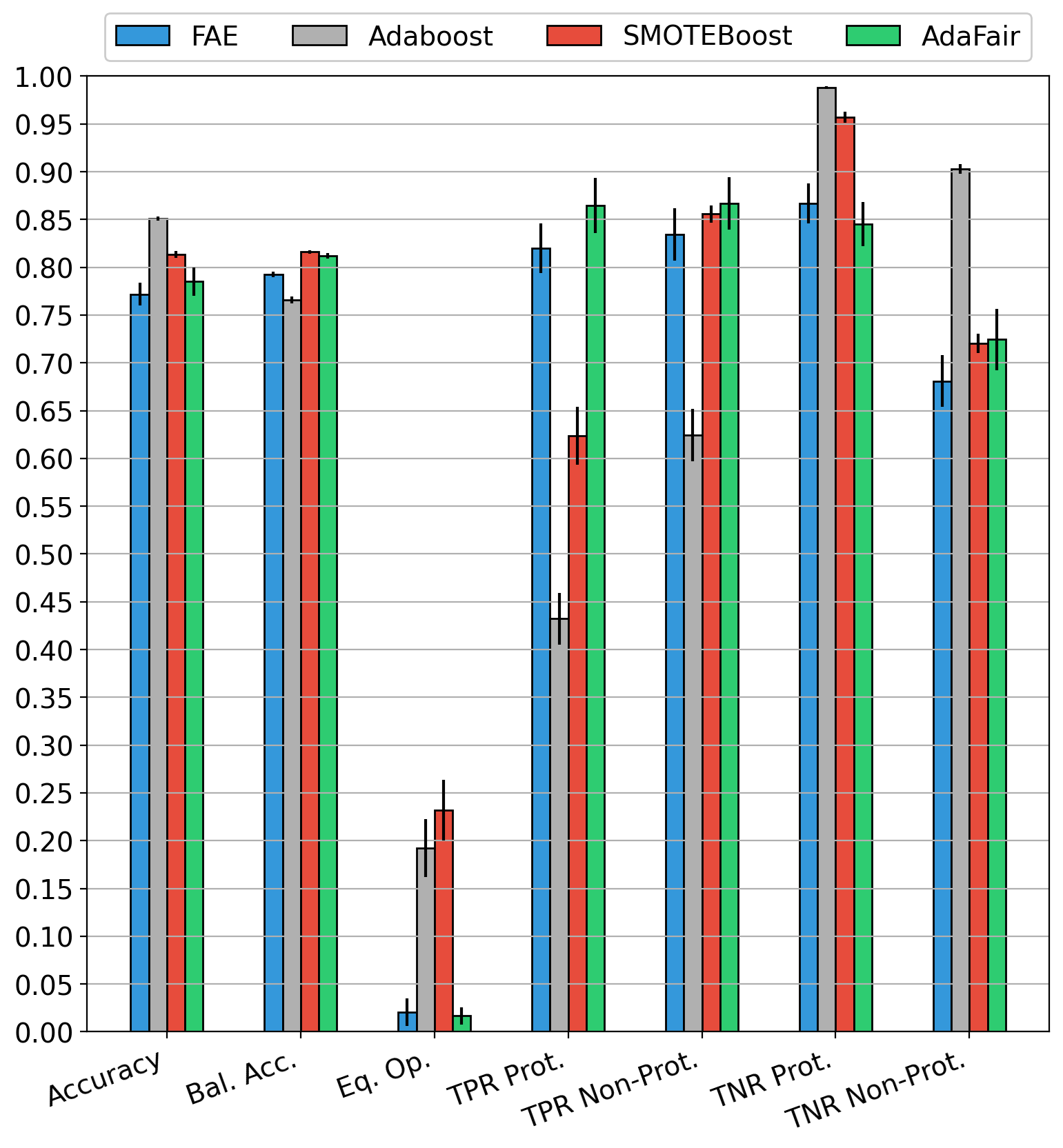}
 \caption{Adult census}
  \label{fig:eq_op_adult}
 \end{subfigure}
 \centering
 \begin{subfigure}[t]{0.40\textwidth}
 \centering
 \includegraphics[width=1.0\columnwidth]{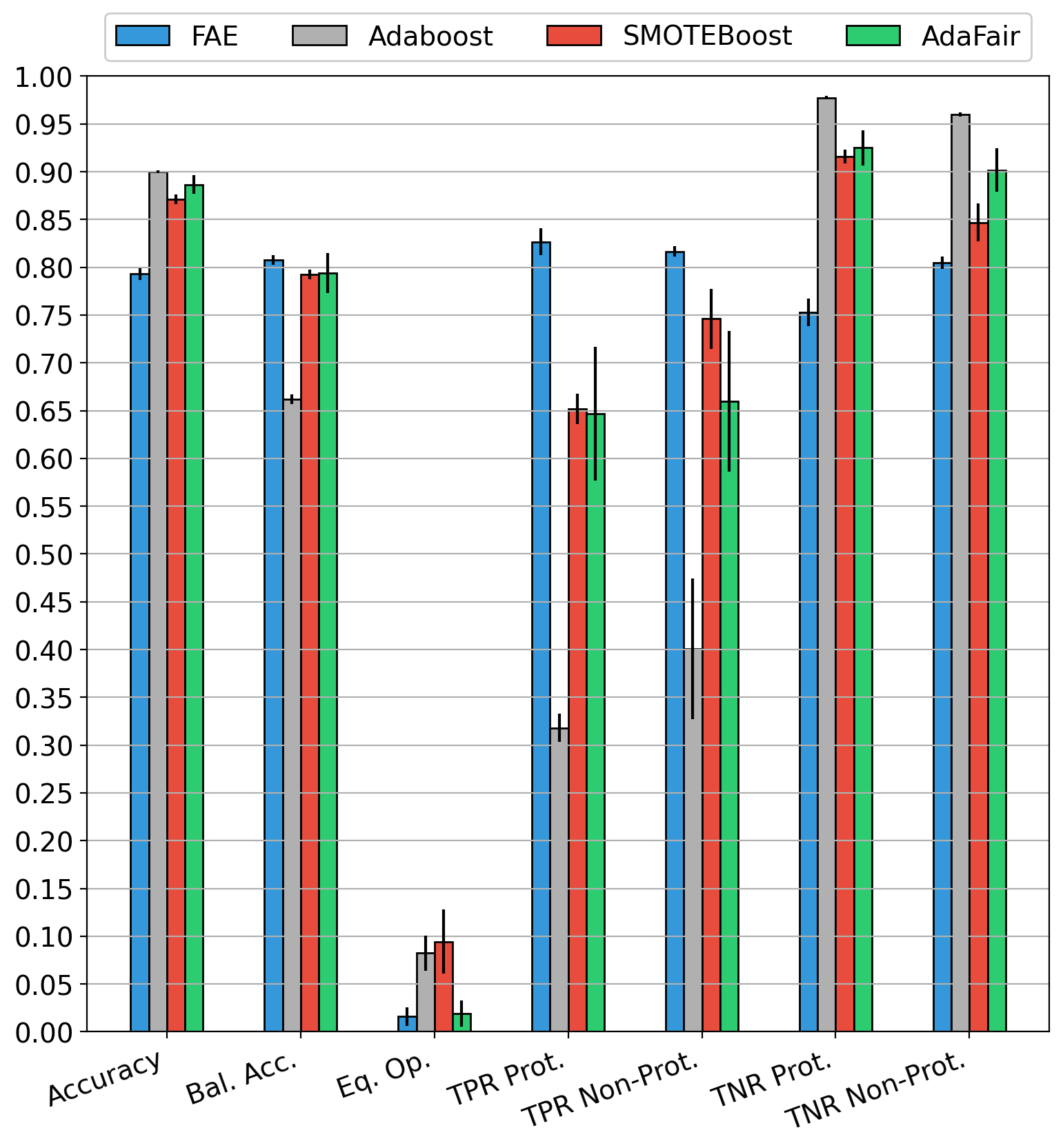}
 \caption{Bank}
   \label{fig:eq_op_bank}
 \end{subfigure}
 \begin{subfigure}[t]{0.40\textwidth}
 \centering
 \includegraphics[width=1.0\columnwidth]{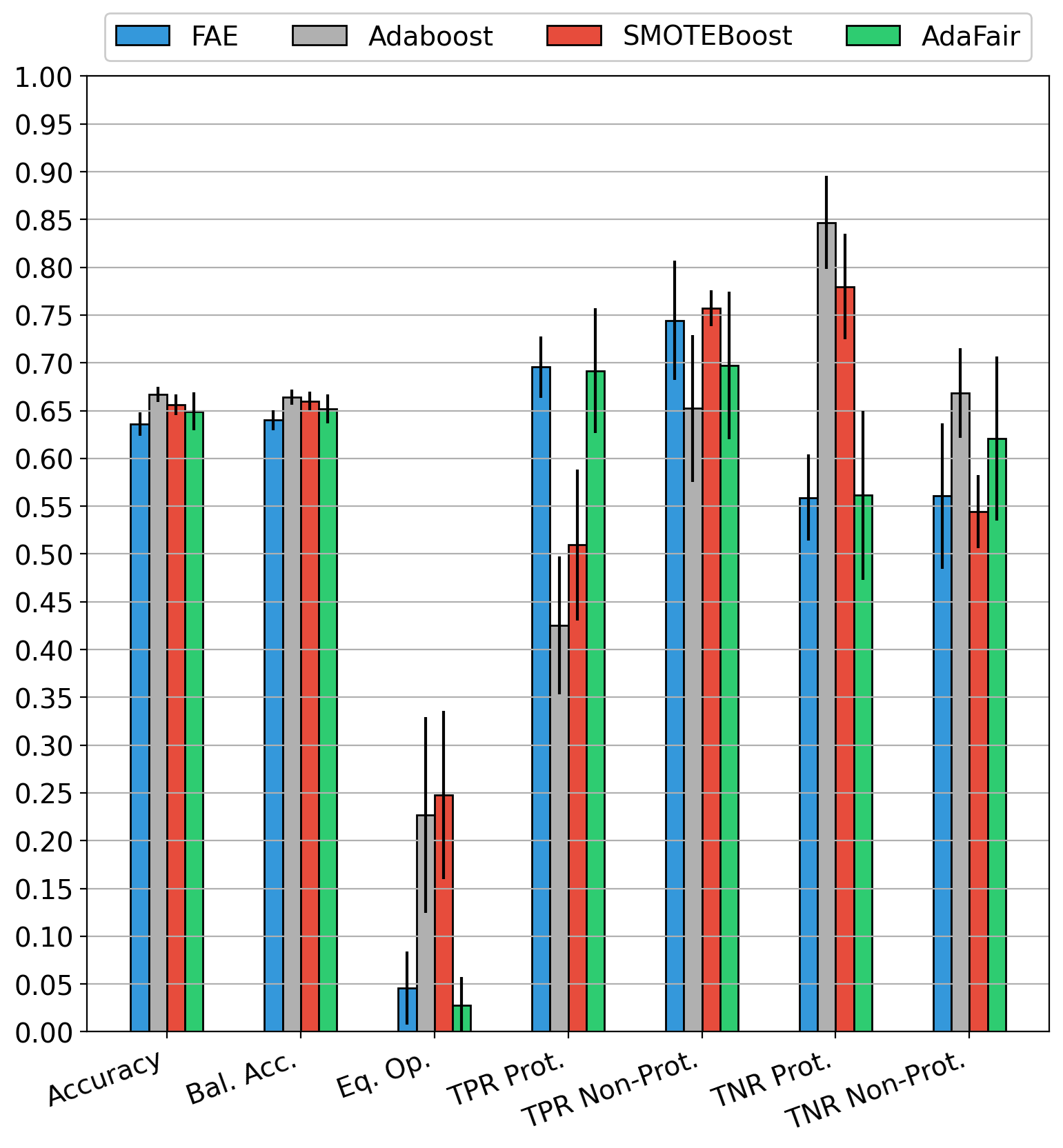}
 \caption{Compass} 
\label{fig:eq_op_compass}
\end{subfigure}
 \begin{subfigure}[t]{0.40\textwidth}
 \centering
 \includegraphics[width=1.0\columnwidth]{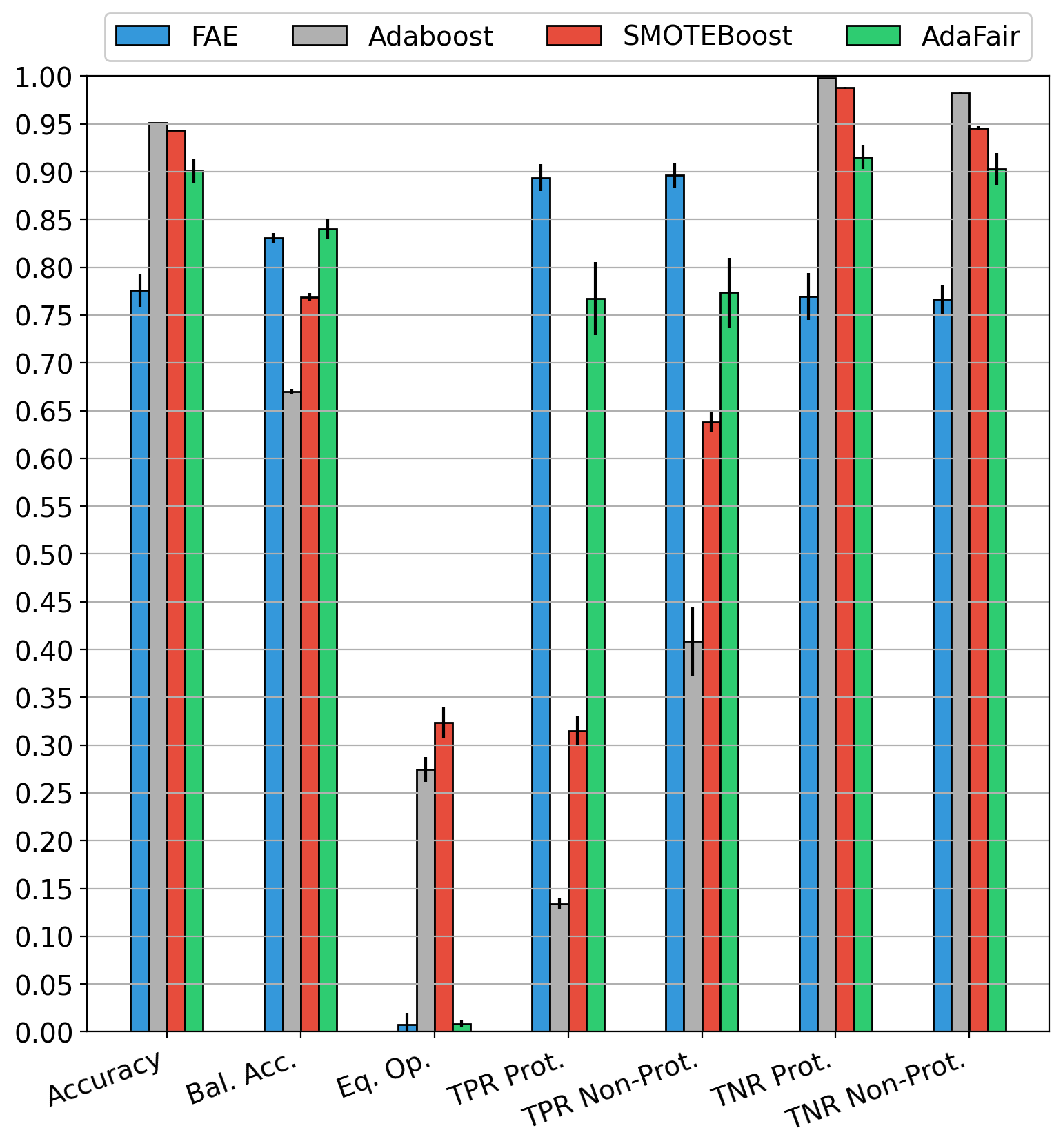}
 \caption{KDD census} 
 \label{fig:eq_op_kdd}
 \end{subfigure}
 \caption{Predictive and fairness, based on Equal Opportunity, performance - higher values are better; for Equal Opportunity, lower values are better.} 
 \label{fig:performance_equal_opportunity}
\end{figure*}

In Figure~\ref{fig:performance_equal_opportunity}, we report on the results for all the approaches w.r.t equal opportunity. Specifically, we report the predictive performance by accuracy and balanced accuracy (Bal. Acc.), and also fairness by equal opportunity (Eq. Op), and TPR and TNR for both protected (Prot.) and non-protected (Non-Prot) groups.

\noindent\textbf{Adult census:} 
In Figure~\ref{fig:eq_op_adult}, we show the results on Adult census dataset. We observe that our AdaFair achieves the lowest equal opportunity score, which FAE follows by a minimal margin. We are also marginally better than FAE in terms of predictive performance, both accuracy and balanced accuracy (around 1\%). FAE achieves almost similar performance for both protected and non-protected groups for both classes. SMOTEBoost achieves the best performance in terms of balanced accuracy (0.5\%$\uparrow$ higher than AdaFair); however, it is the most unfair model (22\%$\uparrow$ higher than AdaFair in terms of equal opportunity).

\noindent\textbf{Bank:} 
In Figure \ref{fig:eq_op_bank}, we report on the results of Bank dataset. AdaFair and FAE have similar equal opportunity scores and balanced accuracy; however, they behave differently. By examining TPR and TNR for both protected and non-protected groups, we observe that FAE outperforms AdaFair by 17\%$\uparrow$ on both groups in positive class. However, it deteriorates its performance in the negative class (17\%$\downarrow$ lower TNR in protected group and 10\%$\downarrow$ lower TNR in non-protected group). SMOTEBoost produces similar predictive performance to AdaFair and FAE, but it cannot mitigate unfair outcomes.

\noindent\textbf{Compass:} 
Figure~\ref{fig:eq_op_compass} shows the results on Compass dataset. We see that AdaFair, on this data, produces the fairest results by achieving the minimum equal opportunity score. By comparing AdaFair to FAE, we observe similar performance. AdaBoost and SMOTEBoost perform similarly in predictive performance and discriminatory outcomes since this dataset does not suffer from class imbalance.

\noindent\textbf{KDD census:} 
The results on KDD census dataset are shown in Figure~\ref{fig:eq_op_kdd}. Same as in the Bank dataset, AdaFair and FAE can mitigate unfair outcomes, but each method's outcome w.r.t TPR and TNR of each group are different. Both AdaFair and FAE outperform SMOTEBoost in terms of balanced accuracy (7\%$\downarrow$ and 6.5\%$\downarrow$, respectively).

\noindent\textbf{Conclusion:} 
FAE is a method that can compete with AdaFair resulting in similar equal opportunity and balanced accuracy scores. In case of severe class imbalance (Bank and KDD census), there is no clear winner:  FAE rejects more negative instances than AdaFair, and AdaFair rejects more positive class instances. For Adult and Compass, both methods behave similarly.

\subsection{Disparate Mistreatment: Predictive and fairness performance}%
\label{sec:exp_comparison}

\begin{figure*}[htp!]
 \centering
 \begin{subfigure}[t]{0.40\textwidth}
 \centering
 \includegraphics[width=1.0\columnwidth]{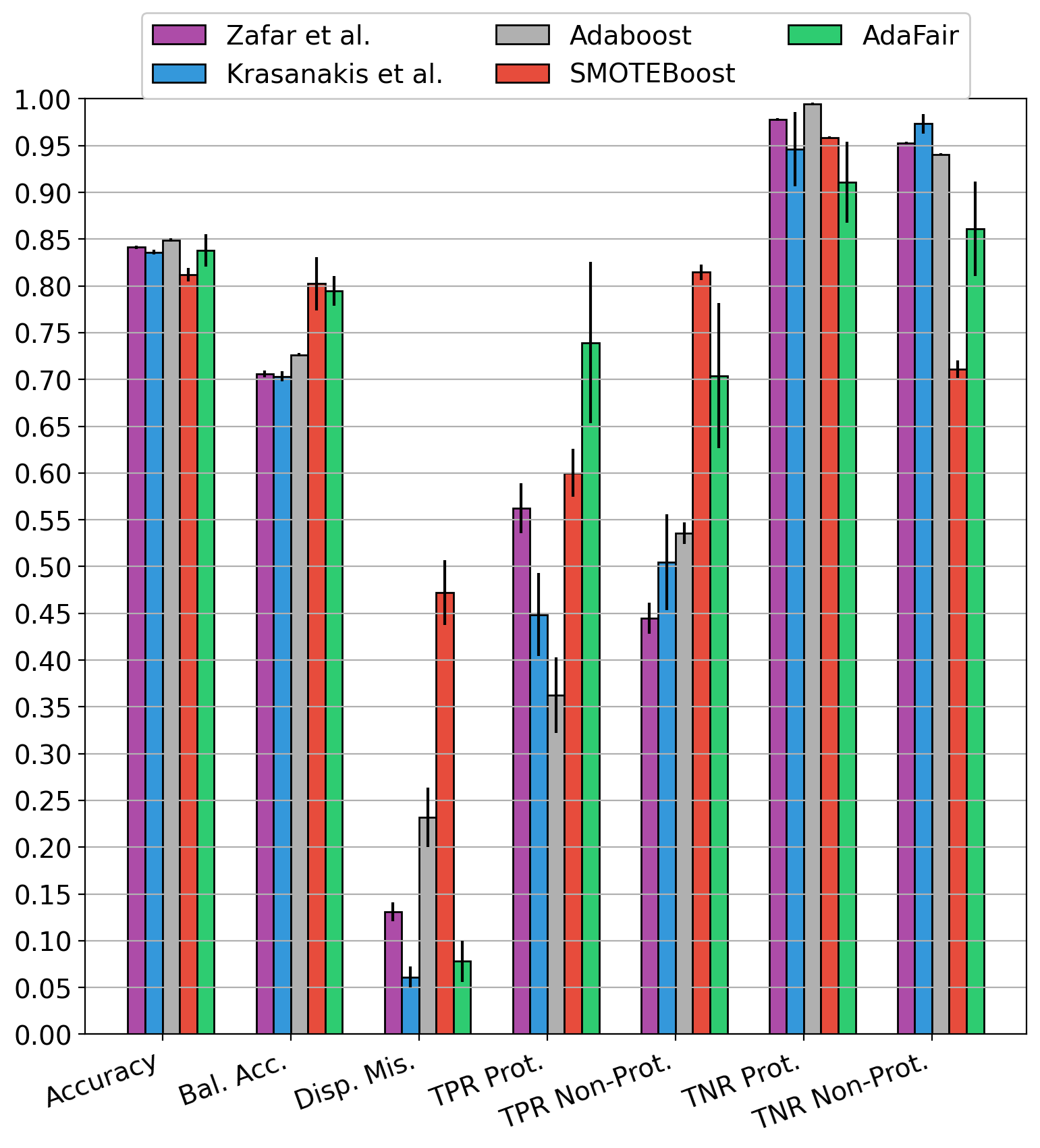}
 \caption{Adult census}
  \label{fig:eq_odds_adult}
 \end{subfigure}
 \centering
 \begin{subfigure}[t]{0.40\textwidth}
 \centering
 \includegraphics[width=1.0\columnwidth]{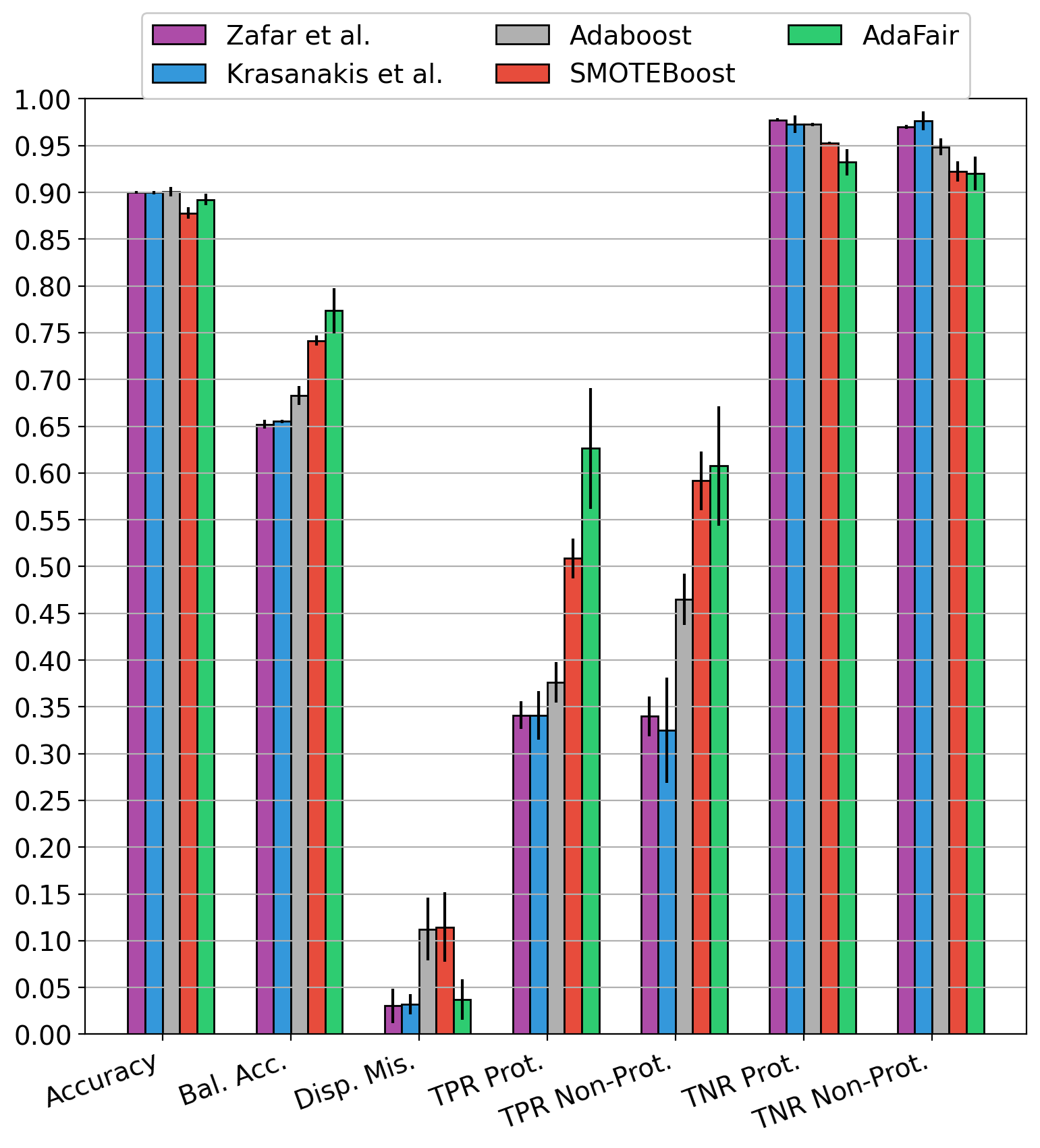}
 \caption{Bank}
   \label{fig:eq_odds_bank}
 \end{subfigure}
 \begin{subfigure}[t]{0.40\textwidth}
 \centering
 \includegraphics[width=1.0\columnwidth]{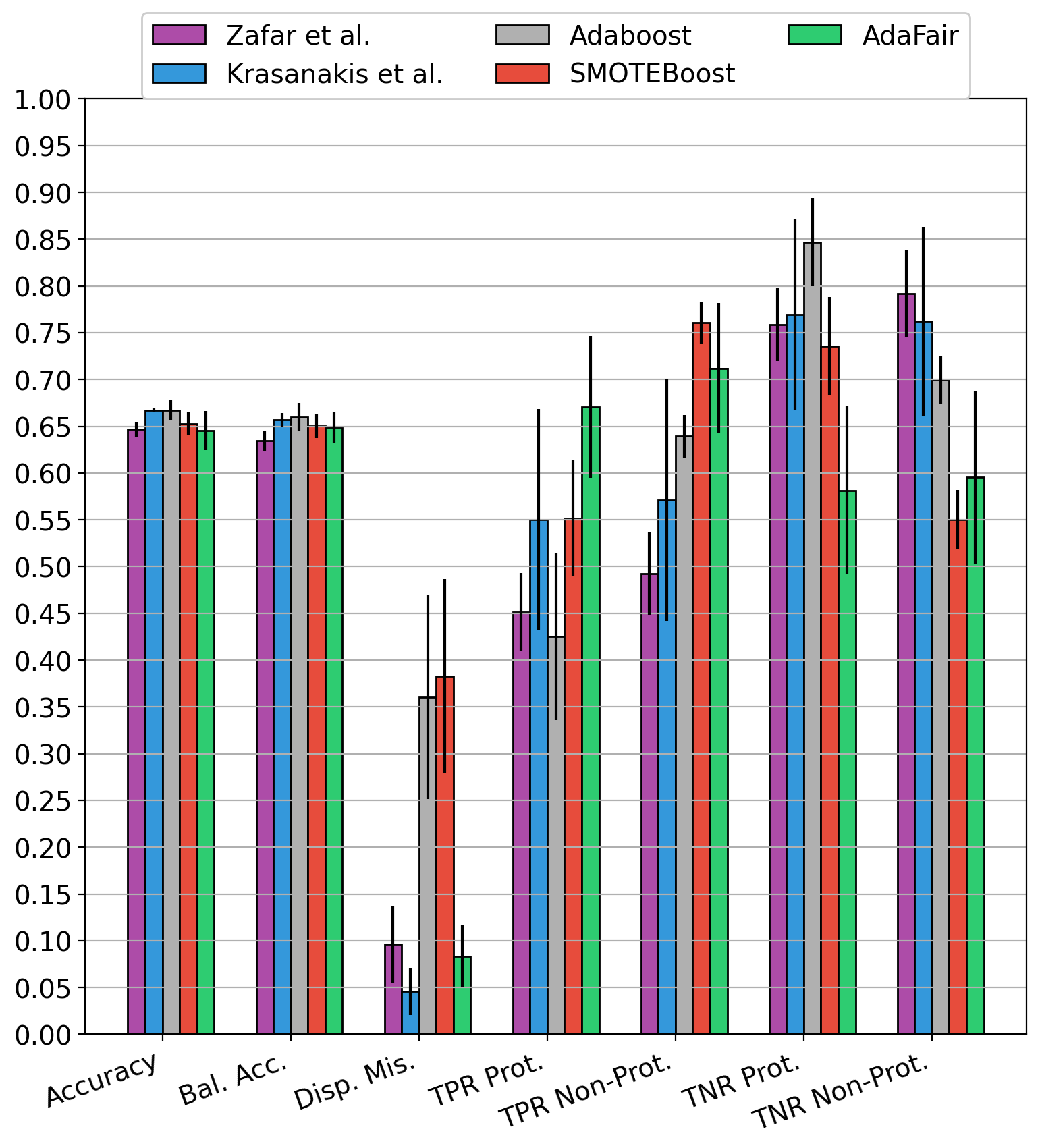}
 \caption{Compass} 
\label{fig:eq_odds_compass}
\end{subfigure}
 \begin{subfigure}[t]{0.40\textwidth}
 \centering
 \includegraphics[width=1.0\columnwidth]{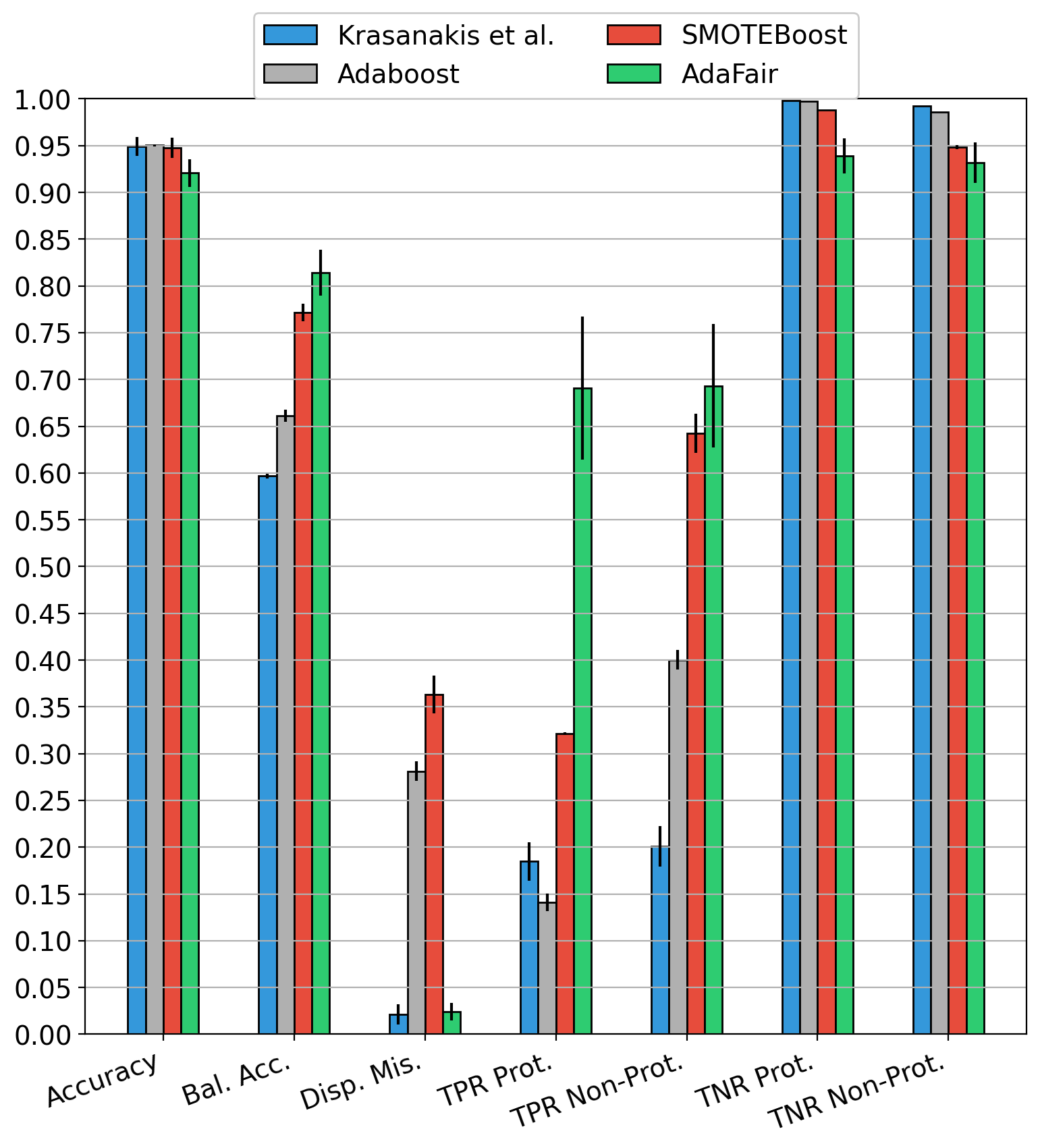}
 \caption{KDD census} 
 \label{fig:eq_odds_kdd}
 \end{subfigure}
 \caption{Predictive and fairness, based on Disparate Mistreatment, performance - higher values are better; for Disparate Mistreatment, lower values are better.} 
 \label{fig:performance_eq_odds}
\end{figure*}

In Figure~\ref{fig:performance_eq_odds}, we show the results w.r.t disparate mistreatment. For reporting, we follow the same measures as in equal opportunity, except for fairness, for which we report on disparate mistreatment ($D.M.$). 

\noindent\textbf{Adult census:} 
In Figure~\ref{fig:eq_odds_adult}, we show the performance of the different approaches on the Adult census dataset. SMOTEBoost achieves the highest balanced accuracy followed by AdaFair ($1$\%$\downarrow$); both methods target class imbalance. The latter, however, also considers fairness. 
AdaBoost, Krasanakis et al. and Zafar et al. that do not consider class imbalance have a 7\%$\downarrow$, 9\%$\downarrow$ and 9\%$\downarrow$, respectively, drop in their balanced accuracy comparing to AdaFair. 
Regarding disparate mistreatment, as expected, AdaBoost and SMOTEBoost perform worse as they do not consider fairness. The best overall disparate mistreatment score is achieved by Krasanakis et al., followed by our AdaFair ($2\%\uparrow$, recall that lower values are better). However, a closer look at the actual TPRs and TNRs values per group show that our method achieves the highest TPRs values for both protected and non-protected groups compared to the other two fairness aware approaches. In particular, for the protected (non-protected) group, our TPR is 29\%$\uparrow$ (20\%$\uparrow$, respectively) higher than the second-best method of Krasanakis et al. 
So, it seems that Krasanakis et al. and Zafar et al. produce low TPRs and high TNRs, i.e., these methods ``reject'' more positive class instances to minimize disparate mistreatment (this explains their high TNRs, low TPRs values). On the contrary, our AdaFair achieves good performance for both classes (high TPRs, high TNRs) while maintaining good disparate mistreatment (i.e., low differences in TPRs, TNRs for both protected and non-protected groups). 

\noindent\textbf{Bank:} 
The results are shown in Figure~\ref{fig:eq_odds_bank}. All approaches, except for AdaBoost and SMOTEBoost, achieve low disparate mistreatment. Interestingly, AdaFair achieves higher balanced accuracy than SMOTEBoost, while it outperforms the other approaches. 
A closer look at disparate mistreatment, namely at TPRs and TNRs, shows significant differences between the approaches.
Namely, w.r.t TPRs, our method outperforms the second-best (Zafar et al.) by almost 29\%$\uparrow$ for each group. Interestingly, AdaBoost and SMOTEBoost maintain higher TPRs than the methods of Krasanakis et al. and Zafar et al., even though they do not consider fairness. The methods of Krasanakis et al. and Zafar et al. have very similar behaviour, and it seems that both focus on the majority class (therefore high TNRs, low TPRs). 
Regarding TNRs, our method has a small drop of 6\%$\downarrow$ and 7\%$\downarrow$ drop for the protected and non-protected groups compared to the second-best approach of Zafar et al.; this is expected as we optimize for balanced error rather than an overall error.

\noindent\textbf{Compass:} 
The results are shown in Figure~\ref{fig:eq_odds_compass}.
Regarding balanced accuracy, AdaBoost performs best and Zafar et al. worse. However, the differences between the approaches are not that high. We expect a similar performance of the different approaches as the dataset is balanced (c.f., Table~\ref{tbl:datasets}), and therefore, imbalance treatment has no substantial effect. 
Regarding fairness, the method of Krasanakis et al. achieves better performance in terms of disparate mistreatment (3.5\%$\downarrow$) and balanced accuracy (0.2\%$\uparrow$) compared to the second-best  AdaFair. 
Zafar et al. have the worst disparate mistreatment (almost twice the value of Krasanakis et al.), recall that its $Bal.Acc.$ was the worse among the approaches. 
By examining the TPRs and TNRs of both protected and non-protected groups, we observe that the performance of Krasanakis et al. is not stable (highest standard deviation among the methods). Our AdaFair has better TPR values for both groups. Our TNRs are the lowest among the approaches ($~57\%-59\%$) as we optimize for balanced error, and the negative class represents 54\% of the population. 

\noindent\textbf{KDD census:} 
We could not use the Zafar et al. approach to this dataset due to its inability to estimate the optimal parameters. 
In balanced accuracy, AdaFair performs 22\%$\uparrow$ than Krasanakis et al., 17\%$\uparrow$ than AdaBoost and 5\%$\uparrow$ than SMOTEBoost. AdaBoost and Krasanakis et al. classify almost perfectly the negative class (i.e., TNRs close to 100\%), which comprises 94\% of the population. SMOTEBoost has 2\%$\downarrow$ to 4\%$\downarrow$ drop in TNRs of protected and non-protected groups, respectively, compared to AdaBoost.
Regarding TPR, AdaFair bags the highest TPR scores for both groups (above 65\%) while the method of  Krasanakis et al. results in values below 20\%. 
Both fairness-aware approaches, AdaFair and Krasanakis et al., minimize discrimination to 2\% while AdaBoost and SMOTEBoost result in 28\% and 36\% disparate mistreatment, respectively.

\noindent\textbf{Conclusion:} AdaFair is capable of achieving high balanced accuracy and low discrimination by maintaining high TPRs and only slightly worse TNRs for both groups. On the contrary, the other fairness-aware approaches, namely Zafar et al. and Krasanakis et al., eliminate discrimination by reducing TPRs, that is, by rejecting more instances of the positive class to achieve parity among the protected and non-protected groups. 
Moreover, Zafar et al. cannot estimate the optimal parameters for datasets with many attributes (see KDD census failure).

\subsection{Performance over the boosting rounds}%
\begin{figure*}[htp!]
 \centering
 \begin{subfigure}[t]{1\textwidth}
 \centering
 \includegraphics[width=1.0\columnwidth]{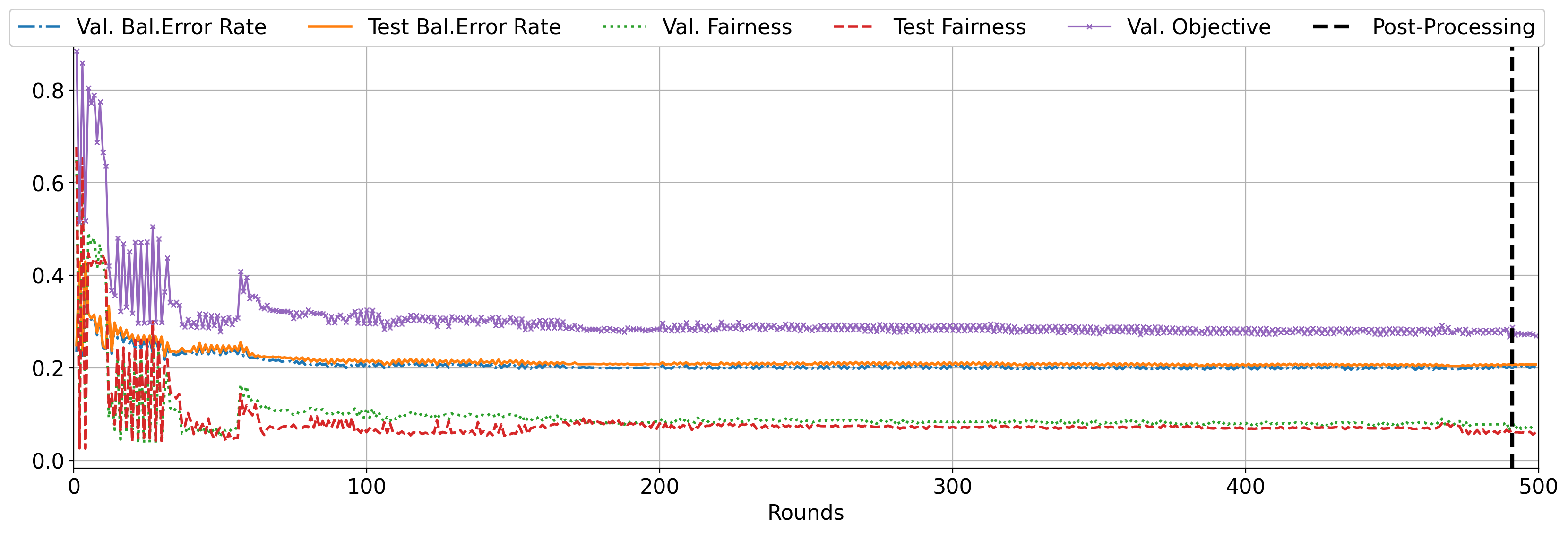}
 \caption{Adult census}
  \label{fig:over_the_rounds_adult}
 \end{subfigure}
 \centering
 \begin{subfigure}[t]{1\textwidth}
 \centering
 \includegraphics[width=1.0\columnwidth]{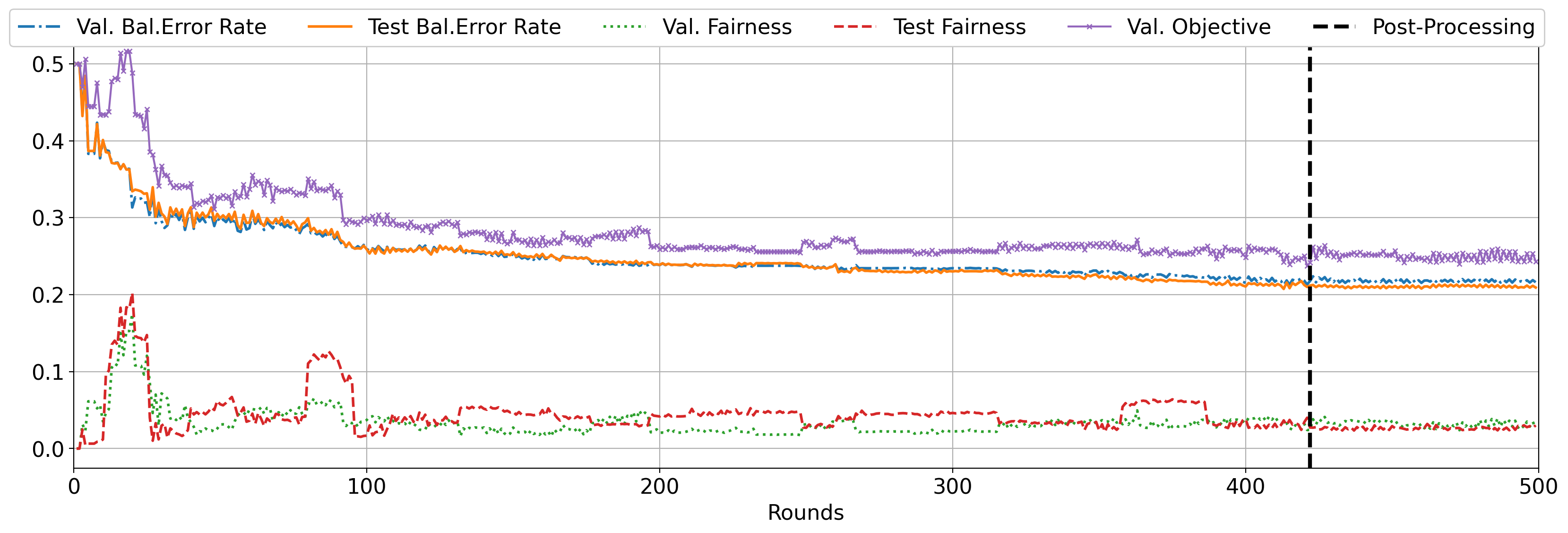}
 \caption{Bank}
   \label{fig:over_the_rounds_bank}
 \end{subfigure}
 \begin{subfigure}[t]{1\textwidth}
 \centering
 \includegraphics[width=1.0\columnwidth]{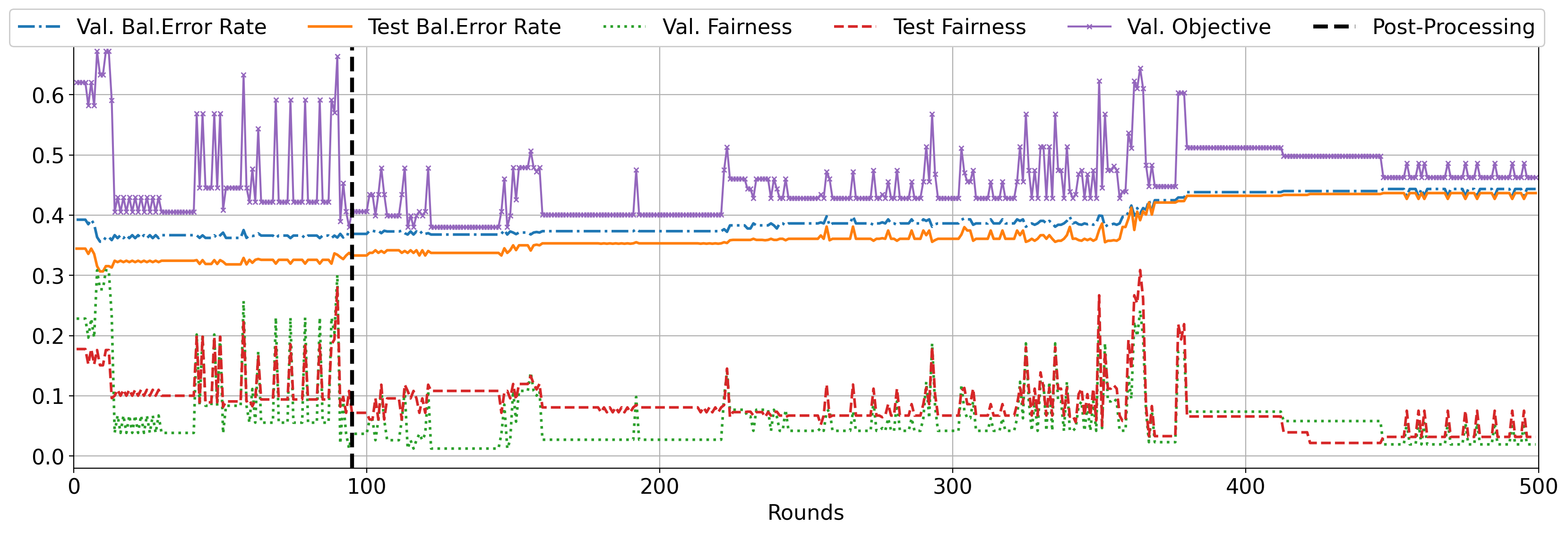}
 \caption{Compass} 
\label{fig:over_the_rounds_compass}
\end{subfigure}
 \begin{subfigure}[t]{1\textwidth}
 \centering
 \includegraphics[width=1.0\columnwidth]{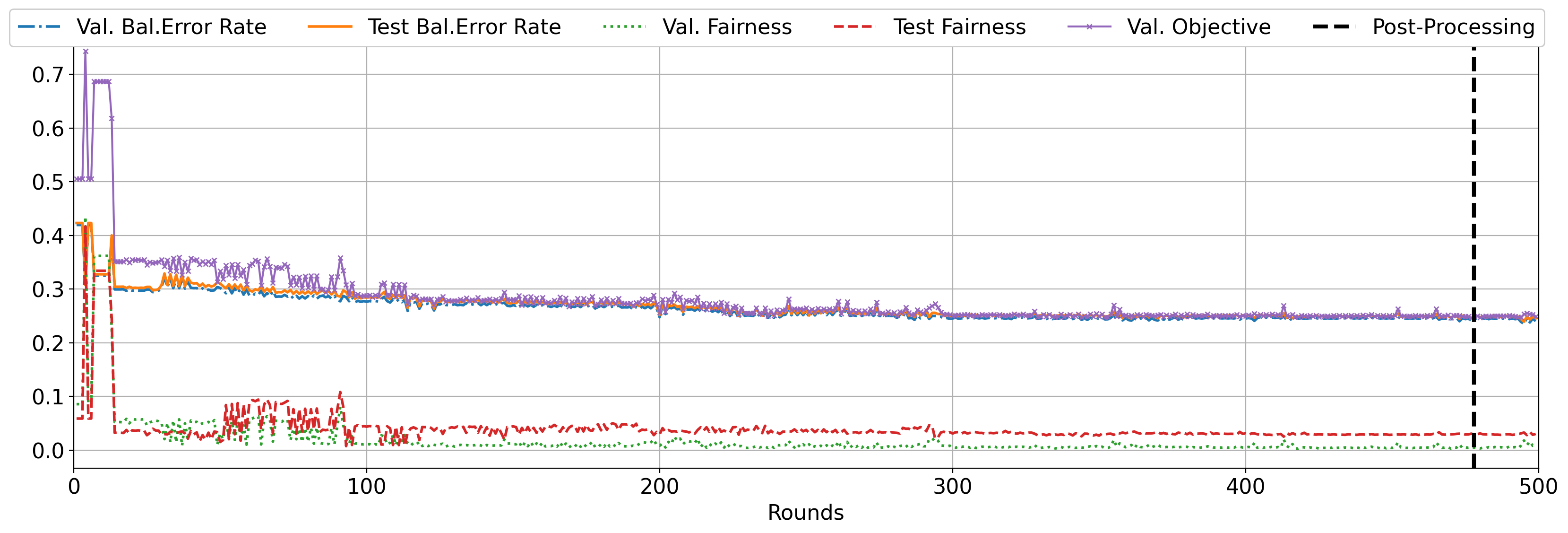}
 \caption{KDD census} 
 \label{fig:over_the_rounds_kdd}
 \end{subfigure}
 \caption{Analysis over the boosting rounds on the validation and test set; lower values are better. Fairness measure: Disparate Mistreatment.} 
 \label{fig:over_the_rounds}
\end{figure*}

\label{sec:over_the_rounds}
In this section, we analyse the performance of AdaFair over the boosting rounds. The purpose of this experiment is to show the per round behaviour of AdaFair w.r.t the objective function, $\theta$ selection and its performance on the test set. For this analysis, we have selected parameters $c=1$ and $T=500$ and report on a holdout evaluation (67\% training - 33\% testing sets) for each dataset. 
We report only on disparate mistreatment as it is the most complex measure comparing to statistical parity and equal opportunity.
In Figure~\ref{fig:over_the_rounds}, we report on the performance of AdaFair on the test set (only used for evaluation, not for training) and the validation set, which is used for selecting the $\theta$ (as mentioned in Section~\ref{sec:parameter_selection}, the validation set is 33\% of the training set). We report on the balanced error rate and fairness on both test and validation set, and also we report on the objective function~\eqref{eq:argmin} w.r.t the validation set. 
\\
\noindent\textbf{Adult census:} The results are shown in Figure~\ref{fig:over_the_rounds_adult}. We can observe that the validation set captures the underlying distribution of the test set since they have almost identical score values. In the early boosting rounds ($T < 100$), AdaFair's performance fluctuates until it stabilizes. As the number of boosting rounds increase, the balanced error rate remains the same and the fairness on the validation set. The values from the objective function stabilize around $[0.27, 0.29]$ for $T > 150$. After the training process, we select the best partial ensemble from weak learners 0 to 488. However, we can see that the performance on the test set does not change significantly after $T > 100$.\\ \noindent\textbf{Bank:} In Figure~\ref{fig:over_the_rounds_bank}, we report the results on the Bank dataset. Like the Adult census dataset, high fluctuation can be observed in the early boosting rounds (e.g., $T < 100$). With boosting rounds increasing, the balanced error rate decreases and stabilises around $0.22$ after $T > 400$ (on validation and test set). The fairness on the validation and test set has some small but insignificant fluctuations. In this dataset, the gradual decrease of balanced error rate is the factor that defines $\theta$ since fairness remains low after $T > 100$ (on the validation set).  \\
\noindent\textbf{Compass:} In Figure~\ref{fig:over_the_rounds_compass}, we report on the results of Compass dataset. In contrast to the other datasets, the objective function in Compass fluctuates for small and high values of $T$. The balanced error rate increases gradually over the rounds, and fairness fluctuates highly, fluctuating the objective function. This behaviour is probably caused due to the complexity of the dataset, e.g., we have seen in Section~\ref{sec:exp_comparison} that none of the methods could achieve more than 70\% accuracy or balanced accuracy on this dataset. \\
\noindent\textbf{KDD census:} In Figure~\ref{fig:over_the_rounds_kdd}, we report on the results of KDD census dataset. This dataset shows similar behaviour to the other two imbalanced datasets (Adult census and Bank). Early boosting rounds ($T < 100$) are characterised by fluctuated values w.r.t predictive performance, fairness and objective function. After 150 rounds, the performance is stabilised, and we observe a non-significant decrease in the balanced error rate after 200 rounds.\\
\noindent\textbf{Conclusions:} 
For the selected datasets, our experiments show that AdaFair needs at least 100 rounds to produce good results. 
Although AdaFair does not tackle class-imbalance in-training, we observe that the balanced error rate decreases over the rounds. This lies in the fairness-related costs that indirectly push the model to also learn the minority class. 
A complete in-training tackling of fairness and class-imbalance is left for future research.

\subsection{Cumulative vs non-cumulative fairness}%
\label{sec:single_vs_accum}
The notion of cumulative fairness, is crucial for AdaFair's ability to mitigate discrimination w.r.t i) statistical parity (Equation~\eqref{eq:cumulsp}), ii) equal opportunity (Equation~\eqref{eq:cumuleqop}), and iii) disparate mistreatment (Equation~\eqref{eq:cumulFairEO}). 
To investigate its impact, we compare AdaFair (with cumulative fairness of models $1:j$, where $j$ is the current boosting round) with a version that considers only the fairness of the individual weak learner in round $j$ (refereed to as AdaFair NoCumul), for all the employed fairness notions. Below, we show the behaviour on disparate mistreatment. The behaviour is similar across all the employed fairness notions (see Appendix for the exact outcome on statistical parity and equal opportunity).

\begin{figure*}[htp!]
 \centering
 \begin{subfigure}[t]{0.40\textwidth}
 \centering
 \includegraphics[width=1.00\columnwidth]{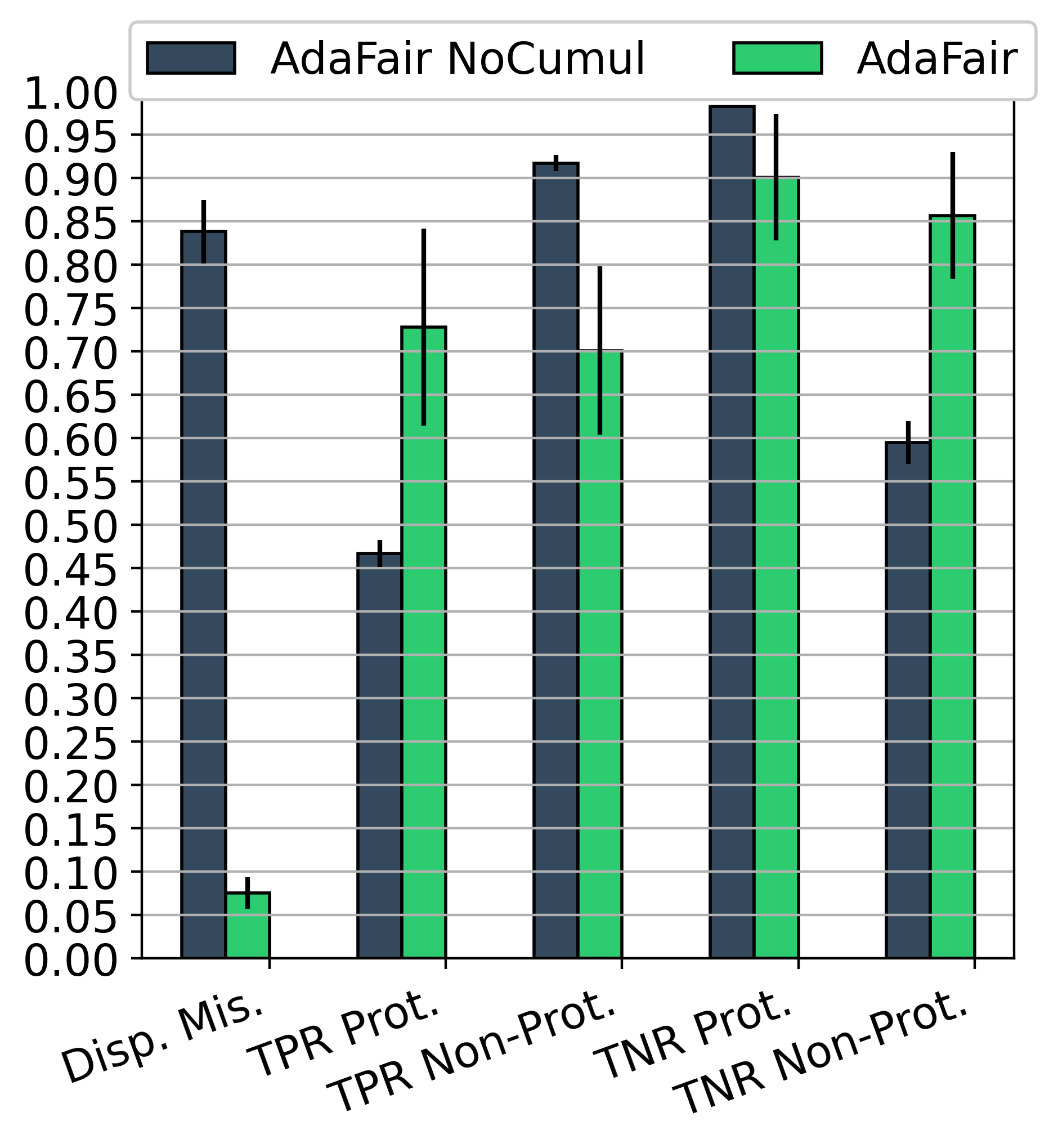}
 \caption{Adult census}
 \end{subfigure}
 \centering
 \begin{subfigure}[t]{0.40\textwidth}
 \centering
 \includegraphics[width=1.00\columnwidth]{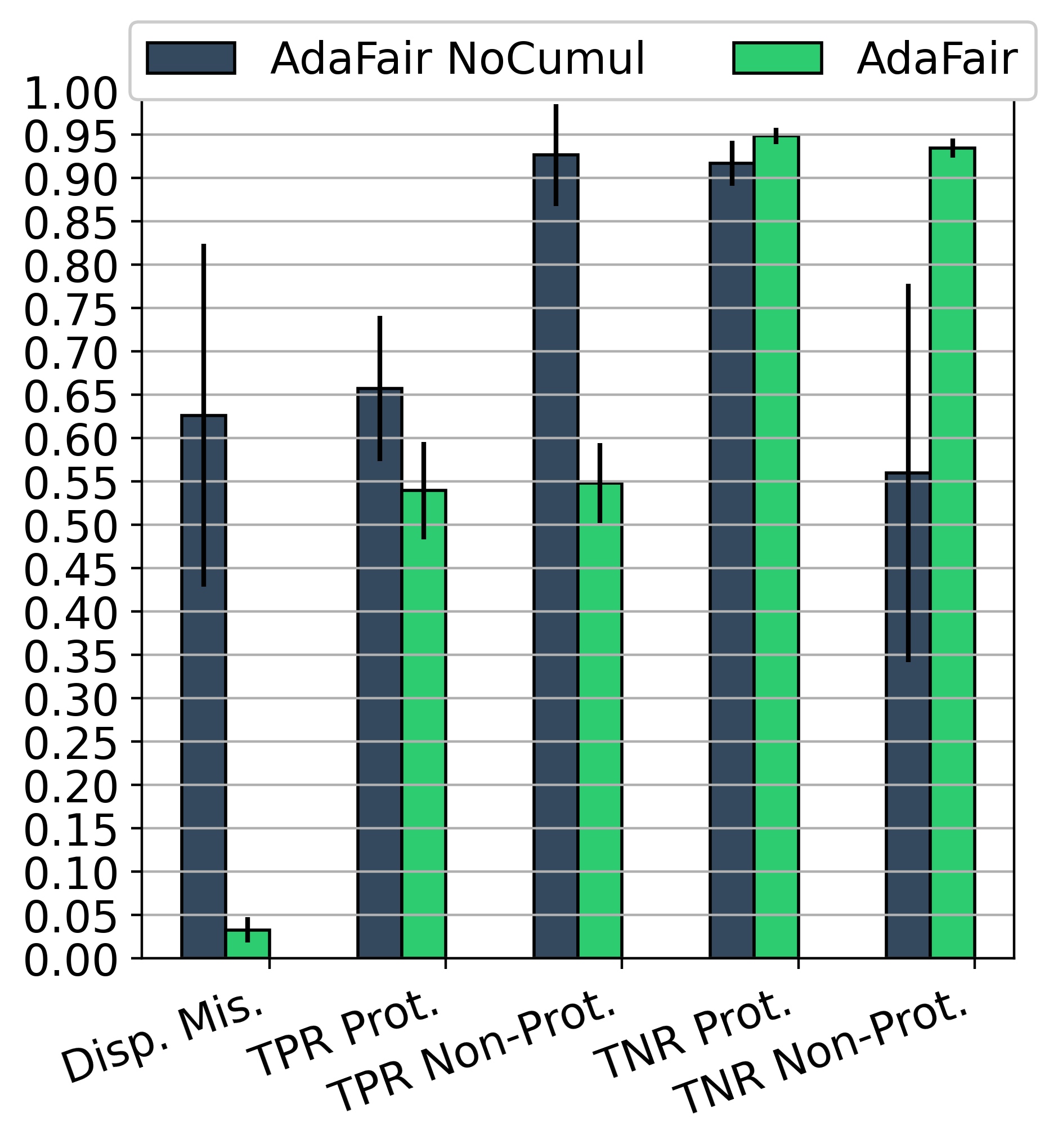}
 \caption{Bank}
 \end{subfigure}
 \begin{subfigure}[t]{0.40\textwidth}
 \centering
 \includegraphics[width=1.00\columnwidth]{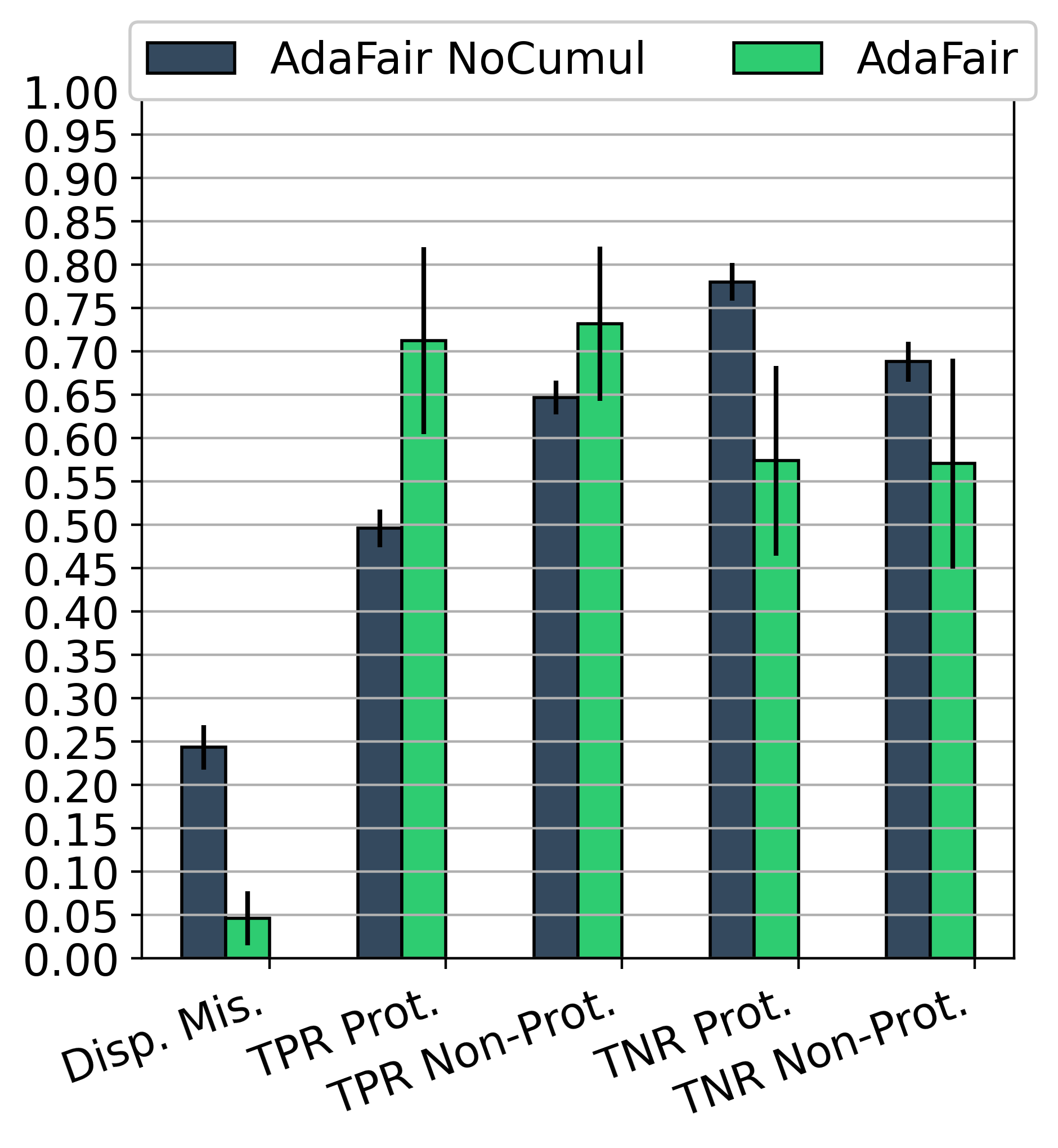}
 \caption{Compass} 
 \end{subfigure}
 \begin{subfigure}[t]{0.40\textwidth}
 \centering
 \includegraphics[width=1.00\columnwidth]{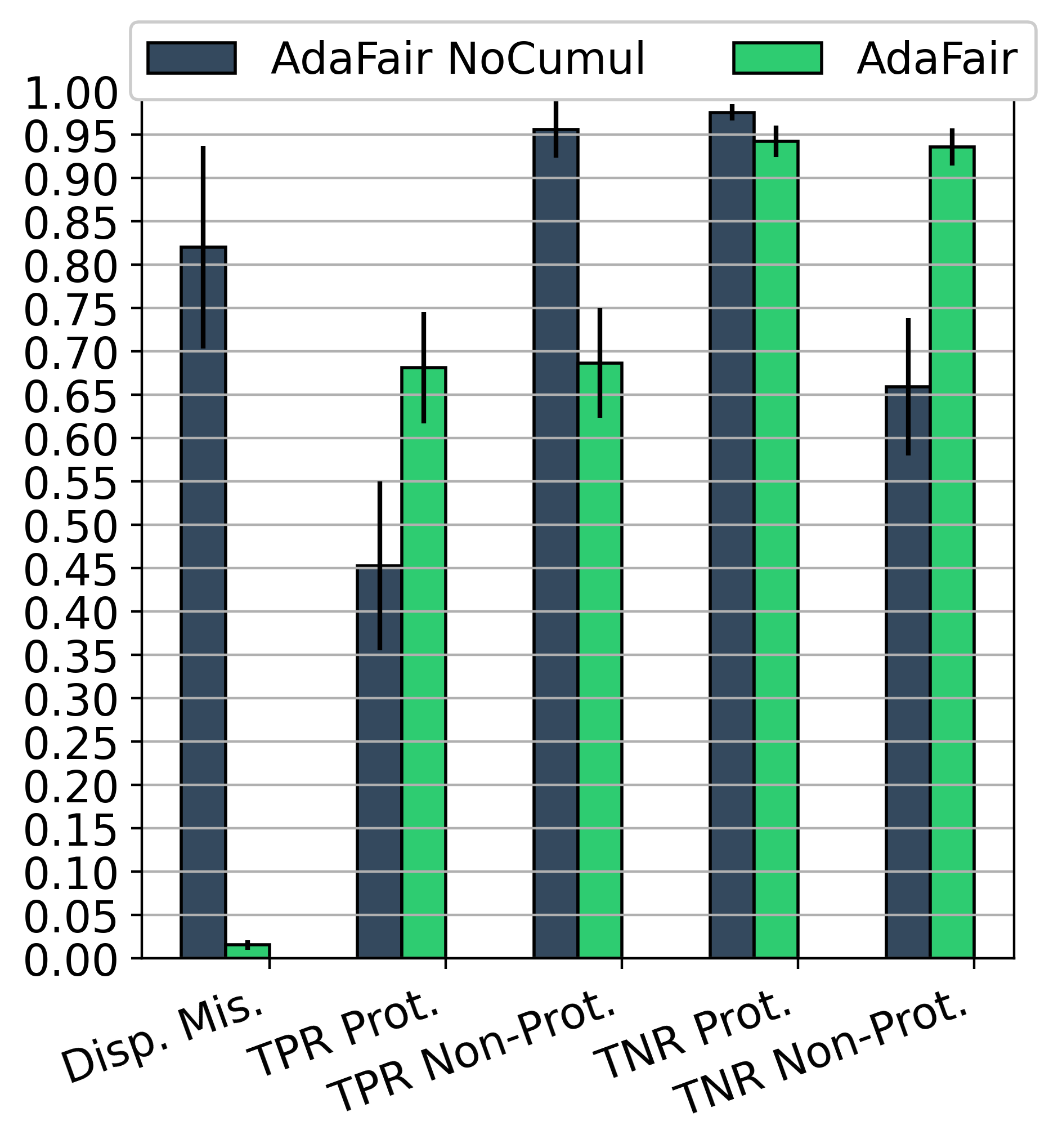}
 \caption{KDD census} 
 \end{subfigure}
 \caption{
Disparate Mistreatment: 
AdaFair vs AdaFair NoCumul} 
 \label{fig:single_performance_equalized_odds}
\end{figure*}

\noindent\textbf{Disparate Mistreatment:}
The fairness performance w.r.t disparate mistreatment, for the different datasets is shown in Figure~\ref{fig:single_performance_equalized_odds}. Overall, the AdaFair NoCumul method results in poor fairness performance with very high $D.M.$ values compared to AdaFair.
In particular, we observe an increase of 78\%$\uparrow$ for the Adult census dataset, 59\%$\uparrow$ for Bank and 20\%$\uparrow$ Compass datasets, of 80\%$\uparrow$ for the KDD census dataset. A closer look at the individual TPR, TNR scores show that the protected group scores are lower regarding TPR. In contrast, w.r.t TNR, the scores of the protected group are higher (we notice opposite behaviour for the Bank dataset). That is, more protected instances are rejected (low TPR, high TNR). Moreover, the standard deviation for the non-cumulative version is higher than AdaFair, indicating AdaFair NoCumul is not stable. 

\begin{figure*}[t!]
 \centering
 \begin{subfigure}[t]{0.40\textwidth}
 \centering
 \includegraphics[width=1.00\columnwidth]{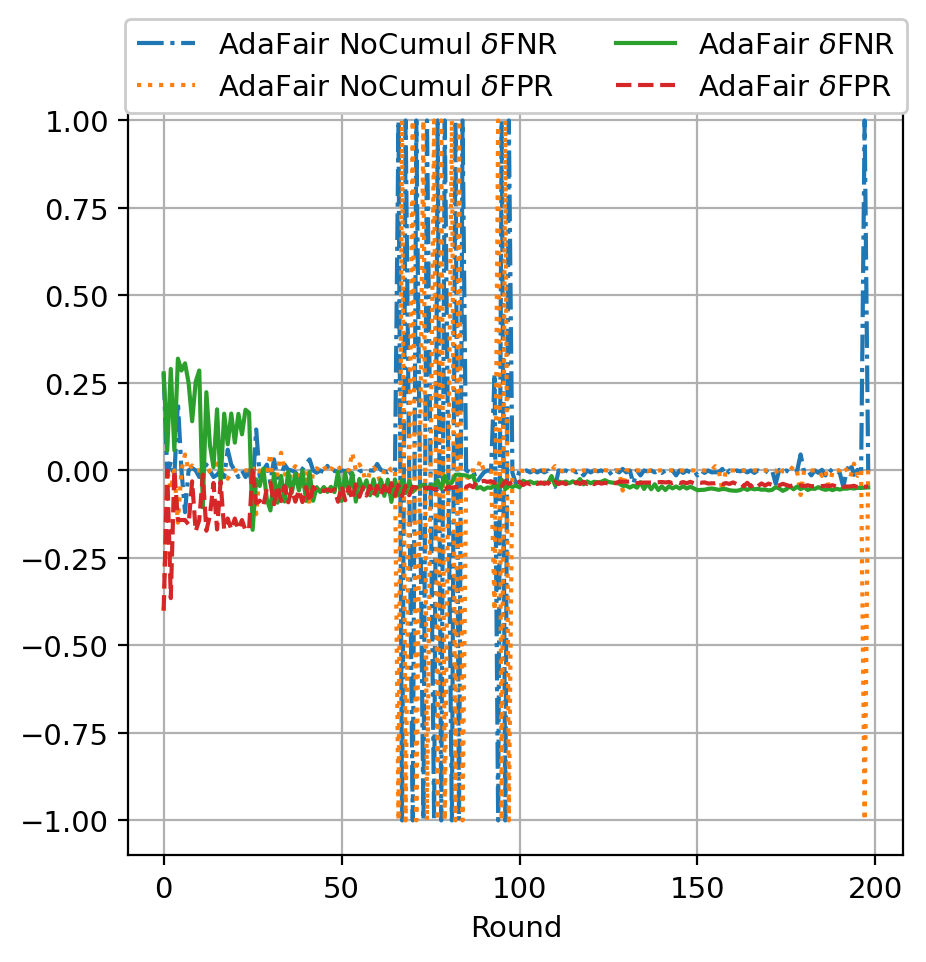}
 \caption{Adult census}
 \end{subfigure}
 \centering
 \begin{subfigure}[t]{0.40\textwidth}
 \centering
 \includegraphics[width=1.00\columnwidth]{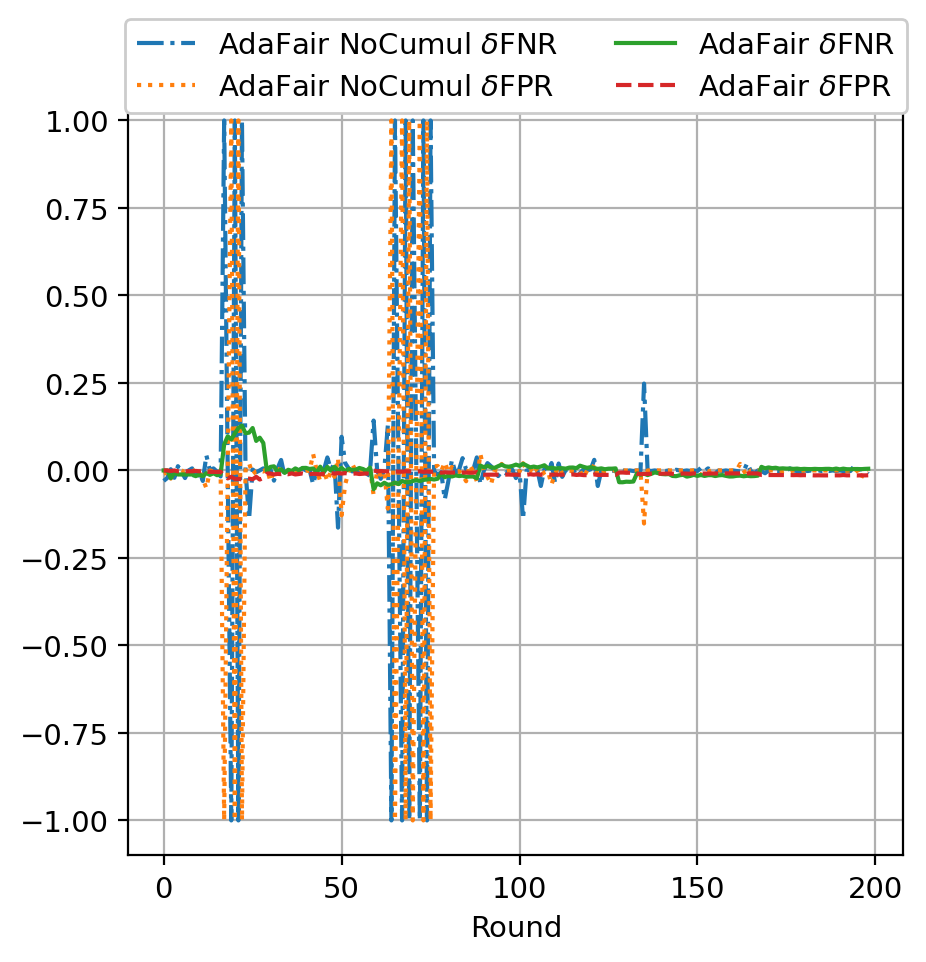}
 \caption{Bank}
 \end{subfigure}
 \begin{subfigure}[t]{0.40\textwidth}
 \centering
 \includegraphics[width=1.00\columnwidth]{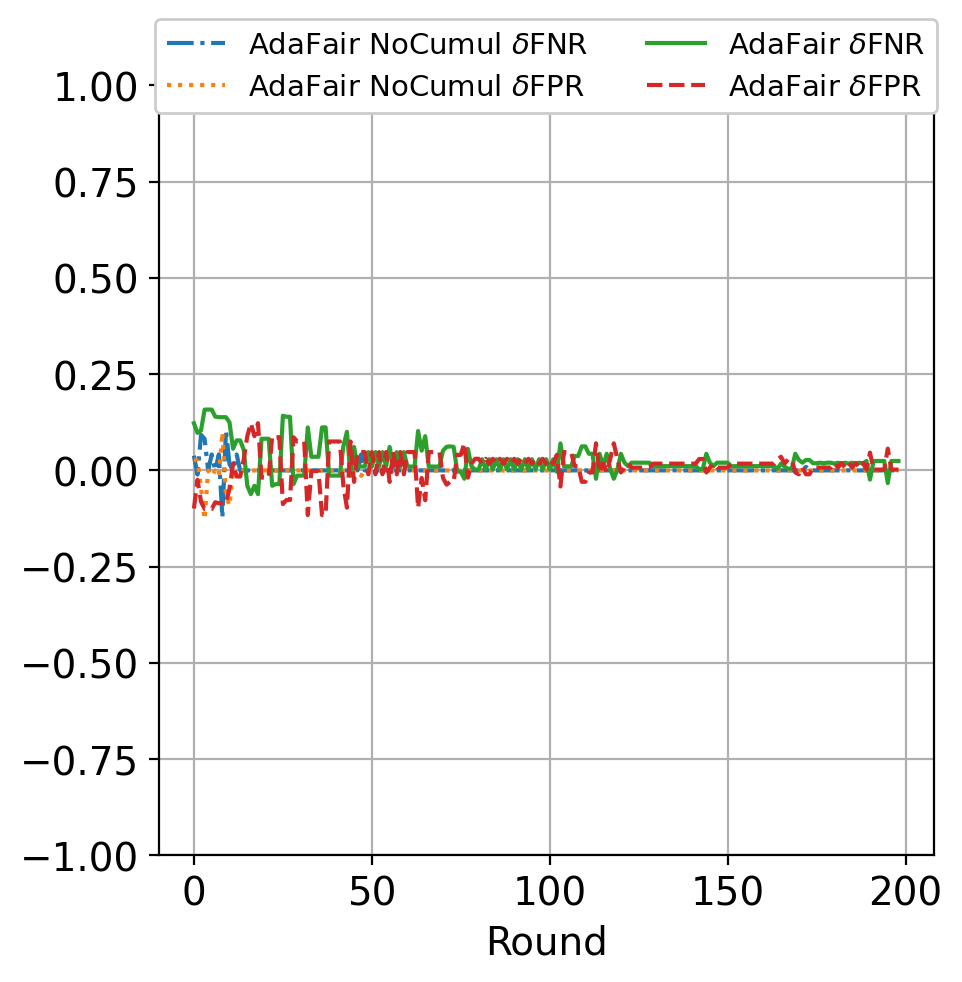}
 \caption{Compass} 
 \end{subfigure}
 \begin{subfigure}[t]{0.40\textwidth}
 \centering
 \includegraphics[width=1.00\columnwidth]{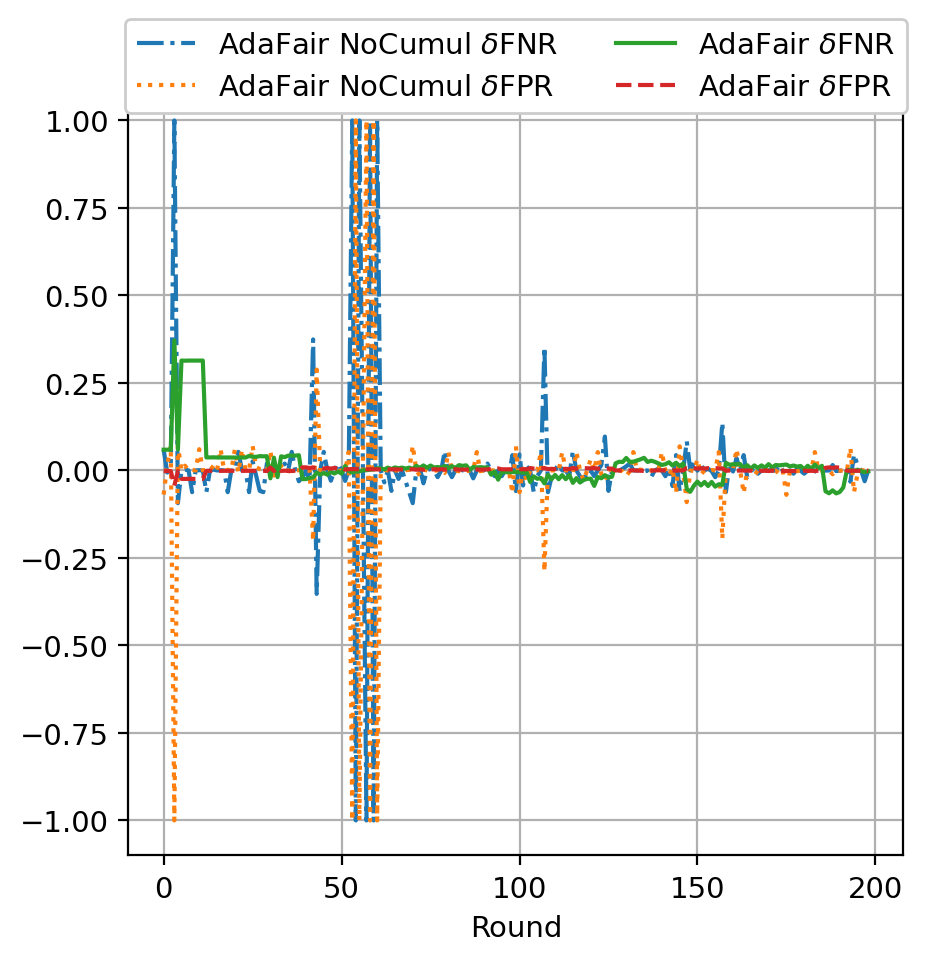}
 \caption{KDD census} 
 \end{subfigure}
 \caption{Disparate Mistreatment, fairness-related costs per boosting round: AdaFair vs AdaFair NoCumul} 
 \label{fig:round_costs_eqodds}
\end{figure*}

In Figure~\ref{fig:round_costs_eqodds}, we compare the per round $\delta FNRs$ and $\delta FPRs$ of the two approaches. Recall that $\delta FNR$ and $\delta FPR$ define the fairness-related cost $u$ related to disparate mistreatment fairness that affects the weighting of the instances for the next round (Equation~\eqref{eq:fairnessCostsEO}).
Fairness-related costs of AdaFair NoCumul exhibit a high fluctuation. On the contrary, the fairness-related costs for our AdaFair are smoother and converge after a sufficient number of rounds to a specific range $[-0.05, 0.05]$. That means that our method mitigates discrimination over the early rounds. These results further confirm that the cumulative definition of fairness is superior to a non-cumulative approach.

\noindent\textbf{Conclusion:} 
By considering the cumulative fairness of the ensemble, AdaFair can produce fairness-related costs which are more stable than its non-cumulative version. These costs allow the model to find a hypothesis that mitigates discriminatory outcomes significantly better than non-cumulative costs. 



\subsection{The effect of balanced error}%
\begin{figure*}[t!]
  \centering
  \begin{subfigure}[t]{0.40\textwidth}
    \centering
  \includegraphics[width=1.0\columnwidth]{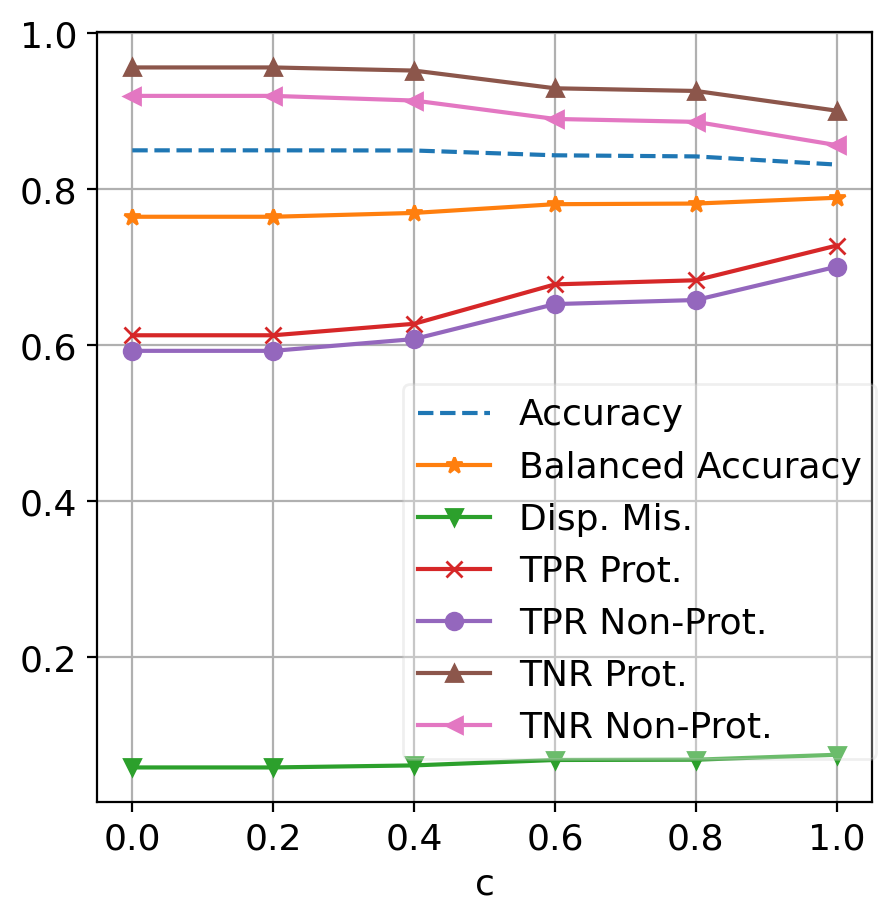}
    \caption{Adult census}
  \end{subfigure}
  \begin{subfigure}[t]{0.40\textwidth}
    \centering
  \includegraphics[width=1.0\columnwidth]{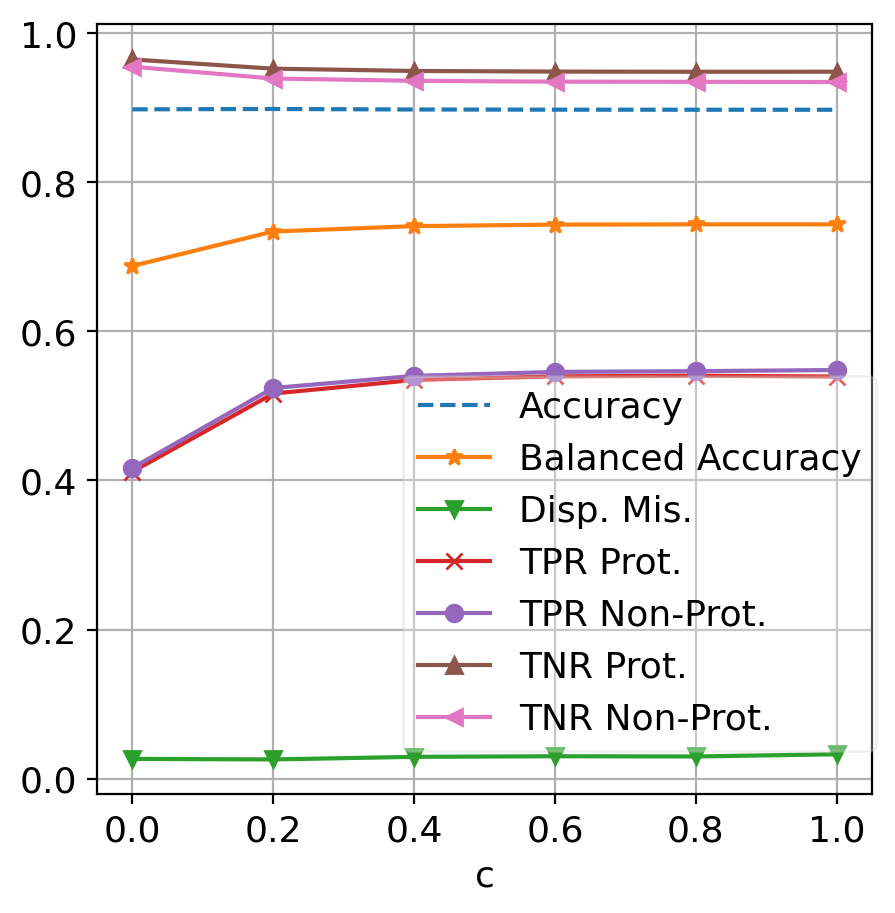}
    \caption{Bank}
  \end{subfigure}
  \begin{subfigure}[t]{0.40\textwidth}
    \centering
    \includegraphics[width=1.0\columnwidth]{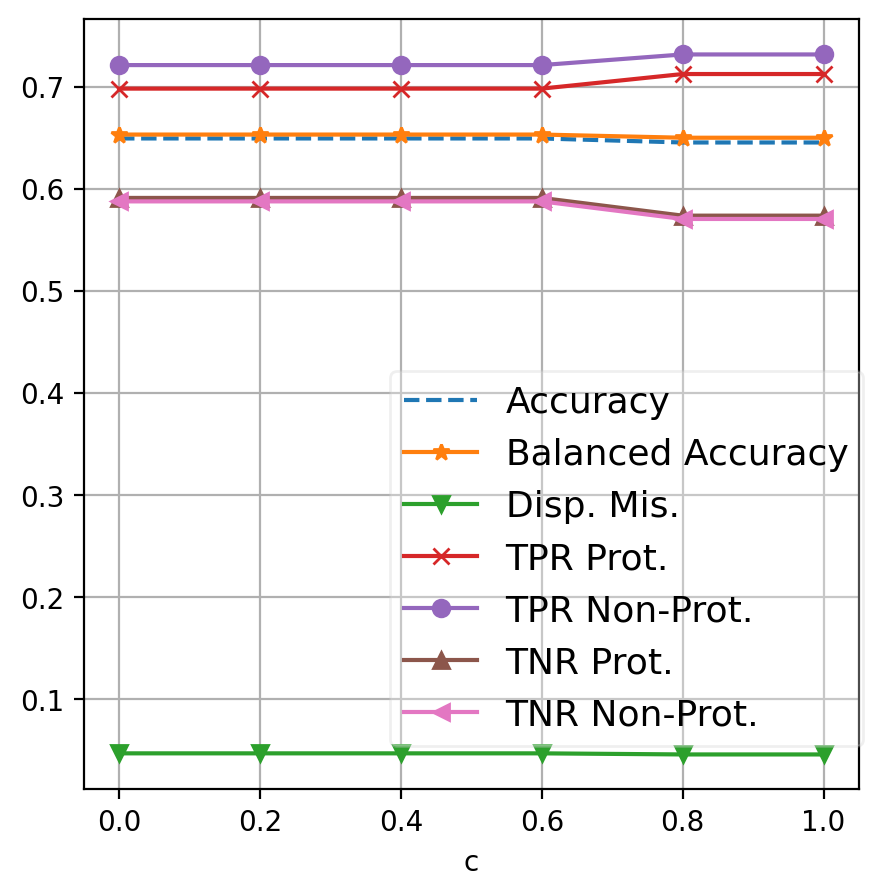}
    \caption{Compass}
  \end{subfigure}
    \begin{subfigure}[t]{0.40\textwidth}
    \centering
    \includegraphics[width=1.0\columnwidth]{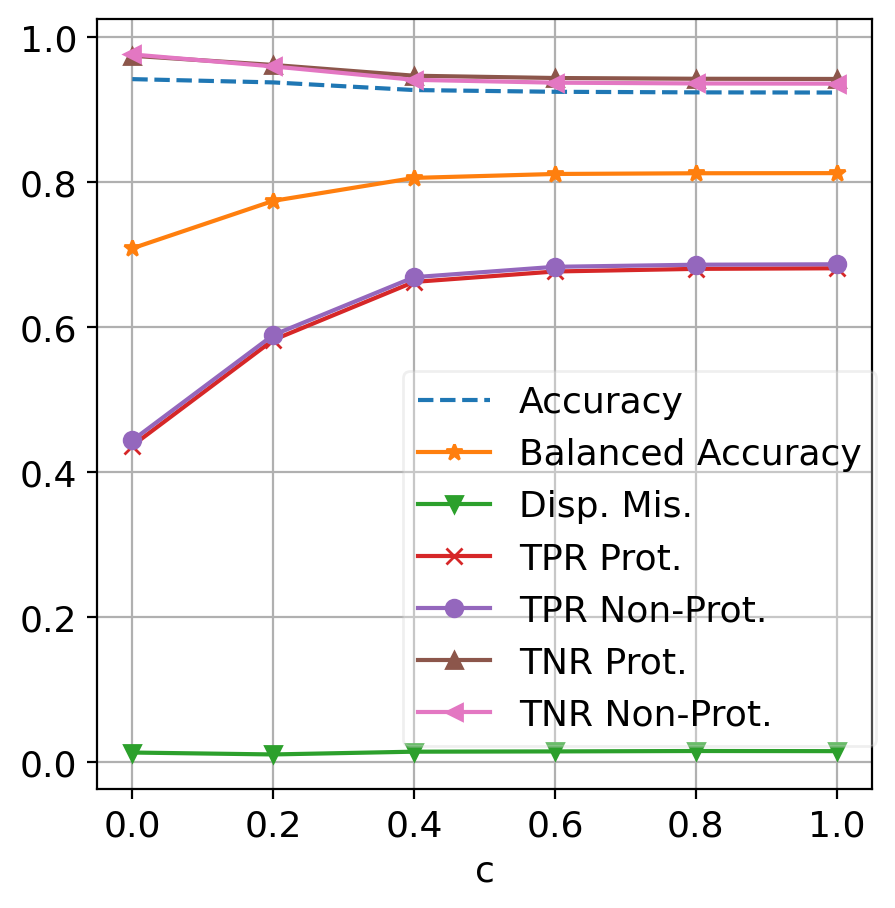}
    \caption{KDD census}
  \end{subfigure}
  \caption{Disparate Mistreatment: impact of parameter $c$}
  \label{fig:impact_of_c}
\end{figure*}

\label{sec:exp_parameter_c}
AdaFair can achieve fairness even when it does not optimize for a balanced error rate because of the cumulative fairness notion that alters the data distribution during training in the direction of fairness. In this section, we show that varying the parameter $c$, which alternates the objective goal (Equation~\eqref{eq:argmin}), does not deteriorate the ability of AdaFair to mitigate discriminatory outcomes. We show the impact of parameter $c$ for all the employed fairness notions in Figure~\ref{fig:impact_of_c}. In these figures, we plot accuracy and fairness related measures for different values of $c \in [0, 1]$. For $c=1$, the balanced error is optimized while for $c=0$, the error rate is optimized. Values in-between (we use a step of 0.2 for $c$) correspond to different balanced error and error rate combinations. We report on the average of 10 random splits for each value of $c$. Same as previously, we show only the results w.r.t disparate mistreatment and urge the interested readers to see the Appendix for the results on other fairness notions.

\noindent\textbf{Disparate Mistreatment:}
As we can see in Figure~\ref{fig:impact_of_c}, for all imbalanced datasets (Adult census, Bank, KDD census), the balanced accuracy increases with $c$. For $c=0$ (only error rate is considered), the TPRs for both groups are very low. The TPRs increase with $c$, reaching their best values at $c=1$, i.e., when balanced accuracy is considered.
TNRs decrease with $c$, though their decrease is lower than the increase of TPRs. This again supports our previous findings that AdaFair achieves parity between the two groups for both TPRs and TNRs while achieving high TPRs, on the contrary to Zafar et al. and Krasanakis et al. (c.f., Section~\ref{sec:evaluation}). For Compass, accuracy and balanced accuracy are very close, and therefore no significant differences are observed by varying $c$. 
More precisely, a comparison of the results for $c=0$ (error rate) and $c=1$ (balanced error rate) shows i) for the Adult census dataset, 12\%$\uparrow$ and 11\%$\uparrow$ increase in TPRs (for $s$, $\bar{s}$ groups respectively) and  4\%$\downarrow$ and 5\%$\downarrow$ reduction in TNRs (for $s$, $\bar{s}$ groups respectively); ii) for the Bank dataset, 16\%$\uparrow$ increase in TPRs (for both groups) and only 1\%$\downarrow$ and 2\%$\downarrow$ decrease in TNRs (for $s$ and $\bar{s}$, respectively) and iii) for the KDD census dataset, 27\%$\uparrow$ growth in TPR (for both groups) and only  4\%$\downarrow$ reduction in TNRs (for both groups).

\noindent\textbf{Conclusion:} 
The parameter $c$ is essential in the presence of class imbalance since it can help the user to examine the trade-off between TPRs and TNRs without losing AdaFair's ability to mitigate discriminatory outcomes for any given fairness measure. Moreover, in balanced datasets, any value of parameter $c$ ($\in [0,1]$) is indifferent since balanced accuracy and accuracy are similar.

\section{Conclusions and Future Work}%
\label{sec:conclusions}
We propose AdaFair, a fairness-aware boosting approach that adapts AdaBoost to fairness by changing the data distribution in each round based on both model error and cumulative fairness performance. Moreover, at post-training, AdaFair selects the best sequence of weak learners that optimizes for balanced error performance.
The notion of cumulative fairness evaluates in each round the fairness-related behaviour of the current partial ensemble and adjusts the fairness-related costs of the discriminated group accordingly. We introduce the cumulative counterparts of three popular parity-based fairness notions, namely: statistical parity, equal opportunity, and disparate mistreatment and we provide an in-depth error analysis of AdaFair for each case. 
Our experiments show that AdaFair can mitigate discriminatory outcomes while maintaining good predictive performance across both classes, even for datasets with severe class-imbalance. 

A possible extension, already discussed in Section~\ref{sec:AdaFair_tuning}, is the direct in-training mitigation of class imbalance. 
Another interesting direction is the online selection of the optimal number of boosting rounds $\theta$.
Finally, we plan to investigate further the notions of cumulative fairness  building upon ideas on combined discrimination from the legal domain~\cite{hand2011combined} .


\section*{Acknowledgements}
The work is supported by the Volkswagen Foundation project BIAS ("Bias and Discrimination in Big Data and Algorithmic Processing. Philosophical Assessments, Legal Dimensions, and Technical Solutions") within the initiative "AI and the Society of the Future".

\bibliographystyle{ieeetr} 
\bibliography{bibliography}

\newpage
\section*{Appendix}
\subsection*{Cumulative vs non-cumulative fairness}
\begin{figure*}[htp!]
 \centering
 \begin{subfigure}[t]{0.48\textwidth}
 \centering
 \includegraphics[width=1.00\columnwidth]{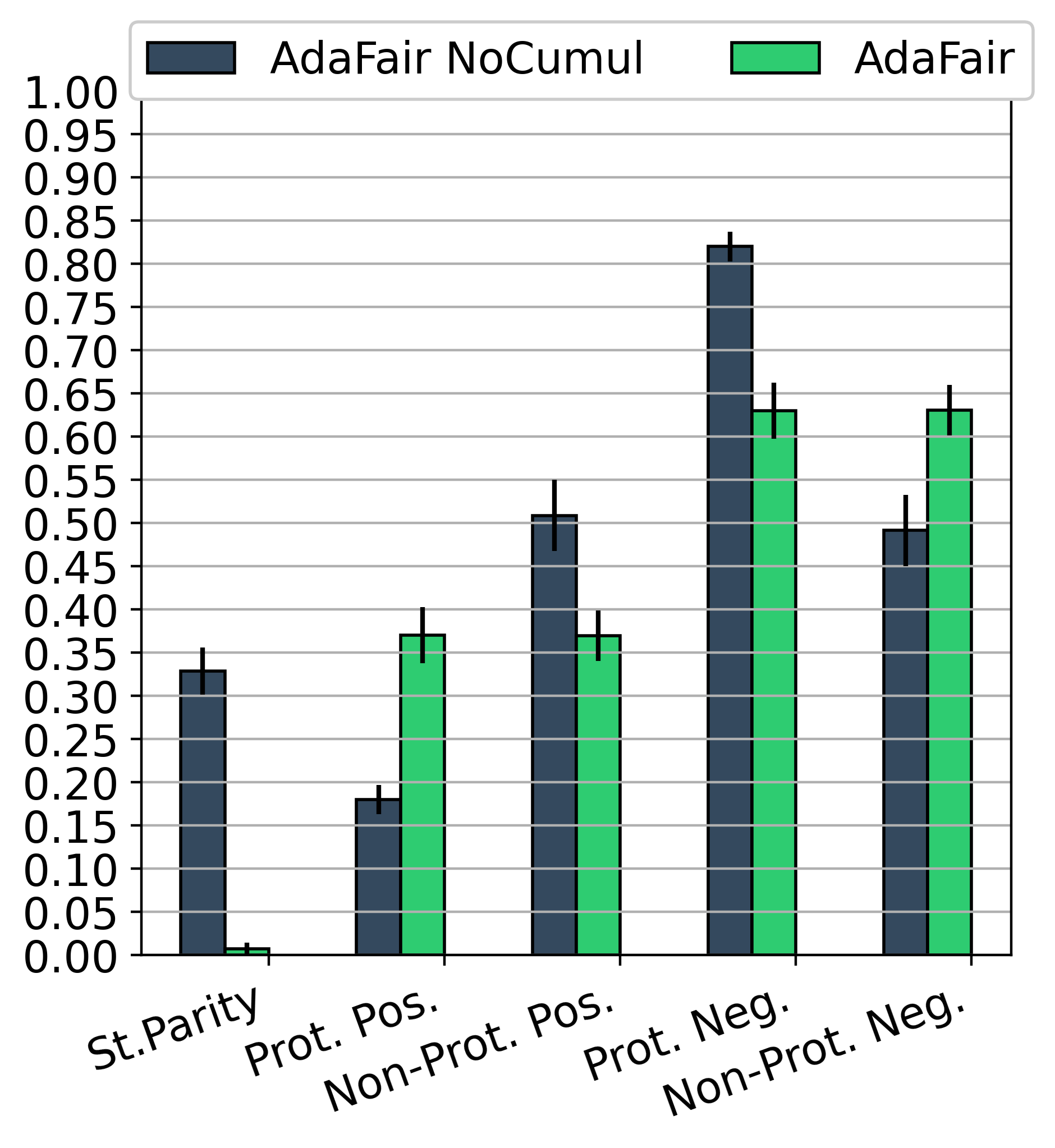}
 \caption{Adult census}
 \end{subfigure}
 \centering
 \begin{subfigure}[t]{0.48\textwidth}
 \centering
 \includegraphics[width=1.00\columnwidth]{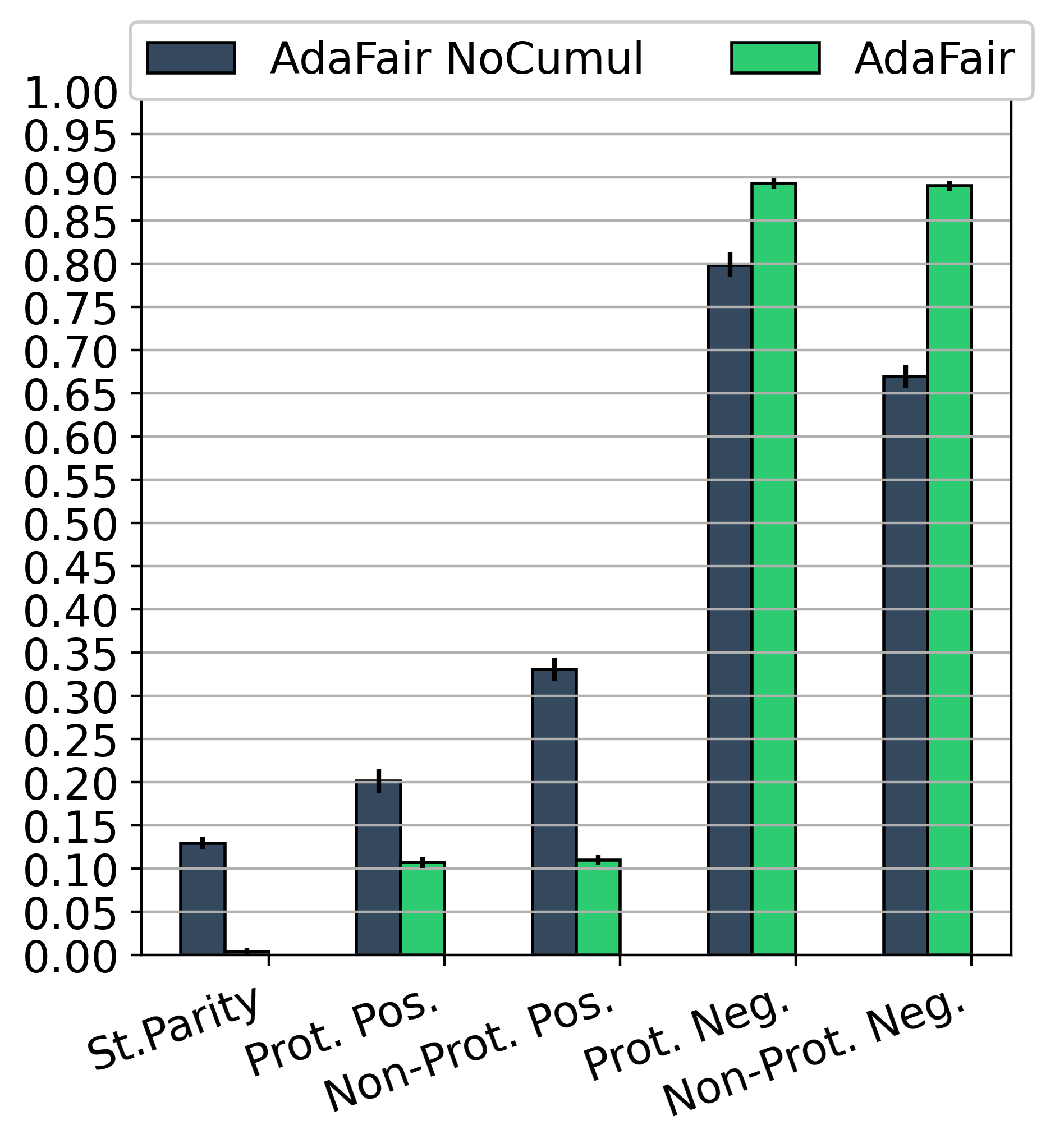}
 \caption{Bank}
 \end{subfigure}
 \begin{subfigure}[t]{0.48\textwidth}
 \centering
 \includegraphics[width=1.00\columnwidth]{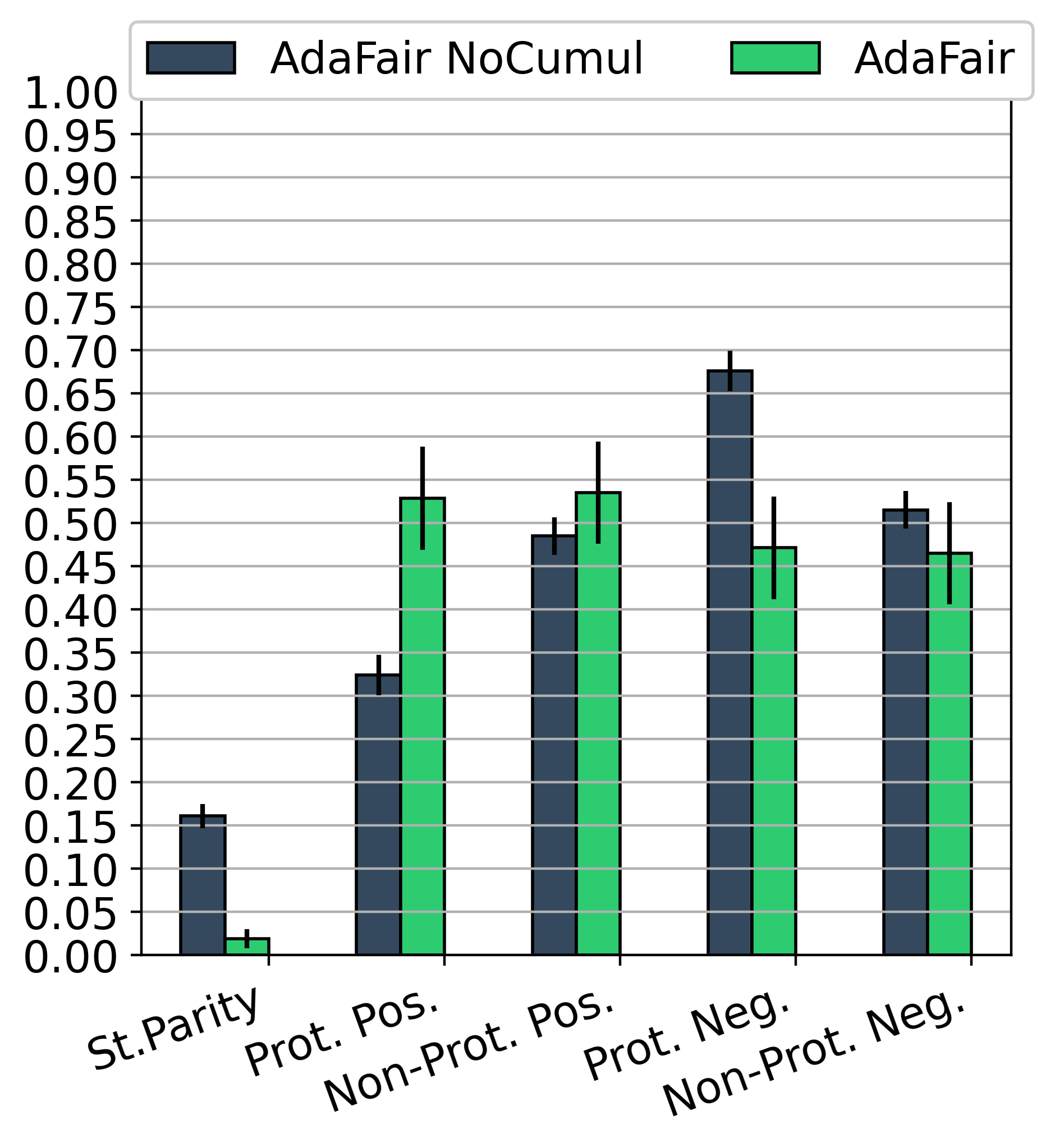}
 \caption{Compass} 
 \end{subfigure}
 \begin{subfigure}[t]{0.48\textwidth}
 \centering
 \includegraphics[width=1.00\columnwidth]{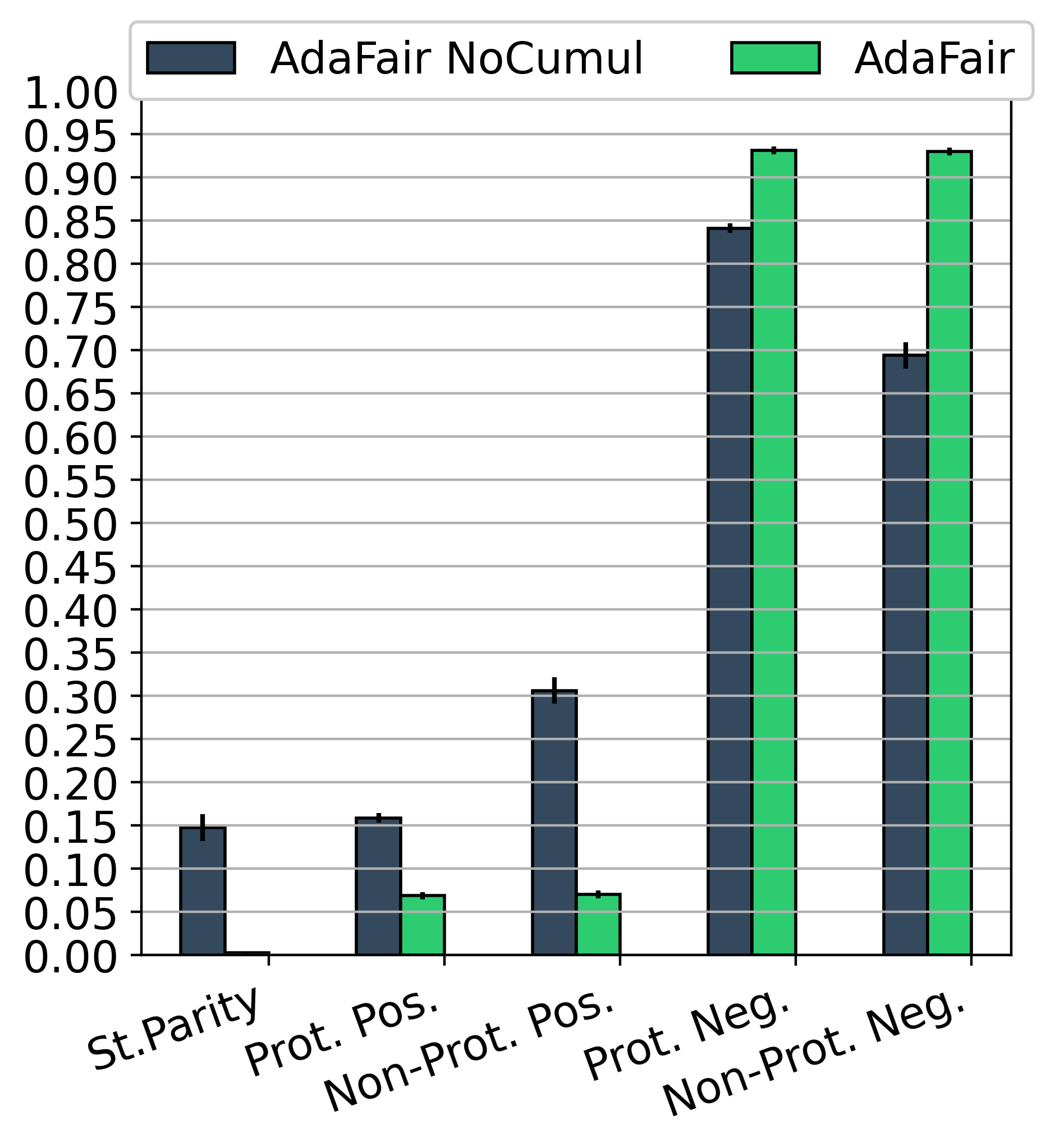}
 \caption{KDD census} 
 \end{subfigure}
 \caption{
Statistical parity: 
AdaFair vs AdaFair NoCumul} 
 \label{fig:single_performance_statistical_parity}
\end{figure*}

\noindent\textbf{Statistical Parity:}
In Figure~\ref{fig:single_performance_statistical_parity}, we show the comparison of AdaFair versus AdaFair NoCumul w.r.t statistical parity for each dataset. As we see, AdaFair NoCumul produces higher discriminatory outcomes than AdaFair on all datasets. For the Adult census dataset, we observe a 31\%$\uparrow$ increase, 12\%$\uparrow$ increase for the Bank dataset, 15\%$\uparrow$ for the Compass and 15\%$\uparrow$ for the KDD census dataset. The cumulative notion of fairness allows AdaFair to effectively mitigate the discriminatory outcomes in contrast to the non-cumulative version.

\begin{figure*}[htp!]
 \centering
 \begin{subfigure}[t]{0.48\textwidth}
 \centering
 \includegraphics[width=1.00\columnwidth]{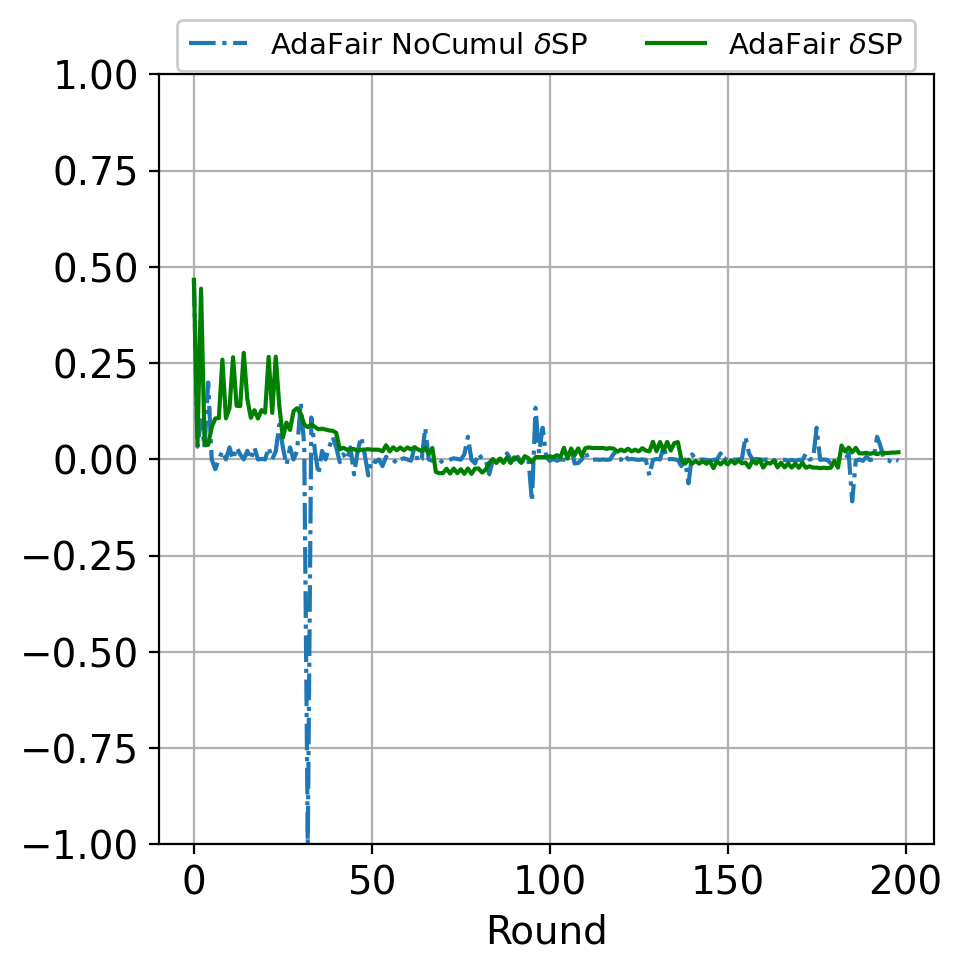}
 \caption{Adult census}
 \end{subfigure}
 \centering
 \begin{subfigure}[t]{0.48\textwidth}
 \centering
 \includegraphics[width=1.00\columnwidth]{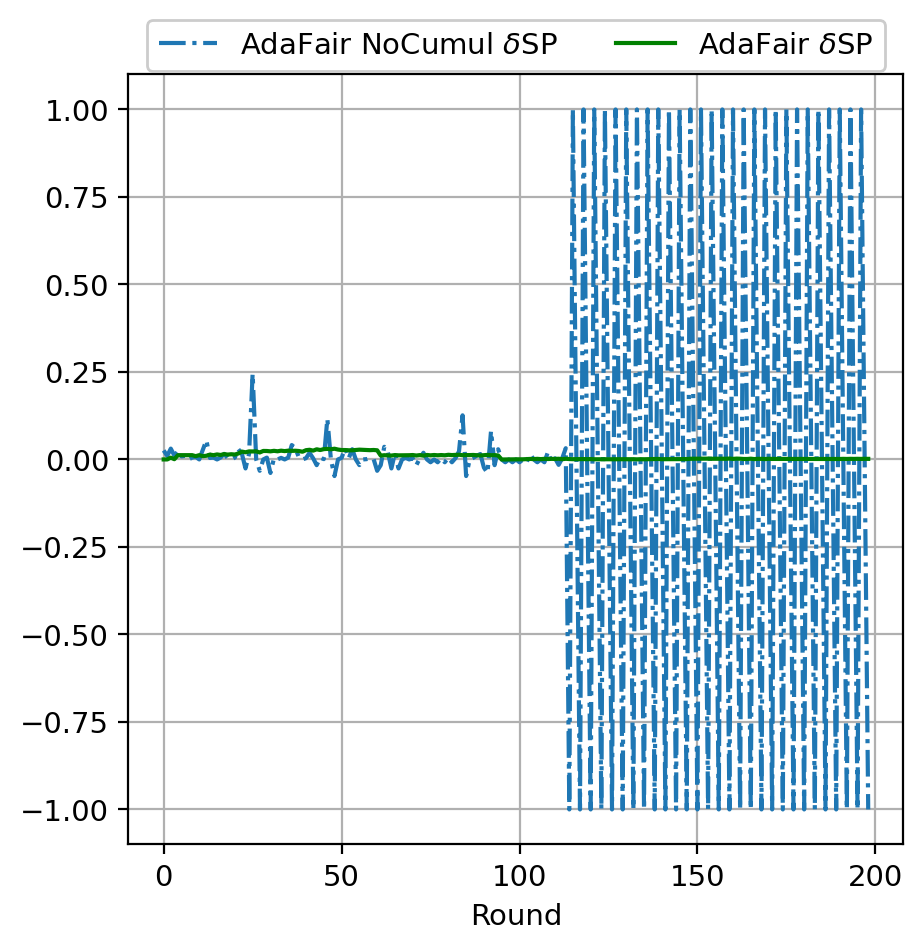}
 \caption{Bank}
 \end{subfigure}
 \begin{subfigure}[t]{0.48\textwidth}
 \centering
 \includegraphics[width=1.00\columnwidth]{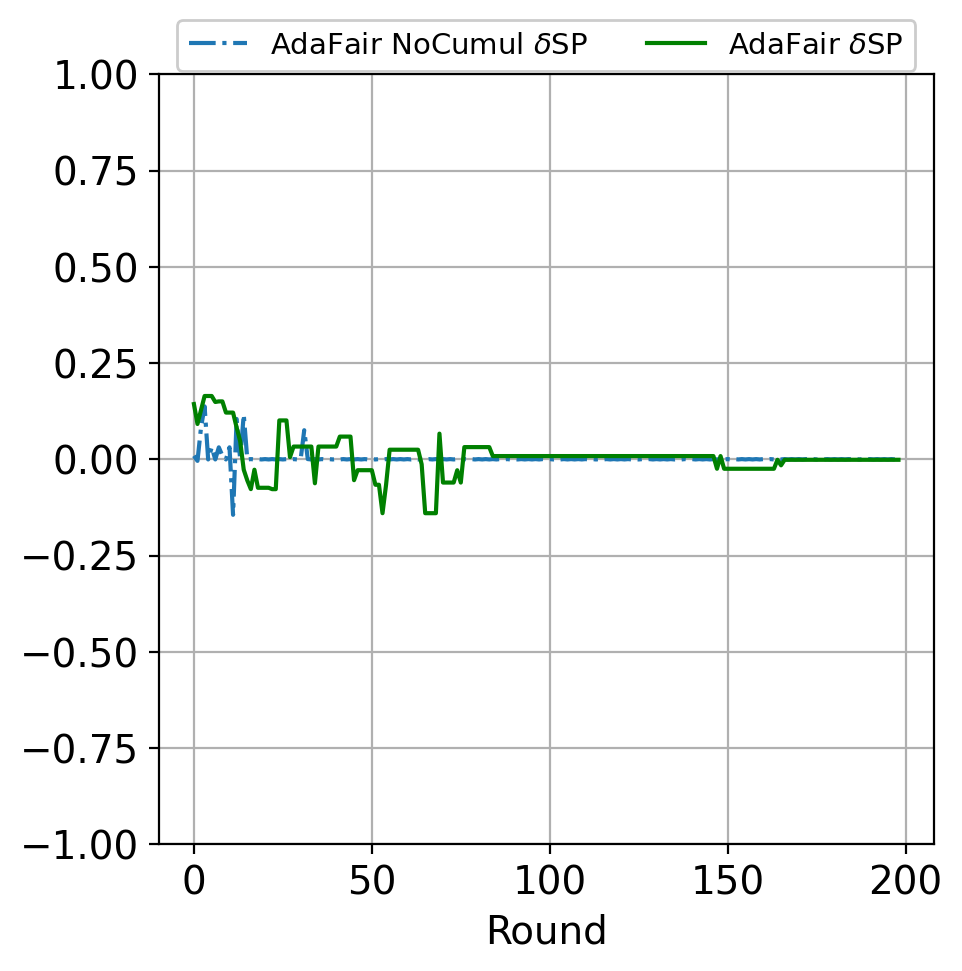}
 \caption{Compass} 
 \end{subfigure}
 \begin{subfigure}[t]{0.48\textwidth}
 \centering
 \includegraphics[width=1.00\columnwidth]{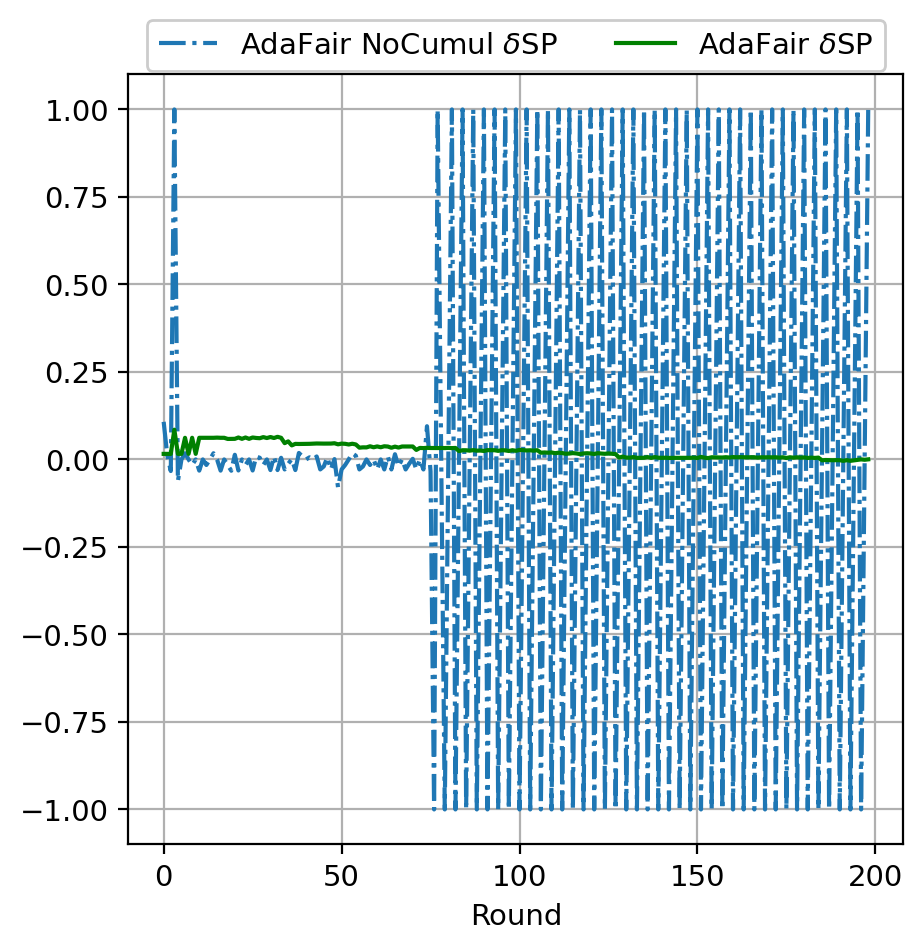}
 \caption{KDD census} 
 \end{subfigure}
 \caption{Statistical parity, fairness-related costs per boosting round: AdaFair vs AdaFair NoCumul} 
 \label{fig:round_costs_stat_parity}
\end{figure*}

In Figure~\ref{fig:round_costs_stat_parity}, we compare the per round $\delta SP$ of AdaFair NoCumul and AdaFair. $\delta SP$ refers to the fairness-related cost ($u$) that is assigned to instances based on the discriminatory behaviour of the model (Equation~\eqref{eq:fairnessCostsSP}). We observe that AdaFair NoCumul produces fairness-related costs, which highly fluctuate, in contrast to AdaFair, in all the datasets. The non-cumulative version cannot stabilize the fairness-related costs since it depends on the behaviour of individual weak learns rather than the cumulative behaviour of the model.

\begin{figure*}[htp!]
 \centering
 \begin{subfigure}[t]{0.48\textwidth}
 \centering
 \includegraphics[width=1.00\columnwidth]{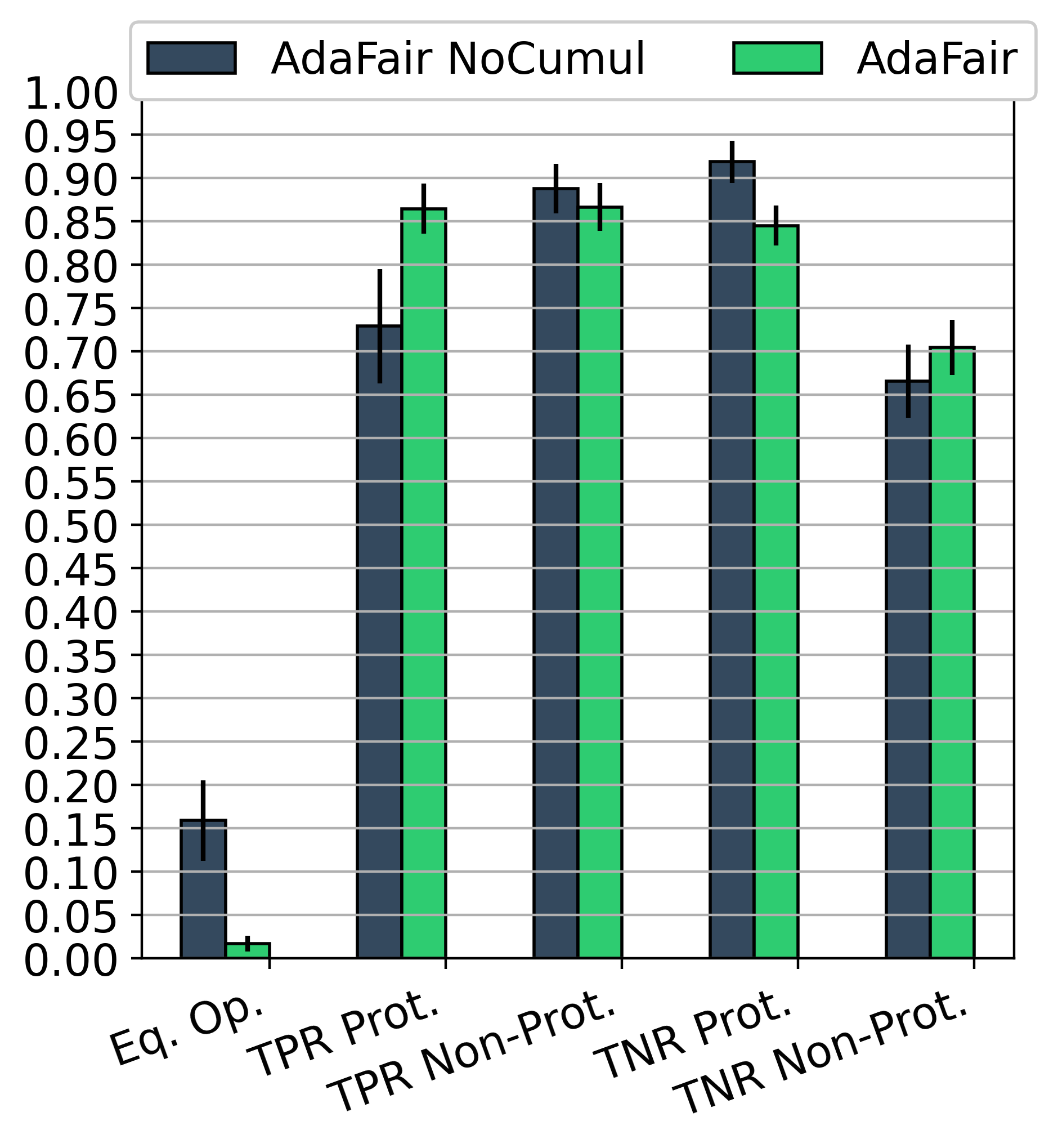}
 \caption{Adult census}
 \end{subfigure}
 \centering
 \begin{subfigure}[t]{0.48\textwidth}
 \centering
 \includegraphics[width=1.00\columnwidth]{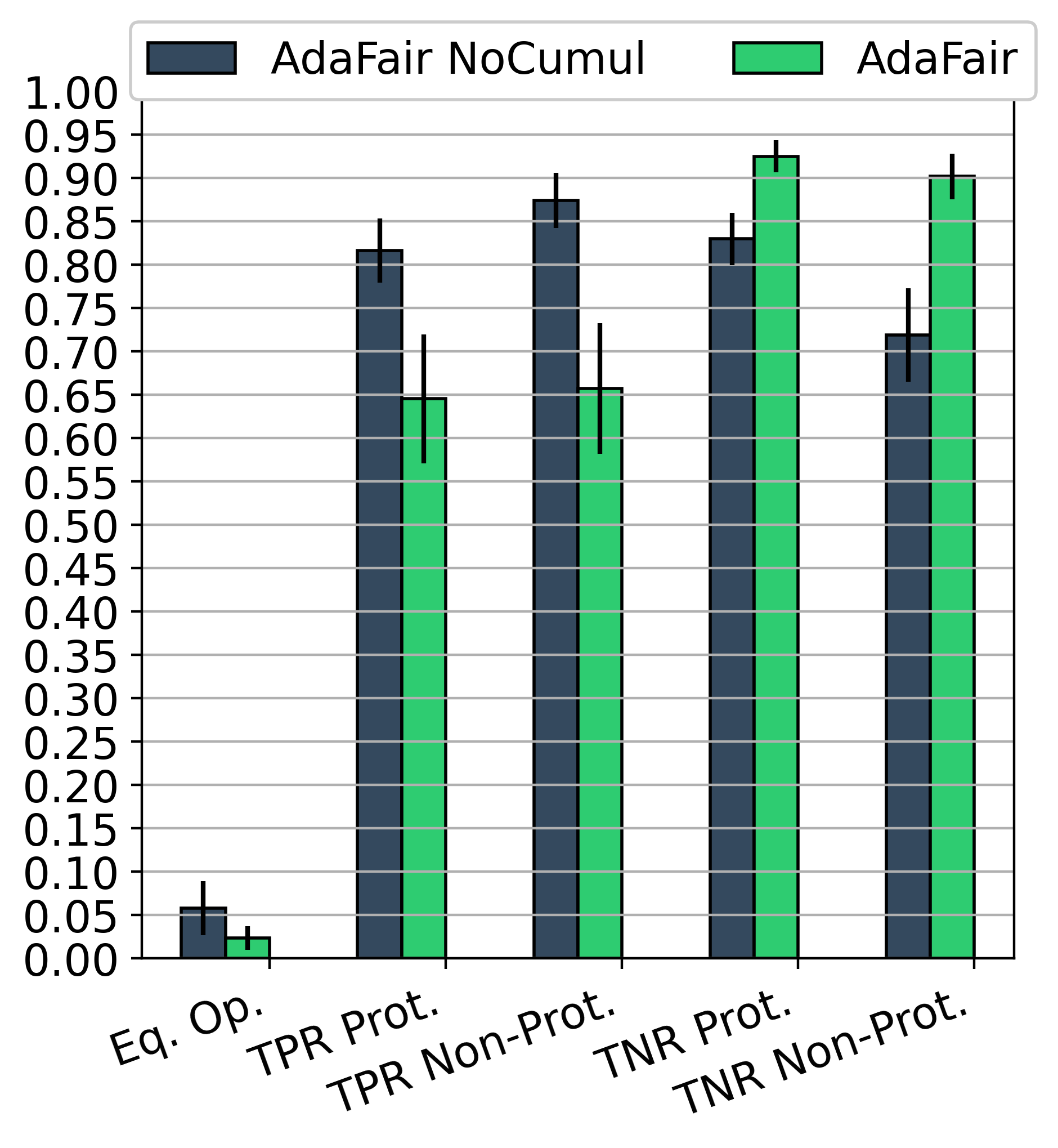}
 \caption{Bank}
 \end{subfigure}
 \begin{subfigure}[t]{0.48\textwidth}
 \centering
 \includegraphics[width=1.00\columnwidth]{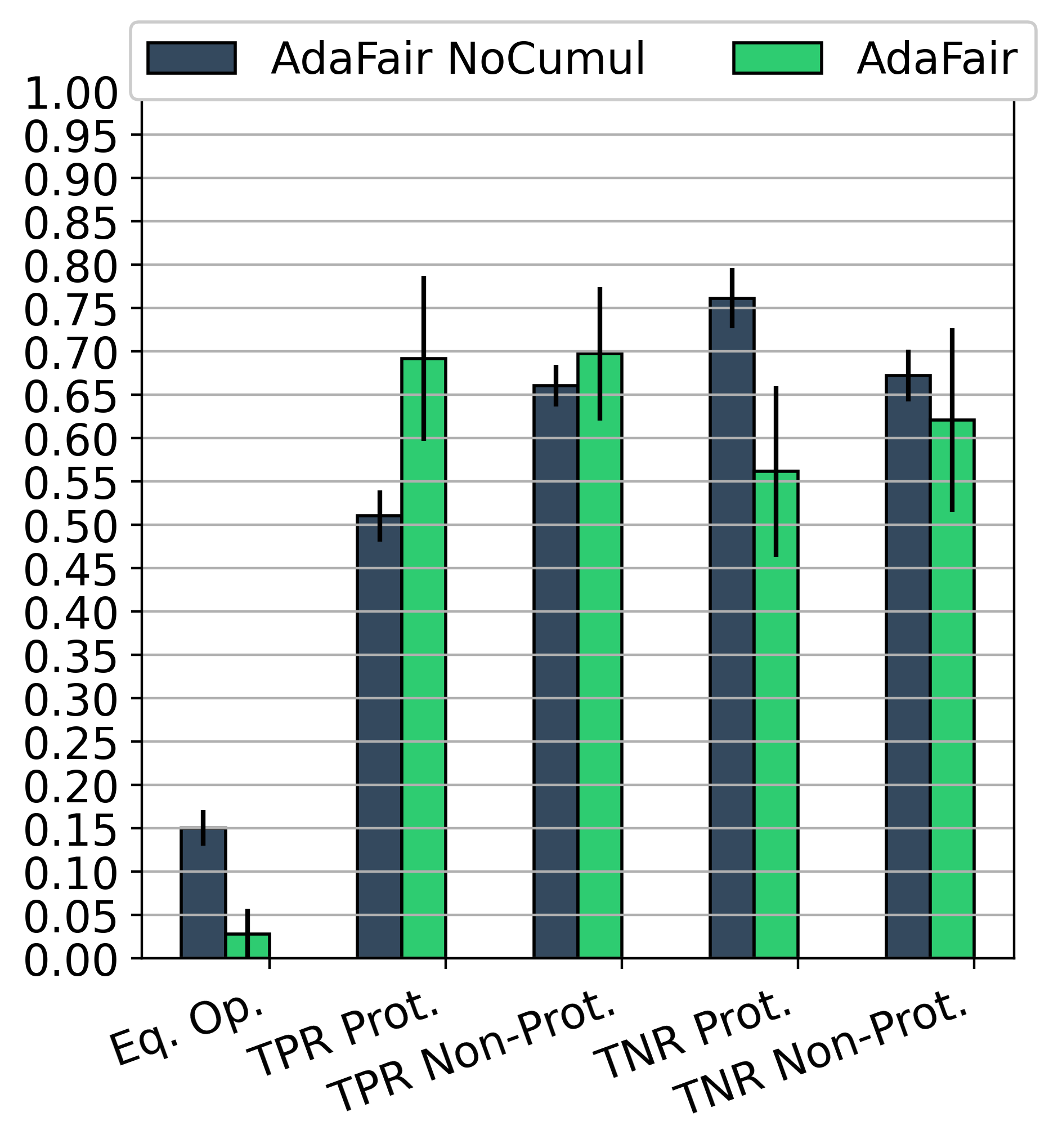}
 \caption{Compass} 
 \end{subfigure}
 \begin{subfigure}[t]{0.48\textwidth}
 \centering
 \includegraphics[width=1.00\columnwidth]{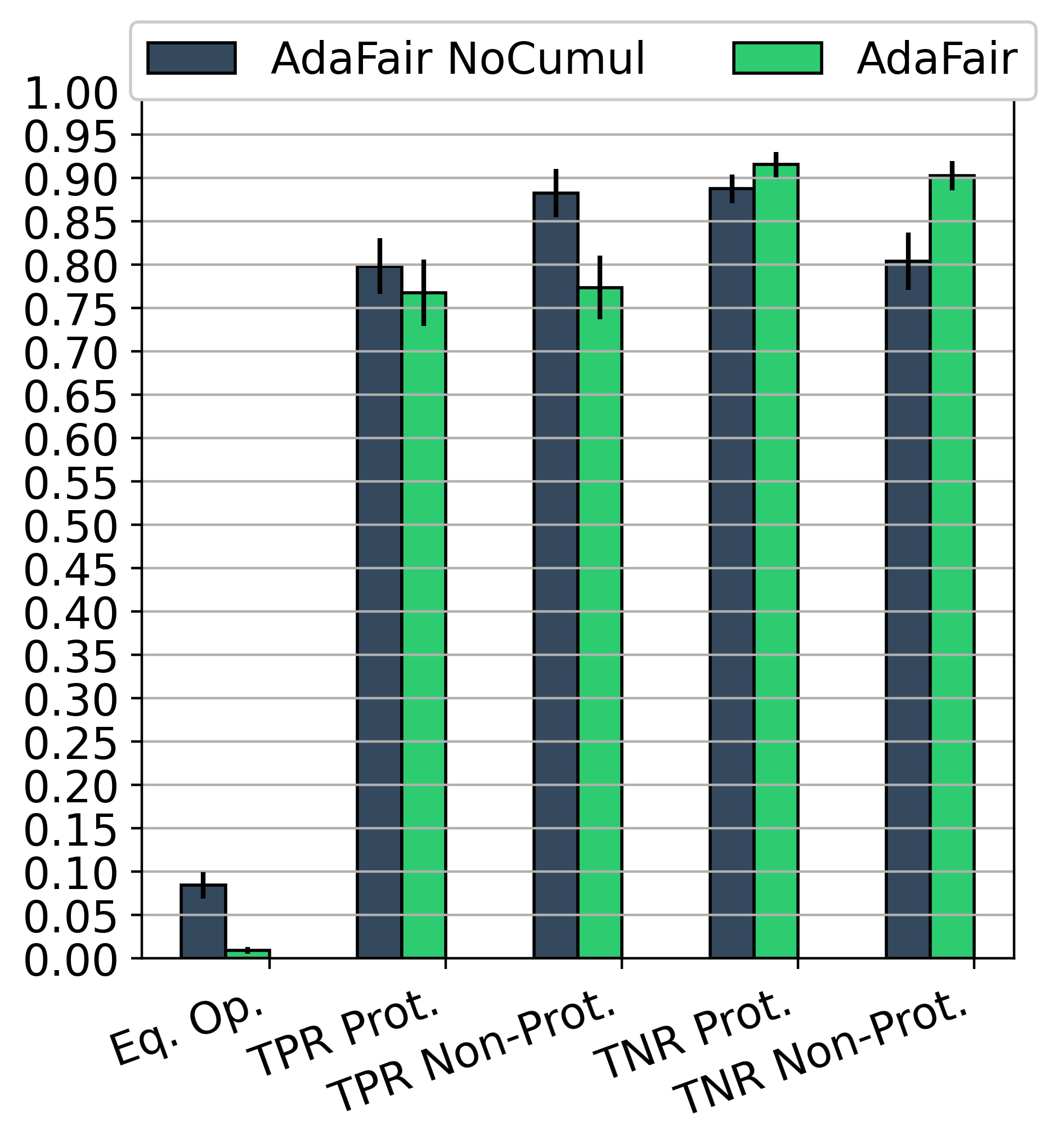}
 \caption{KDD census} 
 \end{subfigure}
 \caption{
Equal opportunity: 
AdaFair vs AdaFair NoCumul} 
 \label{fig:single_performance_equal_opportunity}
\end{figure*}

\noindent\textbf{Equal Opportunity:}
In Figure~\ref{fig:single_performance_equal_opportunity}, we show the comparison of AdaFair versus AdaFair NoCumul w.r.t equal opportunity for each dataset. Same as in the statistical parity case, AdaFair NoCumul produces more discriminatory outcomes in contrast to AdaFair. For the Adult census dataset, there is a 15\%$\uparrow$ increase, 2\%$\uparrow$ increase for the Bank dataset, 12\%$\uparrow$ increase for the Compass, and 8\%$\uparrow$ increase for the KDD census dataset.

\begin{figure*}[htp!]
 \centering
 \begin{subfigure}[t]{0.48\textwidth}
 \centering
 \includegraphics[width=1.0\columnwidth]{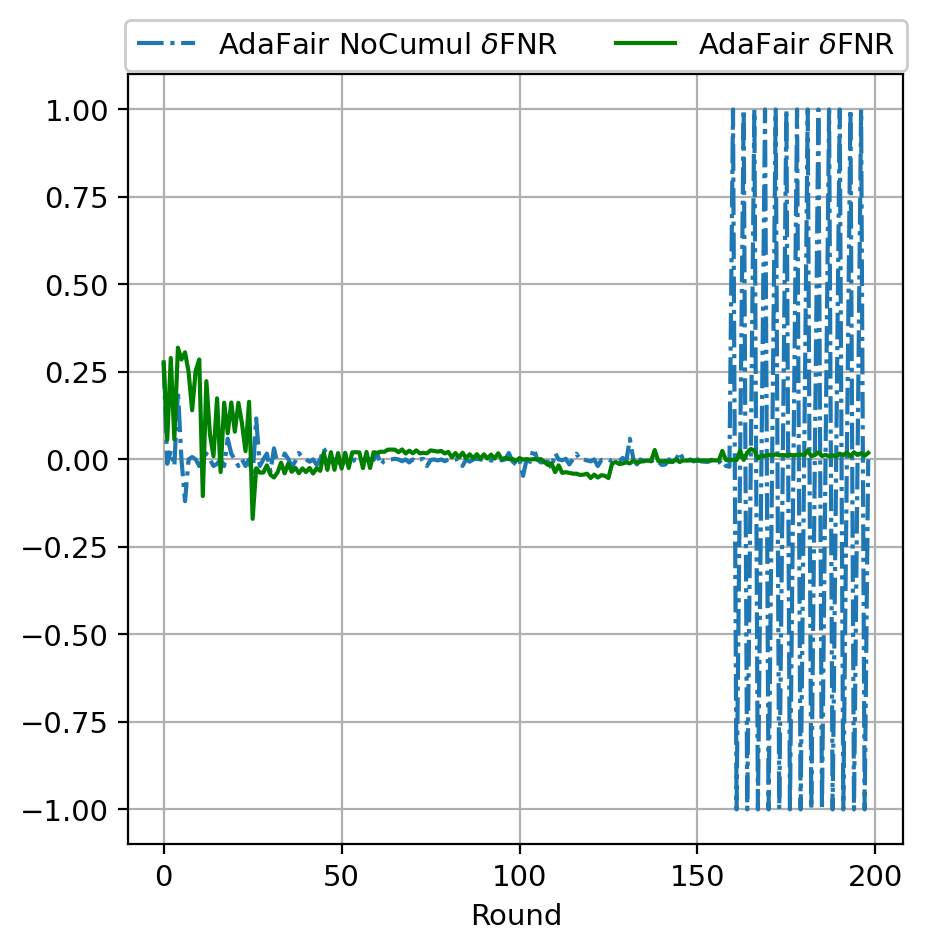}
 \caption{Adult census}
 \end{subfigure}
 \centering
 \begin{subfigure}[t]{0.48\textwidth}
 \centering
 \includegraphics[width=1.0\columnwidth]{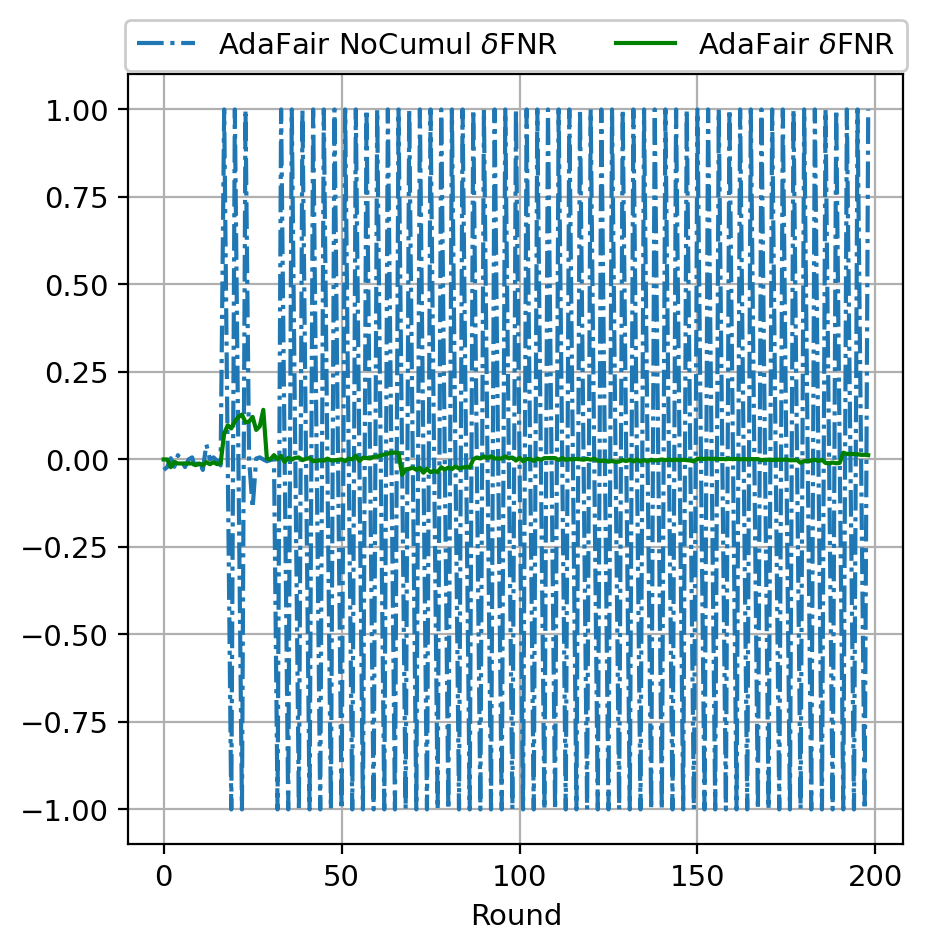}
 \caption{Bank}
 \end{subfigure}
 \begin{subfigure}[t]{0.48\textwidth}
 \centering
 \includegraphics[width=1.0\columnwidth]{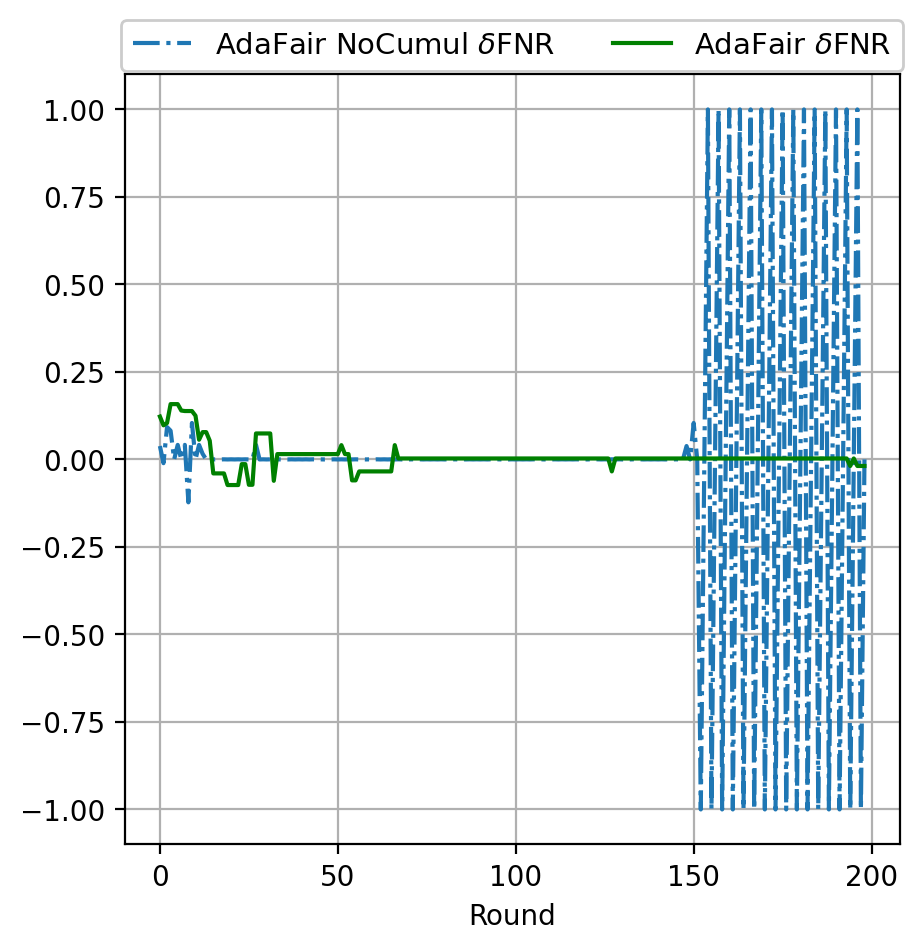}
 \caption{Compass} 
 \end{subfigure}
 \begin{subfigure}[t]{0.48\textwidth}
 \centering
 \includegraphics[width=1.0\columnwidth]{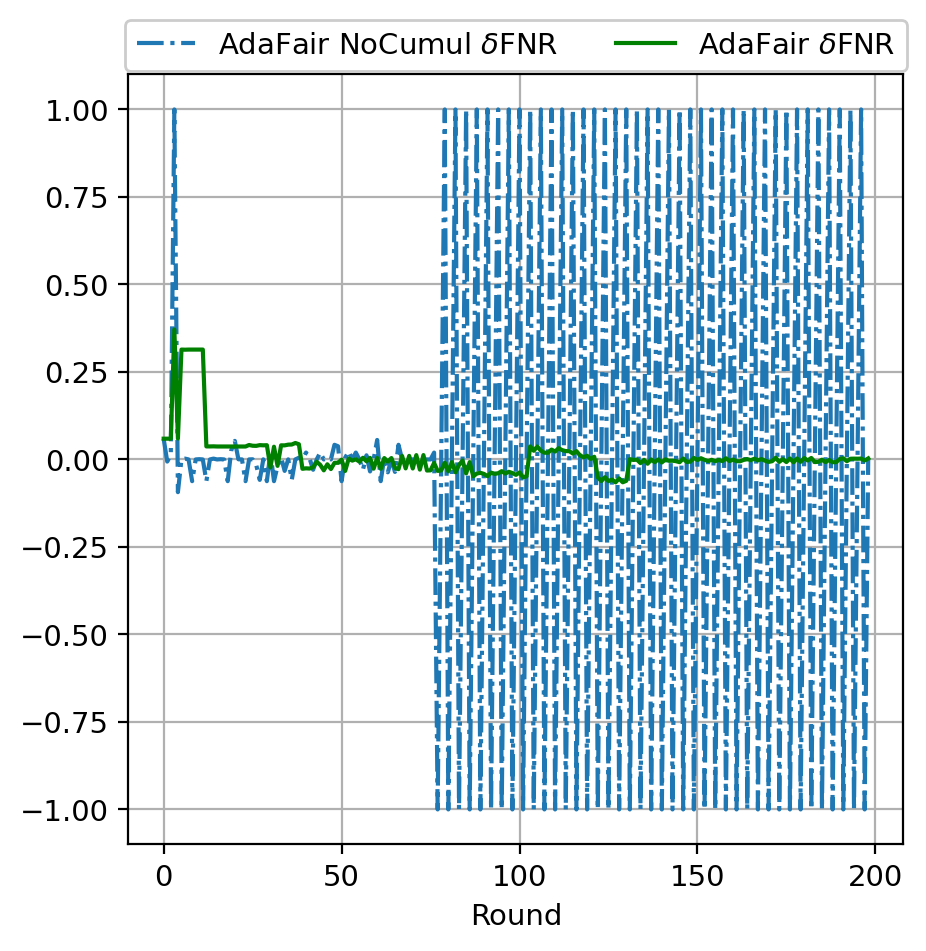}
 \caption{KDD census} 
 \end{subfigure}
 \caption{Equal opportunity, fairness-related costs per boosting round: AdaFair vs AdaFair NoCumul} 
 \label{fig:round_costs_eq_op}
\end{figure*}

Similar behaviour to statistical parity is also observed in Figure~\ref{fig:round_costs_eq_op}, where we report $\delta FNR$ values for the cumulative and non-cumulative approaches; $\delta FNR$ values are employed as fairness-related costs and are derived from Equation~\eqref{eq:fairnessCostsEQOP}. The non-cumulative version is unstable and produces highly fluctuating fairness-related costs in contrast to AdaFair in all datasets.

\subsection*{The effect of balanced error}
\begin{figure*}[htp]
  \centering
  \begin{subfigure}[t]{0.48\textwidth}
    \centering
  \includegraphics[width=1.0025\columnwidth]{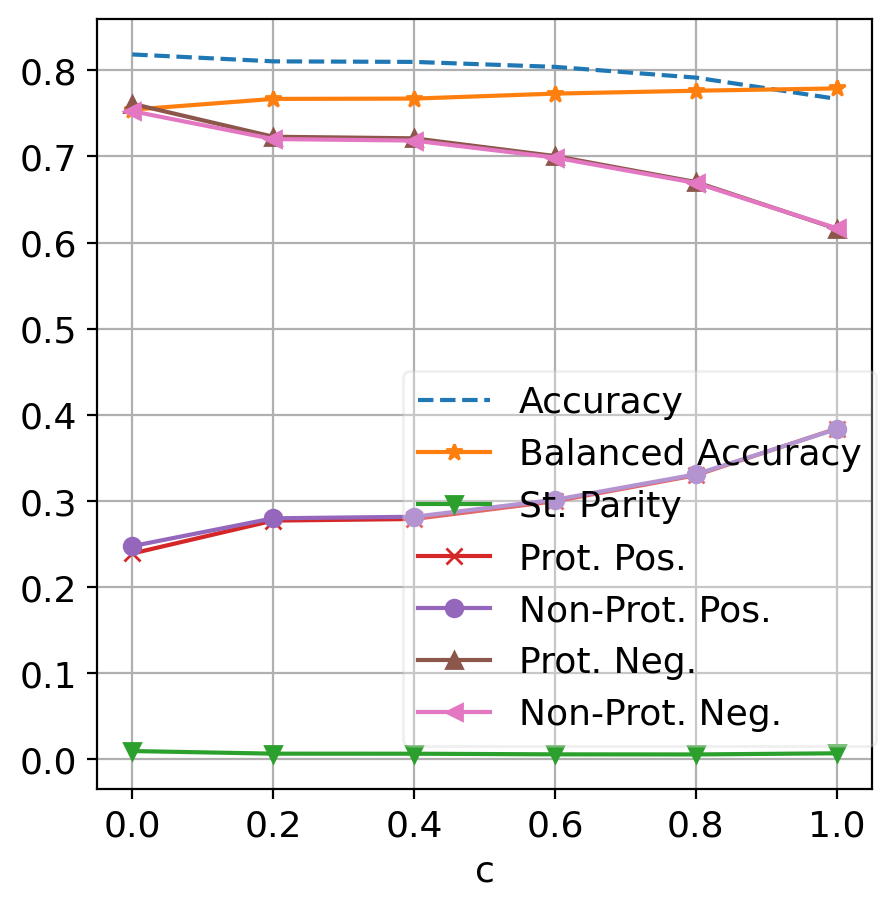}
    \caption{Adult census}
  \end{subfigure}
  \begin{subfigure}[t]{0.48\textwidth}
    \centering
  \includegraphics[width=1.0025\columnwidth]{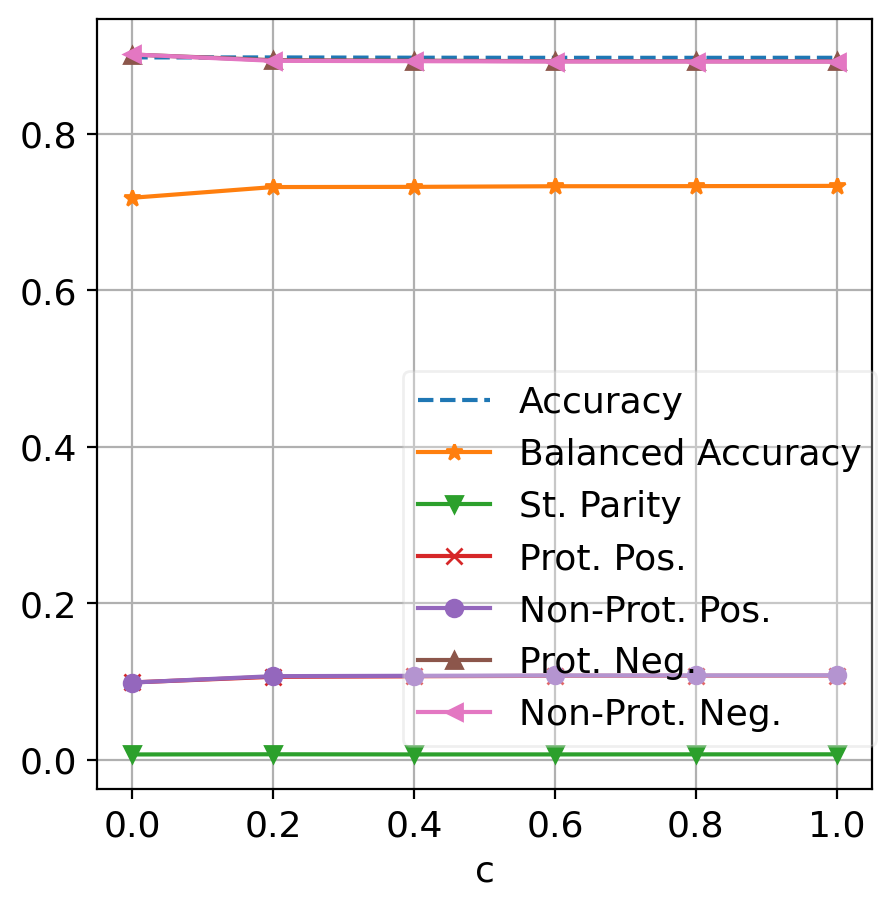}
    \caption{Bank}
  \end{subfigure}
  \begin{subfigure}[t]{0.48\textwidth}
    \centering
    \includegraphics[width=1.0025\columnwidth]{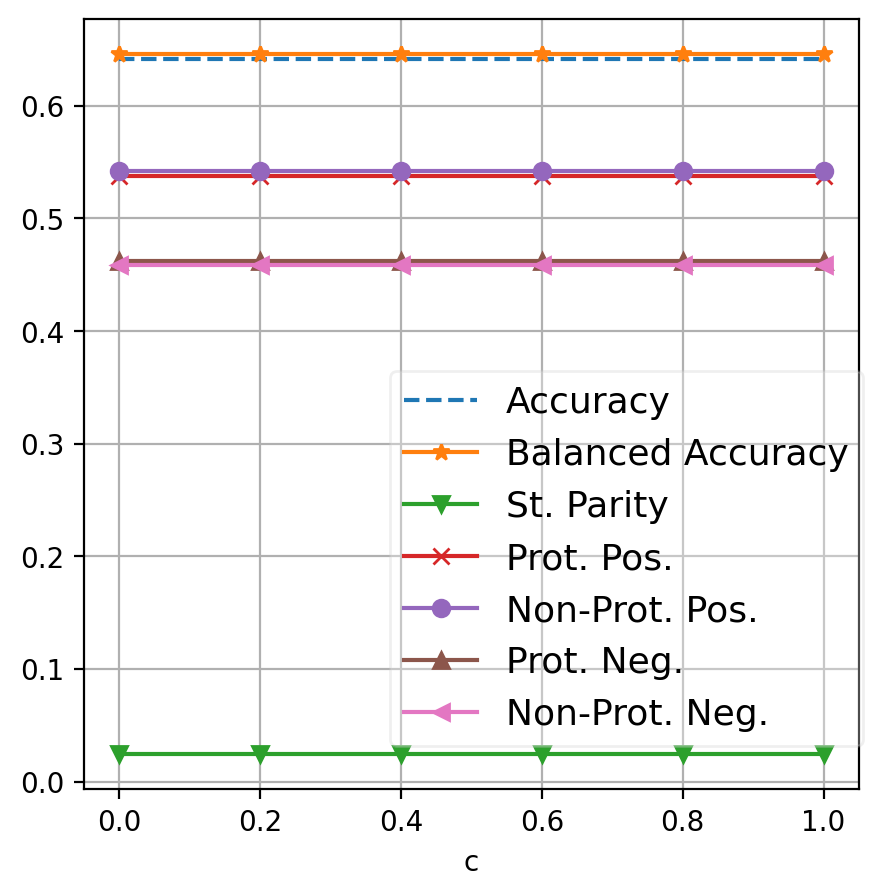}
    \caption{Compass}
  \end{subfigure}
    \begin{subfigure}[t]{0.48\textwidth}
    \centering
    \includegraphics[width=1.0025\columnwidth]{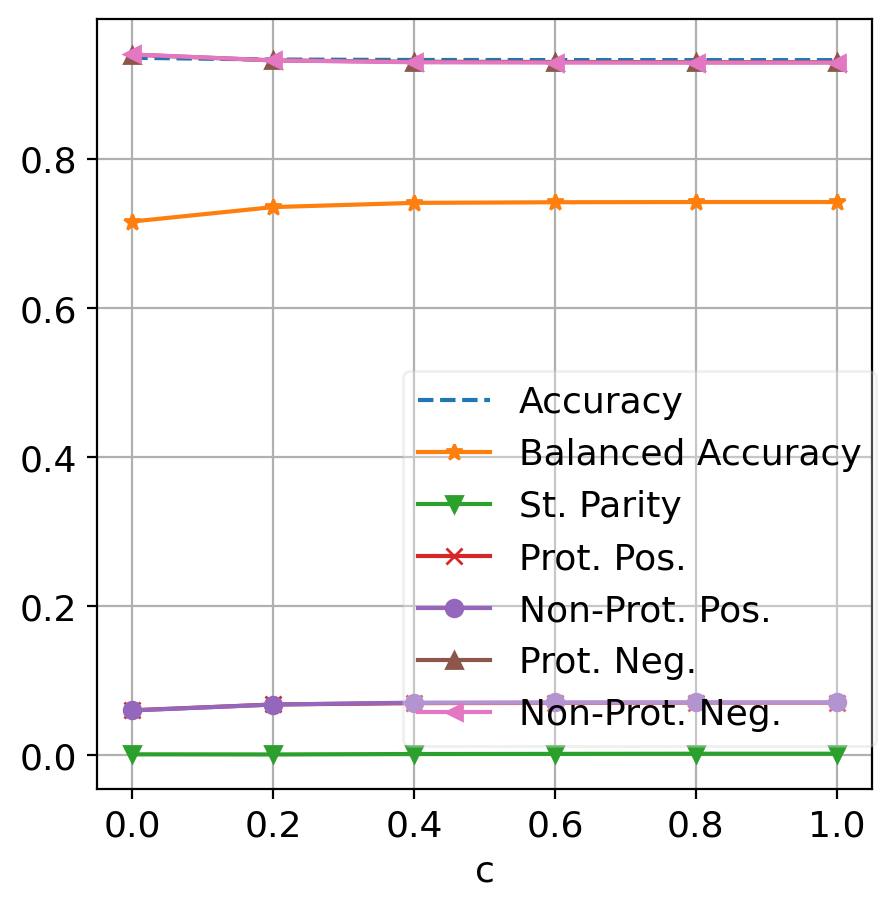}
    \caption{KDD census}
  \end{subfigure}
  \caption{Statistical Parity: impact of parameter $c$}
  \label{fig:impact_of_c_for_statistical_parity}
\end{figure*}

\begin{figure*}[htp]
  \centering
  \begin{subfigure}[t]{0.48\textwidth}
    \centering
  \includegraphics[width=1.0025\columnwidth]{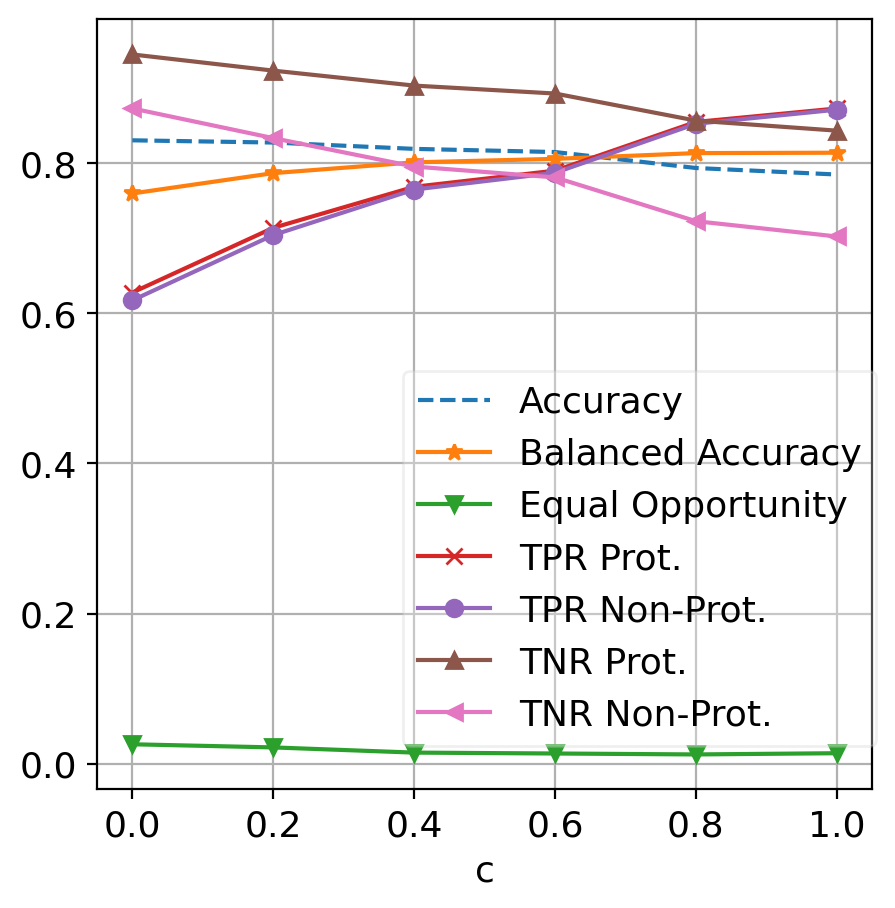}
    \caption{Adult census}
  \end{subfigure}
  \begin{subfigure}[t]{0.48\textwidth}
    \centering
  \includegraphics[width=1.0025\columnwidth]{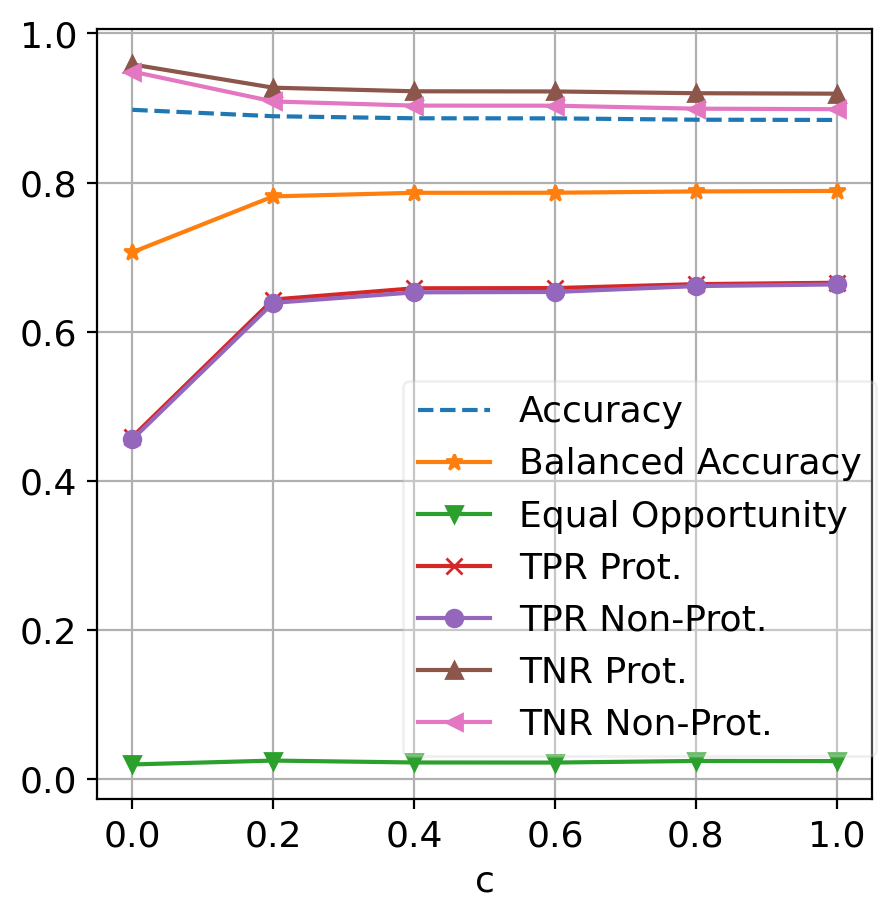}
    \caption{Bank}
  \end{subfigure}
  \begin{subfigure}[t]{0.48\textwidth}
    \centering
    \includegraphics[width=1.0025\columnwidth]{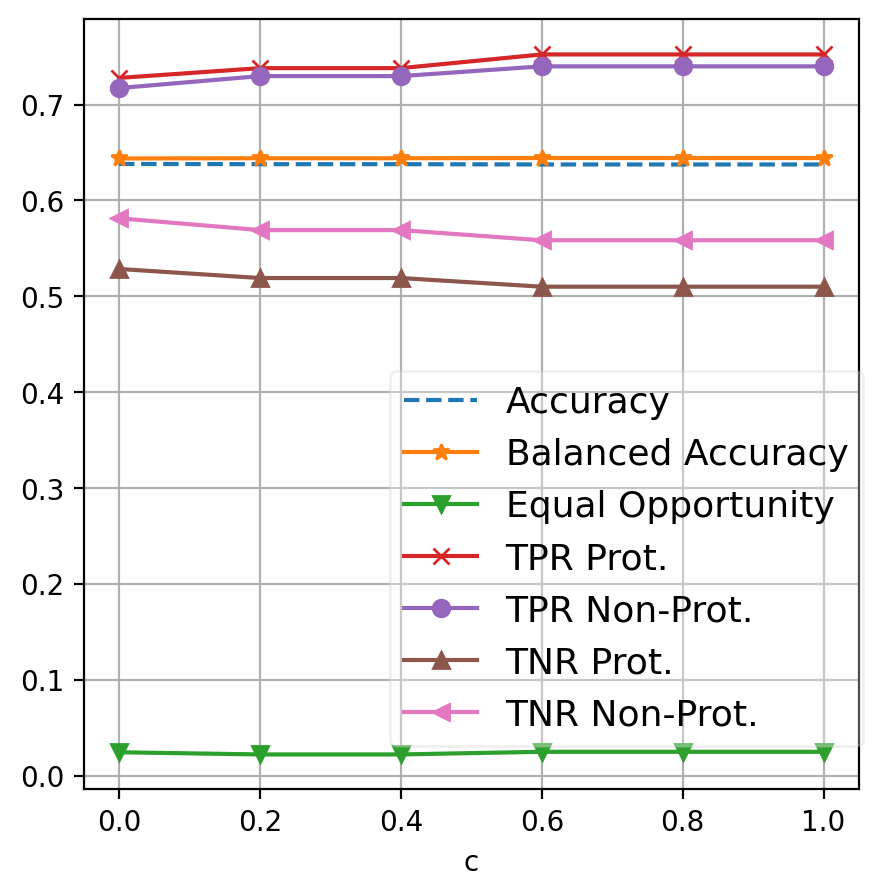}
    \caption{Compass}
  \end{subfigure}
    \begin{subfigure}[t]{0.48\textwidth}
    \centering
    \includegraphics[width=1.0025\columnwidth]{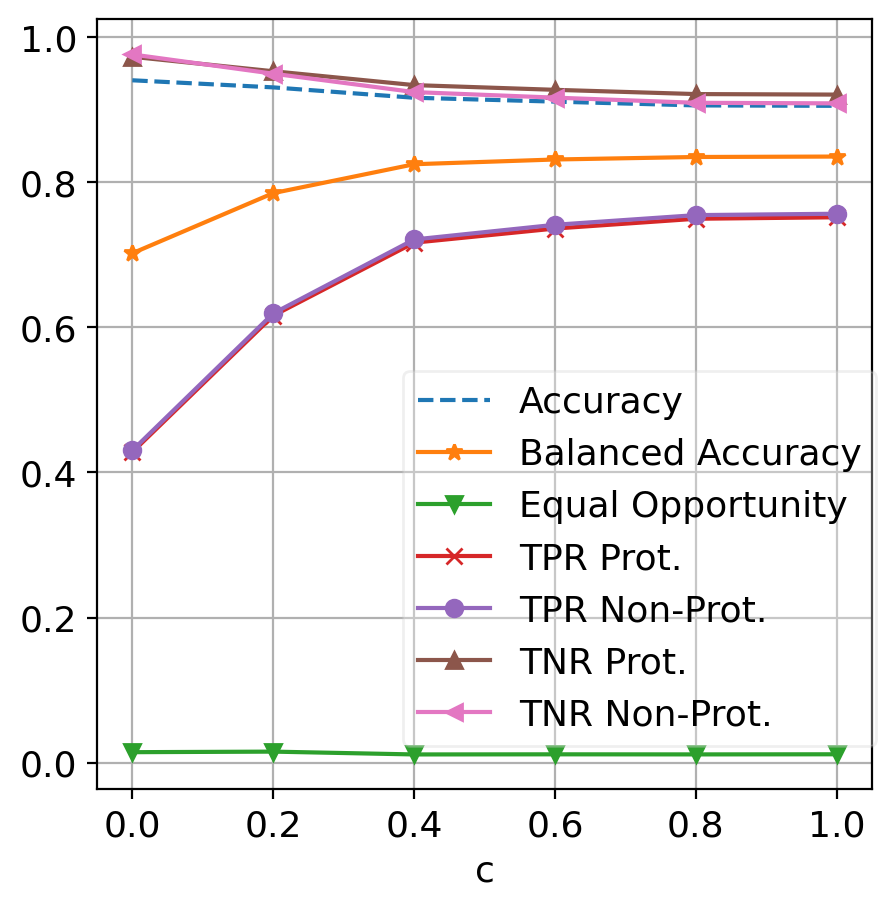}
    \caption{KDD census}
  \end{subfigure}
  \caption{Equal Opportunity: impact of parameter $c$}
  \label{fig:impact_of_c_for_equal_opportunity}
\end{figure*}

We show the impact of parameter $c$ for all the employed fairness notions in Figures~\ref{fig:impact_of_c_for_statistical_parity} and~\ref{fig:impact_of_c_for_equal_opportunity}.
 
\noindent\textbf{Statistical Parity:}
In Figure~\ref{fig:impact_of_c_for_statistical_parity}, we show the impact of parameter $c$ in case of statistical parity. As we observe, all the imbalanced datasets show the worst performance in terms of balanced accuracy when $c=0$; however, statistical parity is close to 0. As the parameter $c$ increases, the balanced accuracy increases and the statistical parity remains close to 0. However, in the case of statistical parity, we observe that the balanced accuracy is not affected significantly in contrast to the other two fairness notions. Such behaviour is caused due to the fairness' notion, which forces parity between protected and non-protected groups on the predicted outcomes; thus, statistical parity can force AdaFair to predict more instances in the positive class indirectly. 

\noindent\textbf{Equal Opportunity:}
In Figure~\ref{fig:impact_of_c_for_equal_opportunity}, we show the impact of $c$ when AdaFair tunes for equal opportunity. Similar to disparate mistreatment, AdaFair can maintain its low discrimination values w.r.t equal opportunity and at the same time increase the balanced accuracy as the parameter $c$ increases. E.g., AdaFair's balanced accuracy increases 8\% for $c=0$ to $c=1$ and at the same time equal opportunity is close to 0. This behaviour is similar for all the employed imbalanced datasets. For the Compass dataset, the parameter $c$ does not affect the performance significantly since the dataset is class balanced.

\end{document}